%% file: main.tex
\PassOptionsToPackage{hyphens}{url}
\PassOptionsToPackage{table,dvipsnames}{xcolor}
\documentclass[11pt, a4paper]{yihaoqiao_preprint}

\usepackage[utf8]{inputenc}
\usepackage[T1]{fontenc}
\usepackage{url}
\usepackage{graphicx}
\usepackage{booktabs}
\usepackage{amsmath, amssymb, amsfonts, bm}
\usepackage{xcolor}
\usepackage{microtype}
\usepackage{xspace}
\definecolor{themec}{HTML}{8C3A4B}
\usepackage[pagebackref=false, breaklinks=true, colorlinks,
            allcolors=themec, bookmarks=false]{hyperref}
\usepackage{natbib}
\usepackage{pgfplots}
\pgfplotsset{compat=1.18}
\usepgfplotslibrary{groupplots,colorbrewer}
\usetikzlibrary{matrix,positioning,calc,arrows.meta}
\usepackage{multirow}
\usepackage{wrapfig}
\usepackage[section]{placeins}
\usepackage{nicematrix}
\usepackage{algorithm}
\usepackage{algpseudocode}
\usepackage[capitalise,noabbrev]{cleveref}
\usepackage{titletoc}%
\dottedcontents{subsection}[4.3em]{}{2.8em}{1pc}%

\input{math_commands.tex}

\renewcommand{\eqref}[1]{\textup{(\ref{#1})}}
\newcommand{\one}{\mathbf{1}}
\newcommand{\tr}{\operatorname{tr}}
\DeclareMathOperator{\diag}{diag}
\input{tables/numbers}
\newcommand{\figslot}[2]{\includegraphics[width=#2]{figures/#1.pdf}}
\setcitestyle{authoryear,round}

\input{style_macros}


\input{figures/pgfplots_style}

\renewcommand{\runningtitle}{RoFE: Rotated Frozen Embeddings for Low-Data Multimodal Tabular Learning}

\makeatletter
\patchcmd{\renderheaderbox}{\textbf{Project Website:}}{\textbf{Project page:}}{}%
  {\errmessage{Could not patch the project page label in \string\renderheaderbox}}
\makeatother
\renewcommand\urlheaderfont{\color{themec}\fontsize{8}{10}\selectfont}
\codelink{https://github.com/yihao-qiao/RoFE}
\projecturl{https://yihao-qiao.github.io/RoFE}

\title{RoFE: Rotated Frozen Embeddings for Low-Data Multimodal Tabular Learning}

\makeatletter\renewcommand\AB@authnote[1]{}\renewcommand\AB@affilnote[1]{}\makeatother
\author{Yihao Qiao}
\author{Jiajian Ma}
\author{Wushuang Rui}
\author{Abhishek Jha}
\author{Aarti Darji}
\author{Yiqiu Shen}

\affil{New York University}

\begin{abstract}
\input{chapters_arxiv/0_abstract_arxiv}
\end{abstract}

\begin{document}
\maketitle

\newenvironment{Itemize}{
    \begin{itemize}[leftmargin=*]
    \setlength{\itemsep}{0pt}
    \setlength{\topsep}{0pt}
    \setlength{\partopsep}{0pt}
    \setlength{\parskip}{1pt}}
{\end{itemize}}
\setlength{\leftmargini}{9pt}

\input{chapters_arxiv/1_intro_arxiv}
\input{chapters_arxiv/2_related_arxiv}
\input{chapters_arxiv/2_background_arxiv}
\input{chapters_arxiv/3_method_arxiv}
\input{chapters_arxiv/4_exp_arxiv}
\input{chapters_arxiv/5_conclusion_arxiv}

\FloatBarrier
\bibliographystyle{plainnat}
\bibliography{references}

\clearpage
\appendix
\crefalias{section}{appendix}%
\crefalias{subsection}{appendix}%
\crefname{appendix}{Appendix}{Appendices}
\Crefname{appendix}{Appendix}{Appendices}
\startcontents[appendix]
\section*{Appendix Contents}
\printcontents[appendix]{l}{1}{\setcounter{tocdepth}{2}}
\input{chapters_arxiv/9_appendix_arxiv}

\end{document}

%% file: math_commands.tex
\usepackage{amsmath,amsfonts,bm}

\def\eqref#1{equation~\ref{#1}}

\def\1{\bm{1}}

\DeclareMathAlphabet{\mathsfit}{\encodingdefault}{\sfdefault}{m}{sl}
\SetMathAlphabet{\mathsfit}{bold}{\encodingdefault}{\sfdefault}{bx}{n}

\newcommand{\R}{\mathbb{R}}

%% file: tables/numbers.tex
\newcommand{\NmanyMax}{8.3}
\newcommand{\NregTasks}{40}
\newcommand{\NregFewWins}{35}
\newcommand{\NregFarComet}{6}
\newcommand{\NregFarRofe}{3}
\newcommand{\NregCurated}{18}
\newcommand{\NregImg}{20}
\newcommand{\NregTxt}{20}
\newcommand{\NregPoolMin}{360}
\newcommand{\NregPoolMax}{10{,}000}
\newcommand{\NtbMmJointLo}{2.1}
\newcommand{\NtbMmJointHi}{3.7}
\newcommand{\NtbMmImgLo}{3.7}
\newcommand{\NtbMmImgHi}{9.6}
\newcommand{\NtbCbLo}{1.6}
\newcommand{\NtbCbHi}{5.2}
\newcommand{\NcbRefit}{0.0 to +0.8}
\newcommand{\NcbRefitRofe}{+1.5 to +5.1}
\newcommand{\NmmValTab}{0.1 to 1.6}
\newcommand{\NmmValImg}{1.2 to 2.5}
\newcommand{\NmmValRofeTab}{3.5 to 4.5}
\newcommand{\NmmValRofeImg}{5.9 to 7.0}
\newcommand{\NmmBestRofeTab}{1.7 to 2.5}
\newcommand{\NmmBestRofeImg}{4.9 to 6.1}
\newcommand{\NimgTasks}{24}
\newcommand{\NtextTasks}{20}
\newcommand{\NimgTenWins}{23}
\newcommand{\NcometProbeTenGap}{2.1}
\newcommand{\NscaleRotated}{+1.5 to +2.2}
\newcommand{\NfadeWithin}{+2.2}
\newcommand{\NtruncHelps}{+1.4 to +2.2}
\newcommand{\NtruncImg}{+0.5 to +0.7}
\newcommand{\NtruncHelpsJoint}{+1.5 to +2.1}
\newcommand{\NrofeImgMany}{+1.9 to +2.8}
\newcommand{\NrofeJointMany}{+1.2 to +1.6}
\newcommand{\NruleImg}{+0.5 to +3.6}
\newcommand{\NtruncText}{+0.3 to +1.2}
\newcommand{\NtruncExcCI}{$-$1.14 to $-$0.01}
\newcommand{\NwidthICLImg}{+0.2 to +0.7}
\newcommand{\NwidthICLFull}{+1.6 to +2.4}
\newcommand{\NwidthPFNsmall}{+0.1 to +0.8}
\newcommand{\NwidthPFNlarge}{+1.0 to +2.7}
\newcommand{\NwidthTextSmall}{+0.1 to +0.4}
\newcommand{\NwidthTextFull}{+1.0 to +2.0}
\newcommand{\NsrcWiden}{0.3}
\newcommand{\NsrcSignTol}{0.1}
\newcommand{\NtruncFourImg}{+0.3 to +1.0}
\newcommand{\NtruncFourJoint}{0.0 to +0.5}
\newcommand{\NtruncFourBeat}{0.2 to 0.7}
\newcommand{\NwidthUncut}{9}
\newcommand{\NtpfnScaleCol}{$-$0.6 to +0.5}
\newcommand{\NtpfnScaleRot}{+0.9 to +1.6}
\newcommand{\NwidthJoint}{0.1}
\newcommand{\NkOne}{$-$0.4 to $-$0.3}
\newcommand{\NkMore}{+0.1}
\newcommand{\NspecCVBefore}{1.24}
\newcommand{\NspecCVAfter}{0.22}
\newcommand{\NspecAllSets}{2,880}
\newcommand{\NspecAllTasks}{24}
\newcommand{\NspecAllRatio}{0.99}
\newcommand{\NepsSmall}{176}
\newcommand{\NepsOther}{0.016}
\newcommand{\NepsRot}{0.16}
\newcommand{\NkOneJoint}{$-$0.4 to $-$0.2}
\newcommand{\NkOneText}{$-$0.6 to $-$0.3}
\newcommand{\NpfnOneRot}{+0.7 to +1.7}
\newcommand{\NpfnFourRot}{+1.5 to +2.8}
\newcommand{\NallJoint}{$-$0.5 to $-$0.2}
\newcommand{\NzhangSame}{+0.5 to +1.6}
\newcommand{\NzhangExcCI}{$-$0.39 to $-$0.03}
\newcommand{\NzhangTextSame}{+0.7 to +1.7}
\newcommand{\NzhangTextOracle}{$-$0.2 to +0.2}
\newcommand{\NzhangTextExcCI}{$-$0.83 to $-$0.03}
\newcommand{\NpoolTextTen}{$-$0.8 to $-$0.3}
\newcommand{\NhadOne}{0.4}
\newcommand{\NcometBench}{+0.4 to +2.9}
\newcommand{\NrawRot}{+0.4 to +0.7}
\newcommand{\NrawSupport}{+0.8 to +1.6}
\newcommand{\NoursRaw}{+0.5 to +1.7}
\newcommand{\NoursRawJoint}{+2.5 to +4.2}
\newcommand{\NrawSupportJoint}{$-$2.7 to $-$1.3}
\newcommand{\NrandFrame}{$-$2.2 to $-$0.3}
\newcommand{\NgridPc}{45}
\newcommand{\NgridRot}{2.8}
\newcommand{\NgridSetsPerTask}{30}
\newcommand{\NpalImg}{+1.1 to +3.5}
\newcommand{\NpalJoint}{+0.2 to +1.5}
\newcommand{\NpoolWideImg}{0.4}
\newcommand{\NpoolWideJoint}{1.3}
\newcommand{\NpoolRotGap}{1.6 to 2.7}

%% file: style_macros.tex
\newcommand{\Method}{RoFE\xspace}

\definecolor{tableheader}{HTML}{F2F2F2}
\definecolor{tablesection}{HTML}{F5F5F5}
\providecolor{themec}{HTML}{8C3A4B}
\colorlet{tablebest}{themec!30}
\colorlet{tablesecond}{themec!12}
\definecolor{ourrow}{HTML}{E8E8FF}
\definecolor{casegood}{RGB}{24,135,84}
\definecolor{casebad}{RGB}{195,55,65}
\providecommand{\cmark}{\textcolor{casegood}{\ding{51}}}%
\providecommand{\xmark}{\textcolor{casebad}{\ding{55}}}

\newcommand{\bestbox}{{\setlength{\fboxsep}{0pt}\colorbox{tablebest}{\kern2pt Best\kern2pt}}}
\newcommand{\secondbox}{{\setlength{\fboxsep}{0pt}\colorbox{tablesecond}{\kern2pt second-best\kern2pt}}}
\newcommand{\tablestyle}{\setlength{\aboverulesep}{0pt}\setlength{\belowrulesep}{0pt}\renewcommand{\arraystretch}{1.16}}

\usepackage{aliascnt}
\newtheorem{proposition}{Proposition}[section]
\newaliascnt{assumption}{proposition}\aliascntresetthe{assumption}
\newaliascnt{definition}{proposition}\aliascntresetthe{definition}
\newaliascnt{lemma}{proposition}\aliascntresetthe{lemma}
\ifdefined\crefname
  \crefname{assumption}{Assumption}{Assumptions}\crefname{definition}{Definition}{Definitions}
  \crefname{lemma}{Lemma}{Lemmas}
\fi

%% file: figures/pgfplots_style.tex
\definecolor{tabblue}{HTML}{1F77B4}
\definecolor{taborange}{HTML}{FF7F0E}
\definecolor{tabgreen}{HTML}{2CA02C}
\definecolor{tabred}{HTML}{D62728}
\definecolor{tabpurple}{HTML}{9467BD}
\definecolor{tabbrown}{HTML}{8C564B}
\definecolor{tabpink}{HTML}{E377C2}
\definecolor{tabgrey}{HTML}{7F7F7F}
\definecolor{tabolive}{HTML}{BCBD22}
\definecolor{tabcyan}{HTML}{17BECF}
\pgfplotsset{
  paperplot/.style={
    scale only axis,
    grid=major, grid style={line width=0.2pt, draw=gray!22}, xminorticks=false, yminorticks=false,
    tick label style={font=\footnotesize}, label style={font=\footnotesize},
    title style={font=\footnotesize, yshift=-0.6ex},
    tick align=outside, tick pos=left, major tick length=2pt,
    every axis plot/.append style={line width=0.9pt},
    error bars/error bar style={line width=0.5pt, solid},
    error bars/error mark options={rotate=90, mark size=1.3pt, line width=0.5pt},
  },
  paperlegend/.style={legend style={font=\footnotesize, draw=none, column sep=0.6em, legend columns=-1},
    legend image post style={line width=0.9pt}},
  witherr/.style={error bars/.cd, y dir=both, y explicit},
}

%% file: chapters_arxiv/0_abstract_arxiv.tex
Multimodal tabular learning combines tabular data with images and text, often in settings where labeled data are scarce. Tabular foundation models such as prior-data fitted networks (PFNs) offer a promising solution via in-context learning without task-specific training. However, modern image and text encoders produce embeddings with hundreds or thousands of dimensions, far exceeding the number of features encountered during PFN pretraining.
Existing approaches bridge this gap using principal component analysis (PCA) to compress frozen embeddings.
Surprisingly, we find that this PCA interface can underperform simple linear classifiers when labeled examples are scarce, even though PFNs are designed for small datasets. We identify the underlying failure mechanism: the PFN's per-column standardization removes the variances of the PCA scores. When the PCA keeps every direction of the labeled examples, the examples collapse into a regular simplex in which all pairwise distances are equal. To address this problem, we propose \underline{Ro}tated \underline{F}rozen \underline{E}mbeddings (\textbf{RoFE}), a training-free method that rotates PCA scores before PFN inference. We theoretically show that rotation makes the column variances nearly equal, so that standardization approximately preserves the embedding geometry. Across image and text classification tasks, encoders, and tabular foundation models, RoFE improves low-data performance over the PCA interface and outperforms baselines that require task-specific training. The improvements also extend to regression.

%% file: chapters_arxiv/1_intro_arxiv.tex
\section{Introduction}
\label{sec:intro}

Multimodal tabular learning aims to predict outcomes by jointly analyzing structured tabular features and unstructured data, such as images and text~\citep{multabench2026,kim2026multimodalpfn}. Applications span healthcare and e-commerce, where patient records may be combined with medical images~\citep{pacheco2020padufes}, or product attributes with text descriptions~\citep{shi2021benchmarking}. A key challenge is effectively integrating these heterogeneous sources of information when labeled data are scarce, which is a common practical constraint in domains such as healthcare ~\citep{jaeger2014two,marcus2010open}.

Existing methods typically use pretrained encoders to transform images and text into high-dimensional embeddings, which are then combined with tabular features through early or late fusion~\citep{huang2020fusion,tang2024autogluon}. These methods generally require training a prediction head or fusion model on each task, which can be challenging with limited labels. Recently, prior-data fitted networks (PFNs), such as TabPFN and TabICL, have emerged as a promising alternative~\citep{muller2022transformers,hollmann2023tabpfn,hollmann2025tabpfn,qu2025tabicl}. PFNs are Transformers pretrained on synthetic prediction tasks that predict labels from a set of labeled examples provided as context, without task-specific training. Their strong performance on small tabular datasets~\citep{erickson2025tabarena} has motivated growing interest in extending them to multimodal learning. Recent methods include CoMET~\citep{bergstrom2026comet}, which combines frozen image and text embeddings with PFNs without additional training, and MultiModalPFN~\citep{kim2026multimodalpfn}, which learns modality projections and fine-tunes the PFN for each dataset.

A key challenge in applying PFNs to multimodal learning is the dimensionality mismatch between pretrained embeddings and tabular data: popular image encoders, such as  DINOv3~\citep{simeoni2026dinov3} and SigLIP~\citep{zhai2023siglip}, produce 768- and 1,152-dimensional embeddings, whereas TabICLv2 was pretrained on tables with at most 100 features~\citep{qu2026tabiclv2}. To bridge this gap, CoMET~\citep{bergstrom2026comet} uses principal component analysis (PCA) to reduce the dimensionality of frozen embeddings before concatenating them with tabular features and passing them to a PFN. We refer to this approach as the \textit{PCA interface}. While this provides a simple, training-free solution, we observe that it performs surprisingly poorly when labeled examples are scarce, which is precisely the setting where PFNs are expected to excel (\cref{tab:main-image}). This unexpected result raises a fundamental question:
\vspace{-5pt}
\begin{center}
\emph{Why does a PFN \textbf{fall behind a simple linear classifier} on frozen embeddings in low-data settings,\\
and can this limitation be overcome \textbf{without additional training}?}
\end{center}
\vspace{-5pt}

\begin{figure}[t]
  \centering
  \figslot{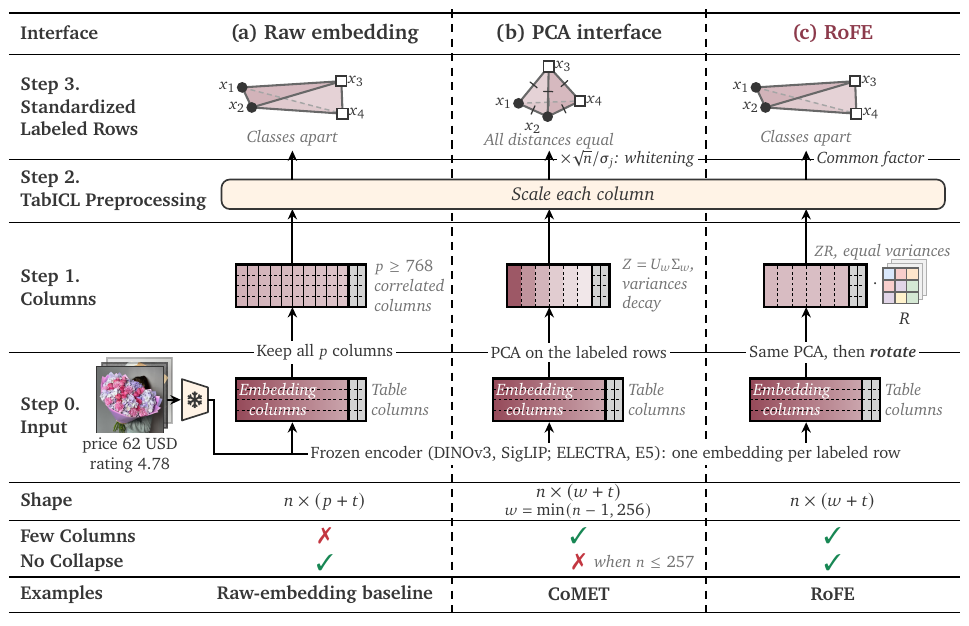}{\linewidth}
  \caption{\textbf{Three approaches to incorporating embeddings into PFNs.} (a) Raw embeddings exceed the 100-feature maximum of TabICLv2 pretraining~\citep{qu2026tabiclv2}. (b) The PCA interface reduces dimensionality but collapses the geometry of labeled examples after standardization (\cref{prop:collapse}). (c) RoFE rotates PCA scores so that standardization approximately preserves the original geometry and class separation. The $t$ tabular features remain unchanged. Step~0 shows labeled examples from the floral-bouquet task in MulTaBench~\citep{multabench2026} (Kaggle, Apache 2.0).}
  \label{fig:overview}
\end{figure}

We identify a previously overlooked interaction between PCA and the per-column standardization used by PFNs. PCA produces components with different variances, reflecting the variance spectrum of the embeddings. However, PFNs standardize each component on its own. This removes the differences in variance and scales up the directions of low variance. Surprisingly, when PCA retains all $n-1$ directions from $n$ labeled examples, this operation causes all labeled examples to become equally distant from one another, regardless of their classes (\cref{sec:collapse}).

To address this problem, we propose \underline{Ro}tated \underline{F}rozen \underline{E}mbeddings (\textbf{RoFE}), a simple, training-free transformation that randomly rotates the PCA scores before they enter the PFN (\cref{fig:overview}). The rotation keeps the distances between rows and makes the column variances nearly equal. The standardization then scales every column by nearly the same factor and approximately preserves the embedding geometry (\cref{prop:geometry}). Without the rotation, replacing the standardization with one global scale keeps the component variances but gains little (\cref{tab:scaling}). The rotation also moves these variances into the correlations between columns, which the PFN reads. RoFE requires no additional training or modifications to the pretrained encoders, PFNs, or original tabular features.

Our contributions are as follows:
\begin{Itemize}
    \item We identify and mathematically characterize a failure mode of PCA-based multimodal learning with PFNs, showing how per-column standardization can destroy embedding geometry in low-data settings (\cref{prop:invariance,prop:collapse}).
    \item We introduce RoFE and theoretically demonstrate how random rotations redistribute variance across components to preserve geometric relationships through standardization (\cref{prop:geometry,prop:haar,prop:share}).
    \item Experiments on \NimgTasks{} image and \NtextTasks{} text classification tasks, with four encoders and four PFNs, show that RoFE consistently improves the PCA interface in low-data settings. It also outperforms baselines that require task-specific training. The gain carries over to regression tasks.
\end{Itemize}

%% file: chapters_arxiv/2_related_arxiv.tex
\section{Related Work}
\label{sec:related}

\paragraph{Multimodal Learning with Tables.}
Most methods for images or text with tables fuse the modalities at the input, inside a jointly trained network, or at
the predictions~\citep{baltrusaitis2019multimodal,huang2020fusion,holste2021end}. These methods include networks that
condition the image features on the table~\citep{polsterl2021combining,pacheco2021attention}, multimodal AutoML that
fine-tunes pretrained encoders next to tabular models~\citep{shi2021benchmarking,tang2024autogluon}, and joint
pretraining of the image and the table~\citep{hager2023best,du2024tip}. MulTaBench~\citep{multabench2026} shows that fine-tuning the encoder on each task's full training set improves PFNs. All of them train on each dataset, while RoFE needs no training and targets tasks with
few labels.

\paragraph{Few-Shot Classification on Frozen Embeddings.}
Few-shot classification often keeps a pretrained encoder frozen and fits a light head on the embeddings of the few
labeled examples, a linear classifier~\citep{chen2019closer,radford2021learning} or the nearest class
mean~\citep{wang2019simpleshot}. A linear head on a good embedding outperforms methods that meta-learn across
tasks~\citep{tian2020rethinking}. These heads are trained on each task and use only the embedding, whereas a PFN needs no gradient updates on the task and can also use the table columns.

\paragraph{PFNs on Frozen Embeddings.}
PFNs~\citep{muller2022transformers} are pretrained to approximate the posterior predictive distribution under a prior
over synthetic datasets. TabPFN~\citep{hollmann2023tabpfn,hollmann2025tabpfn} and TabICL~\citep{qu2025tabicl} draw
these datasets from structural causal models, and their later versions handle larger
tables~\citep{grinsztajn2025tabpfn25,qu2026tabiclv2}. CoMET~\citep{bergstrom2026comet} passes frozen image and text embeddings to TabICL through a PCA fitted on the
labeled rows, the interface studied here. \citet{zhang2026tabpfn} pass them to TabPFN~\citep{hollmann2025tabpfn} after
standardizing each dimension and keeping a few principal components, and \citet{tajjar2026text} train a small adapter
in place of the PCA. ConTextTab~\citep{spinaci2025contexttab} embeds each cell, text included, inside an in-context
learner trained on real tables.

\paragraph{Random Rotations for Axis-Aligned Learners.}
Learners that follow the coordinate axes can need far fewer samples than rotation-invariant ones when few columns
matter~\citep{ng2004feature}, which favors trees on tables of measured variables~\citep{grinsztajn2022tree}. Rotation
Forest~\citep{rodriguez2006rotation} and random rotation ensembles~\citep{blaser2016random} rotate the features before
training decision trees and average over rotations, and random-projection ensembles~\citep{cannings2017random} combine
classifiers trained on random projections. \Method also averages over random rotations, but its learner is a
pretrained PFN that is never trained on the task, and the rotation spreads the variance of the principal component scores almost evenly over the columns before the PFN standardizes each column.

%% file: chapters_arxiv/2_background_arxiv.tex
\section{Background}
\label{sec:background}

\subsection{Preliminaries: PFNs and the PCA Interface}
\label{sec:setup}

Each row carries a frozen embedding $x\in\R^p$ of an image or a text, produced by a pretrained encoder such as
DINOv3 or ELECTRA, and possibly a few table columns. We observe $n$ labeled rows $(x_i,y_i)_{i=1}^n$ with
$y_i\in\{1,\dots,C\}$ and predict the label of a query row $x$. A prior-data fitted network (PFN) such as
TabICL~\citep{qu2025tabicl,qu2026tabiclv2} does this in one forward pass: it reads the labeled rows and their labels
as context and returns class probabilities for the query, with no gradient step. The labeled rows are few, ten to a
hundred per class in most of our experiments, while $p$ is 768 or more.

CoMET~\citep{bergstrom2026comet} passes the embedding to the PFN through a principal component analysis fitted on
the labeled rows. Let $\mu$ be the mean of the labeled embeddings, $X\in\R^{n\times p}$
the centered labeled embeddings with rows $(x_i-\mu)^\top$, and $X=U\Sigma V^\top$ its compact singular value
decomposition with $\sigma_1\ge\sigma_2\ge\cdots$. We take $X$ of rank $\min(n-1,p)$, which holds for embeddings in
general position. Keeping $w=\min(n-1,256)$ components, CoMET's default, every row is replaced by its scores
\begin{equation}
  z(x)=V_w^\top(x-\mu)\in\R^{w},\quad Z=XV_w=U_w\Sigma_w\in\R^{n\times w},
  \label{eq:scores}
\end{equation}
where $Z$ stacks the scores of the labeled rows. The columns of $Z$ have mean zero on the labeled rows, and column
$j$ has variance $\lambda_j=\sigma_j^2/n$ on them. With table columns, the scores are appended to the table columns, and with several embeddings each gets its own PCA.

\subsection{Per-Column Standardization}
\label{sec:standardize}

TabICL's preprocessing starts by standardizing each column with the mean and the standard deviation of the labeled rows. Every later step acts on the standardized values, including a power transform in half of the ensemble members and a clipping of outliers. Pretraining also standardizes every column of every synthetic
table~\citep{qu2026tabiclv2}. The model is therefore trained to receive standardized columns.
This step decides which information reaches the model. The proofs of all statements are in \cref{app:proofs}.

\begin{proposition}[Column-scale invariance]
\label{prop:invariance}
Let $D$ be a positive diagonal matrix and $b$ a vector. If every row, labeled or query, is mapped as
$x\mapsto Dx+b$, the input to TabICL's network, and hence its prediction, is unchanged.
\end{proposition}

Information carried by the scale of a column therefore never reaches the PFN. What reaches it is
each value relative to its column's spread and how the columns vary together.

\begin{figure}[t]
  \centering
  \figslot{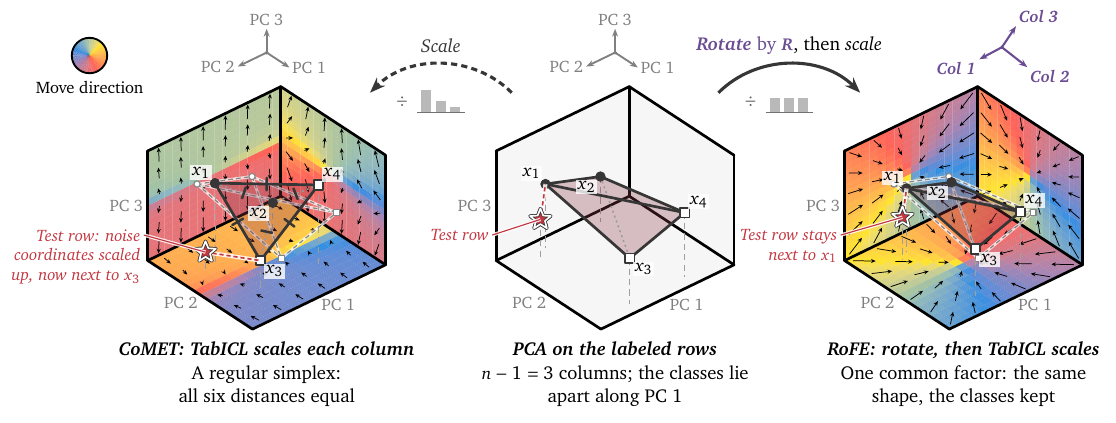}{\linewidth}
  \caption{\textbf{Geometric collapse under per-column standardization and its prevention by RoFE.} Four labeled examples from two classes (black circles and white squares) are represented by $n-1=3$ principal components (middle). \textbf{Left:} TabICL standardizes each component independently, making all pairwise distances equal (\cref{prop:collapse}). \textbf{Right:} RoFE first rotates the PCA scores by $R$ (purple) to equalize column variances, allowing subsequent standardization to preserve class separation. Dashed shapes indicate the geometry before standardization, bars represent column standard deviations, and colored walls indicate the directions of scaling (color wheel).}
  \label{fig:geometry}
\end{figure}

\subsection{Collapse of the Labeled Rows}
\label{sec:collapse}

PCA writes the spectrum into exactly these scales. Column $j$ of $Z=U_w\Sigma_w$ is $\sigma_j u_j$, a unit direction
times its singular value. Hence standardization divides it by $\sigma_j/\sqrt n$ and leaves $\sqrt n\,U_w$, which no
longer depends on $\Sigma$: standardizing principal component scores is PCA
whitening~\citep{kessy2018optimal}. When the PCA keeps all $n-1$ directions, the result has a closed form.

\begin{proposition}[Collapse of the labeled rows]
\label{prop:collapse}
Let $n\le 257$, so that $w=n-1$, let $X$ have rank $n-1$, and write $\tilde Z$ for the standardized scores of the labeled rows. Then
$\tilde Z\tilde Z^\top=n\bigl(I_n-\tfrac1n\one\one^\top\bigr)$: the labeled rows form a regular simplex, every pair
at distance $\sqrt{2n}$, whatever the embeddings and the labels.
\end{proposition}

\begin{figure}[t]
  \centering
  \figslot{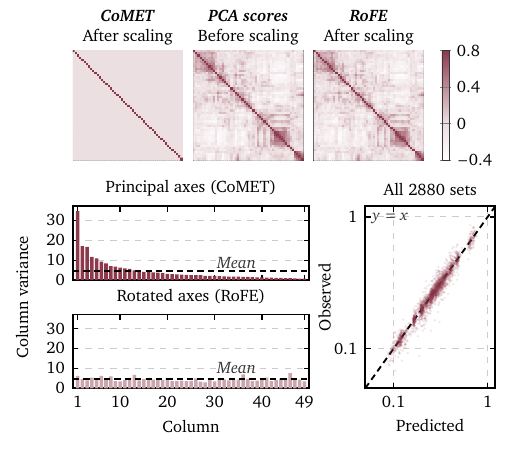}{0.6\linewidth}
  \caption{\textbf{Effect of rotation on embedding geometry and column variances.} MultiModalPFN PetFinder, DINOv3, 10 labeled examples per class ($n=50$). \textbf{Top:} pairwise cosine similarities of the labeled examples. CoMET makes every off-diagonal similarity $-1/(n-1)$, whereas RoFE preserves the geometric structure. \textbf{Bottom left:} column variances; the coefficient of variation falls from \NspecCVBefore{} to \NspecCVAfter{}, and the mean (dashed) stays. \textbf{Bottom right:} this coefficient after one rotation on every labeled set of the image tasks, against the prediction of \cref{prop:haar}.}
  \label{fig:gram}\label{fig:spectrum}
\end{figure}

Previous studies have shown \cref{prop:collapse} in other forms. \citet{pires2019high} prove it for the generalized Mahalanobis
distance, \citet{wadia2021whitening} show that whitening on the training rows erases the information in their Gram
matrix once the dimension reaches their number, and \citet{hall2005geometric} describe the same tendency for raw data
of high dimension and small sample size. In the interface above, the whitening is done by the PFN's own
standardization. The collapse therefore happens inside the model whenever the PCA is fitted on the labeled rows and keeps
$n-1$ directions (\cref{fig:geometry}, left). On real embeddings the standardized rows lose all their structure:
in \cref{fig:gram} (top), every pair of the 50 labeled rows of one task has the same cosine similarity.

The query rows are standardized with the same means and deviations, and what they see of the labeled rows also has
a closed form.

\begin{proposition}[View from a query row]
\label{prop:query}
Under the conditions of \cref{prop:collapse}, the standardized scores $\tilde z(x)$ of a query row satisfy
$\tilde Z\,\tilde z(x)=n\,a(x)$ with $a(x)=(X^{+})^\top(x-\mu)$, the minimum-norm least-squares coefficients of $x-\mu$ on the centered labeled
rows.
\end{proposition}

In $a(x)=U\Sigma^{-1}V^\top(x-\mu)$, direction $j$ is weighted by $1/\sigma_j$. Hence the
directions of smallest variance, which carry mostly noise when $n$ is small~\citep{johnstone2009consistency}, weigh the most. The distances from the query to the labeled rows, whose standardized scores $\tilde z_i$ are the rows of $\tilde Z$,
\begin{equation}
  \|\tilde z(x)-\tilde z_i\|^2=\|\tilde z(x)\|^2+(n-1)-2n\,a_i(x)
  \label{eq:query-distance}
\end{equation}
are ordered by the coefficients $a_i(x)$, which these noise directions dominate. For $n>257$ the collapse is no longer exact, yet every column still receives the same weight after standardization
(\cref{prop:share}).

%% file: chapters_arxiv/3_method_arxiv.tex
\section{Method}
\label{sec:method}

\subsection{Random Rotations}
\label{sec:rotation}

We draw $R\in O(w)$ from the Haar measure, by a QR decomposition of a Gaussian matrix with the signs of the diagonal of
the triangular factor fixed~\citep{mezzadri2007generate}. Here a rotation is any $R\in O(w)$, reflections included. The PFN receives the rotated scores $R^\top z(x)$ of every
row, labeled or query, with the same $R$. The labeled rows thus become $ZR$. We draw $K=4$
independent rotations, run the PFN on each, and average the predicted probabilities. With table columns, the scores of each embedding are rotated by their own
matrices and the table columns stay as they are. \Cref{alg:rofe} in \cref{app:details} lists the steps.

The rotated columns keep mean zero on the labeled rows, and column $j$ has variance
\begin{equation}
  d_j(R)=\tfrac1n\,\|Zr_j\|^2=\sum_{i=1}^{w}\lambda_i R_{ij}^2,
  \label{eq:column-variance}
\end{equation}
where $r_j$ is column $j$ of $R$. The weights $R_{ij}^2$ sum to one over $i$, so each $d_j$ is a weighted average of the $\lambda_i$. Under the principal axes, $R=I$, each $d_j$ is a single eigenvalue, while a rotation that mixes all directions spreads the weight of every column over all of them.

\subsection{Equal Column Variances}
\label{sec:why}

Two groups of maps act on the columns. The standardization makes the PFN invariant to positive diagonal scalings of
the columns (\cref{prop:invariance}), and orthogonal maps keep every inner product between rows. The principal scores
$Z=U_w\Sigma_w$ carry the spectrum in a diagonal factor, which is exactly what the standardization removes. A rotation
keeps the inner products and moves the spectrum from the diagonal factor into the correlations between columns
(\cref{fig:cov}).

Equal column variances turn the PFN's standardization into a single global scale. We state this for any $R\in\R^{w\times
m}$ with orthonormal rows, $m\ge w$, which covers square rotations and transforms applied to scores padded with zero columns. Here $d_j$ is defined by \cref{eq:column-variance} for $j=1,\dots,m$, and $\bar d=\frac1m\sum_{i=1}^{w}\lambda_i$ is
their mean.

\begin{proposition}[Geometry under equal column variances]
\label{prop:geometry}
Let all $d_j$ be positive, and let $\delta(R)=\max_j|\bar d/d_j-1|$. For any two rows with scores $a$ and $b$, labeled or query, the standardized
rotated scores $\tilde a$ and $\tilde b$ satisfy
\[
  \bigl|\langle\tilde a,\tilde b\rangle-\langle a,b\rangle/\bar d\bigr|\le\delta(R)\,\|a\|\,\|b\|/\bar d .
\]
\end{proposition}

Under the principal axes, $\delta(I)$ is at least $\bar d/\lambda_w-1$. Hence, when the spectrum
decays, the bound says nothing, as \cref{prop:collapse} requires. When all $d_j$ are equal, $\delta=0$
and the PFN receives the scores up to one global factor, with all distances kept. A normalized
Hadamard matrix of order $m\ge w$, applied to the scores padded with $m-w$ zero columns, has entries $\pm m^{-1/2}$. Hence
$d_j=\bar d$ for every column and $\delta=0$ exactly. A random rotation comes close.

\begin{proposition}[Column variances under a random rotation]
\label{prop:haar}
Let $R$ be Haar-distributed on $O(w)$, $\bar\lambda=\frac1w\sum_i\lambda_i$ and
$s_\lambda^2=\frac1w\sum_i(\lambda_i-\bar\lambda)^2$. Every column variance satisfies
\[
  \mathbb{E}\,d_j=\bar\lambda,\qquad \operatorname{Var} d_j=\frac{2\,s_\lambda^2}{w+2}.
\]
\end{proposition}

In relative terms, the coefficient of variation of each $d_j$ is $\sqrt{2/(w+2)}$
times that of the spectrum. Averaging the predictions over $K$
rotations reduces the effect of any single draw. \Cref{fig:spectrum} (bottom) shows the column variances of the same labeled rows
before and after a rotation. On all \NspecAllSets{} labeled sets of our \NspecAllTasks{} image tasks the observed
coefficient of variation after one rotation is \NspecAllRatio{} times the predicted one at the median (\cref{fig:spectrum}, bottom right; every task in
\cref{fig:spectrum-grid}).

We call the principal directions of large variance strong and those of small variance weak, and now show why the
rotation has to mix the two. Split the principal
directions into a set $A$ and its complement $B$, let $U_B$ hold the columns of $U_w$ indexed by $B$, and let $G(R)$ be
the Gram matrix of the standardized rotated labeled rows. The share of $G(R)$ that lies in the directions $B$ is
$\tr(U_B^\top G(R)U_B)/\tr G(R)$.

\begin{proposition}[Share of the weak directions]
\label{prop:share}
Let all $d_j$ be positive, and let $A$ and $B$ split the principal directions as above. If $R$ does not mix $A$ with $B$, that is
$R=\diag(R_A,R_B)$ with $R_A$ and $R_B$ orthogonal, the share of the directions $B$ is $|B|/w$, whatever the spectrum.
If instead all $d_j$ are equal, the share of the directions $B$ is $\sum_{i\in B}\lambda_i/\sum_{i}\lambda_i$.
\end{proposition}

Neither case needs $w=n-1$. Under the principal axes, and under any rotation that
stays within the strong directions or within the weak ones, the weak directions keep a share of the labeled rows'
geometry equal to their share of the columns. If 40 of 49 directions are weak, they take 82\% of it. A rotation that
equalizes the variances gives them their share of the variance.

%% file: chapters_arxiv/4_exp_arxiv.tex
\begin{table}[t]
  \centering
  \caption{Image tasks (left) and text tasks (right). Mean over the \NimgTasks{} image and the \NtextTasks{} text tasks. All:
  the whole labeled pool (\cref{tab:task-summary}). Bold: best in each block. $\Delta$: RoFE $-$ CoMET, with 95\% intervals over tasks in
  \cref{fig:labels}.}
  \label{tab:main-image}\label{tab:main-text}
  \footnotesize\setlength{\tabcolsep}{4pt}
  \input{tables/main_side}
\end{table}

\section{Experiments}
\label{sec:exp}

\paragraph{Tasks.}
We evaluate RoFE on \NimgTasks{} image and \NtextTasks{} text classification tasks (\cref{tab:task-summary}; full list in \cref{app:details}). The image tasks include 20 datasets released with MulTaBench (12 in the benchmark and 8 accompanying datasets~\citep{multabench2026}) and the 4 tasks from
MultiModalPFN~\citep{kim2026multimodalpfn}. The text tasks consist of the 20 datasets of the
MulTaBench release, including 10 in MulTaBench itself. For tasks with more than 10 classes, we keep 10 random classes per
fold, following CoMET~\citep{bergstrom2026comet}.

\paragraph{Protocol.}
Each task is evaluated using five folds with fixed query sets. We adopt benchmark-provided splits when available. For each fold, we sample 10, 25, 50, 100, 250, and 500 labeled examples per class, with three independent draws per sample size. We also evaluate using the full labeled training pool. Performance is measured by AUROC for binary classification and macro-averaged one-versus-rest AUROC for multiclass tasks. Results are averaged across draws and folds within each task, then across tasks. We report 95\% bootstrap confidence intervals for performance differences across tasks and the number of tasks on which RoFE outperforms CoMET.

\paragraph{Encoders and PFNs.}
We extract image embeddings using DINOv3 ViT-B/16 (class token)~\citep{simeoni2026dinov3} and SigLIP so400m (pooled output)~\citep{zhai2023siglip}. For text, we concatenate fields in ``name: value'' format and extract mean-pooled embeddings using ELECTRA-base~\citep{clark2020electra} and E5-base-v2~\citep{wang2022e5}. The primary PFN is TabICLv2~\citep{qu2026tabiclv2} with its default eight-member ensemble. We also evaluate TabPFN v2~\citep{hollmann2025tabpfn}, TabPFN 2.5~\citep{grinsztajn2025tabpfn25}, and TabPFN 3.5~\citep{jager2026tabpfn35} (\cref{app:details}).

\paragraph{Baselines.}
Our primary baseline is CoMET's PCA interface, which retains $\min(n-1,256)$ principal components fitted on the labeled examples and uses the same PFN as RoFE. We re-implement this interface following CoMET~\citep{bergstrom2026comet} (\cref{app:comet}). We also compare against a linear probe using logistic regression on standardized embeddings, with one-hot-encoded tabular features and regularization selected by cross-validation on the labeled examples. Task-specific trained baselines include MultiModalPFN~\citep{kim2026multimodalpfn} and CatBoost~\citep{prokhorenkova2018catboost} (\cref{app:trained}). Additional baselines and benchmarks are provided in \cref{app:more-results}.

\subsection{Main Results}
\label{sec:main-results}

\begin{wraptable}{r}{0.37\textwidth}
  \centering
  \caption{Performance gains from rotation across \NimgTasks{} image tasks and \NtextTasks{} text tasks, using embeddings alone. $\Delta$ denotes the AUROC difference (RoFE $-$ CoMET) with the same PFN. Results for all encoders, inputs with tabular features, and 95\% confidence intervals are provided in \cref{tab:pfns-full}.}
  \label{tab:pfns}
  \footnotesize\setlength{\tabcolsep}{4pt}
  \input{tables/pfns_short}
\end{wraptable}
\paragraph{Does RoFE Improve the PCA Interface on Image Tasks?}
\Cref{tab:main-image} (left half) compares RoFE with CoMET and a linear probe. Across 10--100 labeled examples per class, RoFE consistently outperforms CoMET with both image encoders, using either image embeddings alone or combined with tabular features. With DINOv3 image embeddings alone and 10 labels per class, RoFE outperforms CoMET on \NimgTenWins{} of \NimgTasks{} tasks. The improvements diminish as the number of labeled examples increases (\cref{sec:fade}).

\paragraph{Does the Gain Hold on Text Tasks?}
RoFE also consistently outperforms CoMET across 10--100 labels per class with both text encoders, using text embeddings alone or combined with tabular features (\cref{tab:main-image}, right half). These results demonstrate that the benefits extend across image and text modalities and different pretrained encoders (\cref{app:encoders}).

\paragraph{Does the Gain Carry Over to Other PFNs?}
\label{sec:pfns}
\Cref{tab:pfns} evaluates RoFE with different PFNs. RoFE improves TabPFN v2, TabPFN 2.5, and TabPFN 3.5 across 10--100 labels per class, with all encoders and both embedding-only and multimodal inputs (\cref{tab:pfns-full}). This suggests that the improvement is not specific to TabICLv2.

\paragraph{Is RoFE Competitive with Trained Models?}
Among the evaluated prediction heads, the linear probe is the strongest head (\cref{app:heads}). With image embeddings alone and 10 labels per class, CoMET underperforms the linear probe by \NcometProbeTenGap{} AUROC points, whereas RoFE achieves comparable performance. RoFE outperforms the linear probe from 25 labels per class with image-only inputs and across all label budgets when tabular features are included. Despite requiring no task-specific training, RoFE also outperforms MultiModalPFN and CatBoost (\cref{app:trained}).

\paragraph{Is the PCA Needed?}
We compare RoFE with directly feeding raw embeddings into the PFN without PCA (\cref{fig:overview}). With DINOv3, RoFE outperforms this baseline across 10--100 labels per class, with and without tabular features (\cref{app:need}). This demonstrates that PCA remains beneficial when combined with the proposed rotation.

\subsection{Mechanism}
\label{sec:mechanism}
\begin{figure}[t]
  \centering
  \figslot{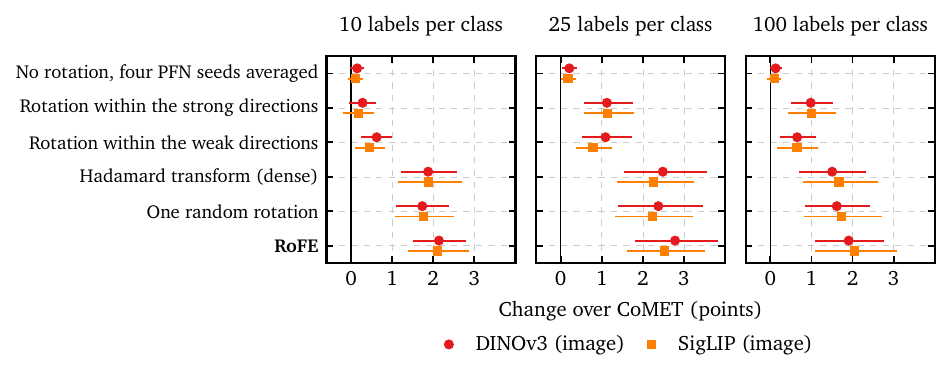}{\linewidth}
  \caption{\textbf{Effect of different changes of basis.} AUROC gains over CoMET on \NimgTasks{} image classification tasks using image embeddings alone. Error bars indicate 95\% confidence intervals across tasks. RoFE and the no-rotation baseline average four runs; all other methods use a single transformation.}
  \label{fig:basis}
\end{figure}

\paragraph{Role of Dense Rotation.}
\Cref{fig:basis} compares different transformations of the same PCA scores. Averaging multiple PFN seeds without rotation provides little improvement. Rotating only within the strong or the weak directions also yields limited gains, particularly at 10 labels per class, consistent with \cref{prop:share}. In contrast, a dense Hadamard transform performs comparably to random rotation, with further gains from averaging four rotations. Sparse Givens rotations, despite equalizing column variances, perform worse than dense rotations (\cref{app:givens}). Furthermore, replacing the PCA span with a random orthonormal basis reduces performance (\cref{app:need}). These results suggest that RoFE benefits from densely mixing principal directions while preserving the span of the labeled embeddings (\cref{app:basis}).

\paragraph{Role of Standardization.}
\Cref{tab:scaling} compares per-column standardization with global scaling, which preserves the original PCA variance spectrum. Without rotation, global scaling changes performance by less than one AUROC point on image tasks and up to about one point on text tasks, suggesting that TabICL makes limited use of variance information in individual column scales. Even under identical global scaling (hence identical distances), rotated scores outperform unrotated PCA scores by \NscaleRotated{} AUROC points on image-only tasks. These findings suggest that rotation improves how the variance spectrum is represented, beyond merely preventing its removal by standardization.

\begin{wraptable}[31]{r}{0.46\textwidth}
  \centering
  \caption{\textbf{Per-column standardization and global scaling.} Mean AUROC across \NimgTasks{} image and \NtextTasks{} text tasks using embeddings alone and TabICL without its power transform. ``Rotated'' indicates one random rotation; ``Standardized'' indicates TabICL's per-column standardization. Otherwise, a single global scaling factor normalizes the mean column variance of the labeled examples to one.}
  \label{tab:scaling}
  \footnotesize\setlength{\tabcolsep}{4pt}
  \input{tables/scaling}
\end{wraptable}
\paragraph{Does a PCA Fitted on More Data Help?}
\label{sec:where-pca}
Because PCA does not require labels, it can be fitted on additional unlabeled examples. Fitting PCA on the full training pool yields comparable performance to RoFE when the scores are rotated, but without rotation it underperforms by \NpoolRotGap{} AUROC points on image-only tasks. Similarly, fitting PCA on 100{,}000 ImageNet~\citep{russakovsky2015imagenet} embeddings does not improve over RoFE (\cref{app:pca-source}). Thus, increasing the data used to fit PCA does not eliminate the benefit of rotation.

\begin{figure}[t]
  \centering
  \figslot{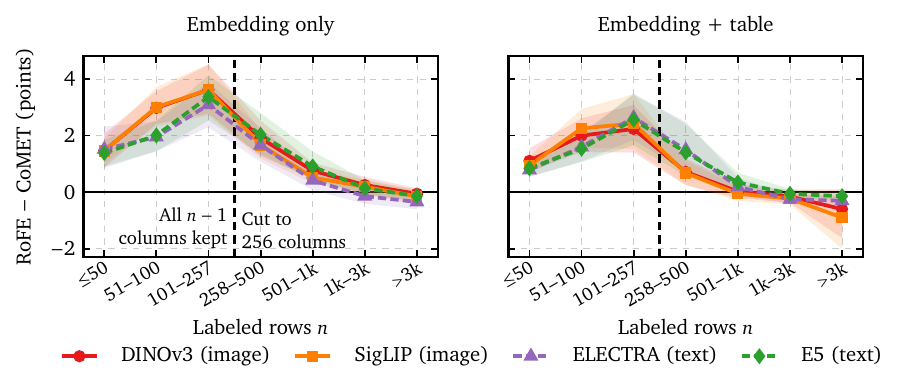}{\linewidth}
  \caption{\textbf{Performance gains by number of labeled examples.} AUROC difference (RoFE $-$ CoMET) across \NimgTasks{} image and \NtextTasks{} text classification tasks, grouped by the total number of labeled examples ($n$). Each task contributes its average gain within each bin. Shaded bands indicate 95\% confidence intervals across tasks.}
  \label{fig:fade}
\end{figure}

\paragraph{When Does the Gain Fade?}
\label{sec:fade}
As shown in \cref{fig:fade}, RoFE achieves its largest gains when $n\le257$, where PCA retains all $n-1$ directions and the labeled examples undergo geometric collapse. The gains diminish as $n$ increases and approach zero for $n>1000$, where PCA retains 256 directions. Within the same task, settings that retain all $n-1$ directions show an additional \NfadeWithin{}-point improvement over those with truncated PCA representations. Similar trends are observed on text tasks.

\paragraph{Would Fewer PCA Columns Suffice?}
Reducing the number of components provides another way to mitigate the PCA interface's limitations by discarding low-variance directions. This improves CoMET from 25 labels per class onward. RoFE still outperforms CoMET with its best PCA dimensionality on image-only tasks, while the two perform comparably when tabular features are included. Rotation also improves performance with fewer PCA components, although the gains are smaller (\cref{app:fewer-columns}).

\subsection{Other Settings}
\label{sec:other-settings}

\paragraph{Does RoFE Work with More Than Ten Classes?}
\label{sec:many-classes}
The main experiments limit tasks with more than 10 classes to 10 randomly selected classes per fold. In this experiment, we retain all classes in Hearthstone (11), wine reviews (30), and Spotify genres (114). With image or text embeddings alone, RoFE outperforms CoMET on all three tasks at 10 and 25 labels per class, with gains of up to \NmanyMax{} AUROC points. At 100 or more labels per class, the methods perform within one point of each other (\cref{app:many-classes}).

\paragraph{What About Regression?}
\label{sec:regression}
The geometric collapse in \cref{prop:collapse} also applies to regression. We evaluate RoFE using TabICLv2's regressor on \NregTasks{} regression tasks from MulTaBench. With 10--250 labeled examples, RoFE has a higher $R^2$ than CoMET on at least \NregFewWins{} tasks at each sample size, across all encoders, with and without tabular features. The improvements diminish as the labeled dataset grows, and the two methods perform within one point of each other with the full training pool (\cref{app:regression}).

%% file: tables/main_side.tex
\tablestyle
\begin{NiceTabular}{lccccccc@{\hspace{10pt}}ccccccc}
\CodeBefore
  \rowcolor{tablesection}{4,9,14,19}
  \rowcolor{ourrow}{7,12,17,22}
\Body
\toprule
\Block{3-1}{\textbf{Method}} & \Block{1-7}{\textbf{Image tasks} (\NimgTasks{})} & & & & & & & \Block{1-7}{\textbf{Text tasks} (\NtextTasks{})} & & & & & & \\
\cmidrule(lr){2-8}\cmidrule(lr){9-15}
 & \Block{1-7}{\textbf{AUROC} ($\uparrow$) by labeled rows per class} & & & & & & & \Block{1-7}{\textbf{AUROC} ($\uparrow$) by labeled rows per class} & & & & & & \\
 & \textbf{10} & \textbf{25} & \textbf{50} & \textbf{100} & \textbf{250} & \textbf{500} & \textbf{All} & \textbf{10} & \textbf{25} & \textbf{50} & \textbf{100} & \textbf{250} & \textbf{500} & \textbf{All} \\
\midrule
 & \Block[l]{1-7}{\textit{Image only, DINOv3}} & & & & & & & \Block[l]{1-7}{\textit{Text only, ELECTRA}} & & & & & & \\
Linear probe & \textbf{70.1} & 74.0 & 76.3 & 78.2 & 81.0 & 81.6 & 81.9 & 64.2 & 68.1 & 71.0 & 73.5 & 75.7 & 77.0 & 79.2 \\
CoMET & 68.0 & 72.0 & 75.2 & 78.3 & 82.0 & 83.3 & \textbf{84.3} & 62.8 & 66.7 & 69.7 & 73.3 & 76.5 & \textbf{78.7} & \textbf{81.3} \\
\textbf{RoFE} & \textbf{70.1} & \textbf{74.8} & \textbf{77.9} & \textbf{80.2} & \textbf{82.8} & \textbf{83.8} & \textbf{84.3} & \textbf{64.5} & \textbf{68.9} & \textbf{72.2} & \textbf{74.8} & \textbf{77.1} & 78.6 & 81.1 \\
$\Delta$ & +2.1 & +2.8 & +2.7 & +1.9 & +0.8 & +0.5 & 0.0 & +1.8 & +2.3 & +2.5 & +1.5 & +0.6 & 0.0 & $-$0.2 \\
\midrule
 & \Block[l]{1-7}{\textit{Image + table, DINOv3}} & & & & & & & \Block[l]{1-7}{\textit{Text + table, ELECTRA}} & & & & & & \\
Linear probe & 71.2 & 75.3 & 77.8 & 80.0 & 83.1 & 84.1 & 84.6 & 65.5 & 70.0 & 73.4 & 76.3 & 79.1 & 80.7 & 83.2 \\
CoMET & 73.9 & 78.2 & 80.9 & 83.5 & 86.8 & \textbf{88.0} & \textbf{89.0} & 71.5 & 75.0 & 77.3 & 80.7 & 83.7 & \textbf{85.1} & \textbf{86.8} \\
\textbf{RoFE} & \textbf{75.2} & \textbf{79.7} & \textbf{82.4} & \textbf{84.7} & \textbf{87.1} & 87.8 & 88.6 & \textbf{72.4} & \textbf{76.8} & \textbf{79.8} & \textbf{81.8} & \textbf{84.0} & \textbf{85.1} & 86.6 \\
$\Delta$ & +1.3 & +1.6 & +1.5 & +1.2 & +0.2 & $-$0.1 & $-$0.4 & +0.9 & +1.8 & +2.5 & +1.2 & +0.3 & $-$0.1 & $-$0.2 \\
\midrule
 & \Block[l]{1-7}{\textit{Image only, SigLIP}} & & & & & & & \Block[l]{1-7}{\textit{Text only, E5}} & & & & & & \\
Linear probe & 72.9 & 76.7 & 79.2 & 80.9 & 83.7 & 84.4 & 84.7 & \textbf{69.3} & 72.1 & 74.2 & 75.9 & 77.5 & 78.7 & 81.0 \\
CoMET & 71.0 & 75.1 & 78.3 & 80.9 & 84.7 & 86.1 & \textbf{86.6} & 67.4 & 70.5 & 73.0 & 75.7 & 79.0 & 81.0 & \textbf{83.5} \\
\textbf{RoFE} & \textbf{73.1} & \textbf{77.6} & \textbf{80.8} & \textbf{82.9} & \textbf{85.5} & \textbf{86.4} & \textbf{86.6} & 68.9 & \textbf{72.7} & \textbf{75.8} & \textbf{78.0} & \textbf{79.8} & \textbf{81.3} & 83.4 \\
$\Delta$ & +2.1 & +2.5 & +2.5 & +2.0 & +0.8 & +0.4 & 0.0 & +1.5 & +2.2 & +2.7 & +2.3 & +0.8 & +0.3 & 0.0 \\
\midrule
 & \Block[l]{1-7}{\textit{Image + table, SigLIP}} & & & & & & & \Block[l]{1-7}{\textit{Text + table, E5}} & & & & & & \\
Linear probe & 73.5 & 77.5 & 80.2 & 82.2 & 85.1 & 86.1 & 86.4 & 70.3 & 73.3 & 75.8 & 77.8 & 80.0 & 81.6 & 84.1 \\
CoMET & 75.7 & 80.0 & 82.7 & 84.8 & 88.2 & \textbf{89.3} & \textbf{90.0} & 74.2 & 77.4 & 79.5 & 82.1 & 84.9 & 86.4 & \textbf{88.0} \\
\textbf{RoFE} & \textbf{77.0} & \textbf{81.5} & \textbf{84.2} & \textbf{86.2} & \textbf{88.5} & 89.1 & 89.5 & \textbf{75.2} & \textbf{79.0} & \textbf{81.9} & \textbf{83.6} & \textbf{85.3} & \textbf{86.5} & 87.8 \\
$\Delta$ & +1.3 & +1.6 & +1.5 & +1.4 & +0.3 & $-$0.2 & $-$0.5 & +1.0 & +1.6 & +2.4 & +1.5 & +0.3 & +0.1 & $-$0.2 \\
\bottomrule
\end{NiceTabular}

%% file: tables/pfns_short.tex
\tablestyle
\begin{NiceTabular}{lcccc}
\CodeBefore
  \rowcolor{tablesection}{3,8}
\Body
\toprule
\Block{2-1}{\textbf{PFN}} & \Block{1-4}{\textbf{$\Delta$ AUROC}\\ by labeled rows per class} & & & \\
 & \textbf{10} & \textbf{25} & \textbf{50} & \textbf{100} \\
\midrule
\Block[l]{1-5}{\textit{Image only, DINOv3}} & & & & \\
TabICLv2 & +2.1 & +2.8 & +2.7 & +1.9 \\
TabPFN v2 & +2.8 & +2.4 & +2.0 & +1.5 \\
TabPFN 2.5 & +2.6 & +2.1 & +1.7 & +0.9 \\
TabPFN 3.5 & +2.4 & +2.3 & +1.8 & +1.4 \\
\midrule
\Block[l]{1-5}{\textit{Text only, ELECTRA}} & & & & \\
TabICLv2 & +1.8 & +2.3 & +2.5 & +1.5 \\
TabPFN v2 & +2.4 & +3.0 & +2.6 & +1.8 \\
TabPFN 2.5 & +2.4 & +2.8 & +1.9 & +1.1 \\
TabPFN 3.5 & +1.9 & +2.8 & +2.0 & +1.3 \\
\bottomrule
\end{NiceTabular}

%% file: tables/scaling.tex
\tablestyle
\begin{NiceTabular}{cccccc}
\CodeBefore
  \rowcolor{tablesection}{3,8,13,18}
\Body
\toprule
\Block{2-1}{\textbf{Rotated}} & \Block{2-1}{\textbf{Standardized}} & \Block{1-4}{\textbf{AUROC} ($\uparrow$)\\ by labeled rows per class} & & & \\
 &  & \textbf{10} & \textbf{25} & \textbf{50} & \textbf{100} \\
\midrule
\Block[l]{1-6}{\textit{Image only, DINOv3}} & & & & & \\
-- & -- & 68.3 & 72.8 & 75.4 & 78.0 \\
-- & \ding{51} & 68.0 & 72.2 & 75.3 & 78.4 \\
\ding{51} & -- & \textbf{69.8} & \textbf{74.5} & \textbf{77.7} & \textbf{80.0} \\
\ding{51} & \ding{51} & \textbf{69.8} & 74.4 & 77.6 & \textbf{80.0} \\
\midrule
\Block[l]{1-6}{\textit{Image only, SigLIP}} & & & & & \\
-- & -- & 71.2 & 75.5 & 78.4 & 80.8 \\
-- & \ding{51} & 71.1 & 75.2 & 78.4 & 81.0 \\
\ding{51} & -- & \textbf{72.8} & \textbf{77.3} & \textbf{80.5} & \textbf{82.7} \\
\ding{51} & \ding{51} & \textbf{72.8} & \textbf{77.3} & \textbf{80.5} & \textbf{82.7} \\
\midrule
\Block[l]{1-6}{\textit{Text only, ELECTRA}} & & & & & \\
-- & -- & 62.6 & 66.5 & 69.4 & 72.3 \\
-- & \ding{51} & 62.8 & 66.8 & 69.9 & 73.4 \\
\ding{51} & -- & 64.0 & \textbf{68.6} & 71.8 & \textbf{74.4} \\
\ding{51} & \ding{51} & \textbf{64.2} & \textbf{68.6} & \textbf{71.9} & \textbf{74.4} \\
\midrule
\Block[l]{1-6}{\textit{Text only, E5}} & & & & & \\
-- & -- & 67.7 & 71.5 & 74.2 & 76.5 \\
-- & \ding{51} & 67.4 & 70.6 & 73.2 & 75.8 \\
\ding{51} & -- & \textbf{68.4} & \textbf{72.5} & \textbf{75.5} & \textbf{77.8} \\
\ding{51} & \ding{51} & 68.3 & 72.4 & \textbf{75.5} & \textbf{77.8} \\
\bottomrule
\end{NiceTabular}

%% file: chapters_arxiv/5_conclusion_arxiv.tex
\section{Conclusion and Limitation}
\label{sec:conclusion}

In this study, we investigate why PFNs can underperform simple linear classifiers on frozen embeddings when labeled data are scarce. We show that per-column standardization removes the variances of the principal component scores and collapses labeled examples into a regular simplex when all $n-1$ principal directions are retained. To address this issue, we propose RoFE, which rotates the principal component scores at random before they reach the PFN. The
rotation moves the spectrum into the column correlations that the PFN reads and makes the standardization
divide every column by nearly the same factor. Across image and text classification tasks, four encoders and four PFNs,
RoFE consistently improves on the PCA interface when labeled rows are few. Without any training, it also outperforms baselines that require task-specific training, and the
gain carries over to regression tasks.

This work has several limitations. First, RoFE's gain is largest with limited labeled data and diminishes as the dataset grows. Second, averaging predictions over multiple rotations increases inference cost. Finally, our evaluation focuses on image and text embeddings. Future work should evaluate RoFE on additional modalities.

%% file: chapters_arxiv/9_appendix_arxiv.tex
\section{Proofs}
\label{app:proofs}

Throughout, standardization uses the mean and the standard deviation with divisor $n$ of the labeled rows, as
TabICL does, and we ignore the constant $10^{-6}$ that TabICL adds to each standard deviation. In the labeled sets of
the image tasks with 10 to 100 labels per class, every principal column has a standard deviation of at least
\NepsOther{}, except in \NepsSmall{} sets, all from tasks whose data contain byte-identical images. Such sets can repeat
an embedding and then have rank below $n-1$, outside \cref{prop:collapse}. After one rotation no column has a standard
deviation below \NepsRot{}.

\subsection{Proof of \cref{prop:invariance}}
\label{app:invariance}

TabICL first casts numeric inputs to floating point and removes columns with a single distinct value on the labeled
rows. Write $D=\diag(c_1,\dots,c_p)$ with $c_j>0$. The map $t\mapsto c_jt+b_j$ is injective and keeps the number of distinct values in every column. Hence the same
columns are removed. Each ensemble member then fits its preprocessing pipeline on the labeled rows, and its first step
standardizes column $j$ with the labeled-row mean $m_j$ and standard deviation $s_j$. After the map, the column has
mean $c_jm_j+b_j$ and standard deviation $c_js_j$. Every standardized value, labeled or query, is thus
$\bigl((c_jt+b_j)-(c_jm_j+b_j)\bigr)/(c_js_j)=(t-m_j)/s_j$, as before. The remaining steps, clipping to $[-100,100]$,
the power transform, the removal of outliers, the feature shuffles and the class permutations, are fitted on or applied
to these standardized values or depend only on the random state. The network receives only their output.\qed

\subsection{Proof of \cref{prop:collapse,prop:query}}
\label{app:collapse}

Column $j$ of $Z=U_w\Sigma_w$ is $\sigma_ju_j$. Since $\one^\top X=0$, we have $\one^\top U=0$, so the column has mean
zero and variance $\sigma_j^2/n$ on the labeled rows, and standardization gives $\tilde Z=\sqrt n\,U_w$. For $w=n-1$
and $\operatorname{rank}X=n-1$, $U_w=U$ has $n-1$ orthonormal columns orthogonal to $\one$. Hence $UU^\top$ is the projector onto $\one^\perp$:
\[
  \tilde Z\tilde Z^\top=n\,UU^\top=n\bigl(I_n-\tfrac1n\one\one^\top\bigr).
\]
This gives \cref{prop:collapse}. The diagonal entries are $n-1$ and the off-diagonal ones
$-1$, so $\|\tilde z_i-\tilde z_k\|^2=2(n-1)+2=2n$.

For \cref{prop:query}, a query's standardized scores are $\tilde z(x)=\sqrt n\,\Sigma^{-1}V^\top(x-\mu)$. Hence, with
$X^+=V\Sigma^{-1}U^\top$,
\[
  \tilde Z\tilde z(x)=n\,U\Sigma^{-1}V^\top(x-\mu)=n\,(X^+)^\top(x-\mu).
\]
Since
$(X^+)^\top=(X^\top)^+$, $a(x)$ is the minimum-norm least-squares solution of $X^\top a=x-\mu$.
\Cref{eq:query-distance} follows from $\|\tilde z_i\|^2=n-1$.\qed

\subsection{Proof of \cref{prop:geometry}}
\label{app:geometry}

Because $\one^\top Z=0$, the rotated labeled scores $ZR$ have column means zero and column variances $d_j$. Hence the
standardized rotated scores of a row with scores $a$ are $\tilde a=D^{-1/2}R^\top a$ with $D=\diag(d_1,\dots,d_m)$.
Since $RR^\top=I_w$,
\begin{equation*}
  \langle\tilde a,\tilde b\rangle-\langle a,b\rangle/\bar d=a^\top R\bigl(D^{-1}-\bar d^{-1}I_m\bigr)R^\top b ,
\end{equation*}
and $\|R^\top a\|=\|a\|$, so its absolute value is at most
$\|a\|\,\|b\|\max_j|1/d_j-1/\bar d|=\|a\|\,\|b\|\,\delta(R)/\bar d$.\qed

\subsection{Proof of \cref{prop:haar}}
\label{app:haar}

The Haar measure is invariant under left multiplication by orthogonal matrices. Hence each column $u=r_j$ is uniformly
distributed on the unit sphere of $\R^w$. Write $u=g/\|g\|$ with $g\sim N(0,I_w)$. Since $u$ is independent of
$\|g\|$, $\mathbb{E}g_i^2g_k^2=\mathbb{E}\|g\|^4\,\mathbb{E}u_i^2u_k^2$ for all $i,k$, and with $\sum_iu_i^2=1$,
$\mathbb{E}\|g\|^4=w(w+2)$, $\mathbb{E}g_i^4=3$ and $\mathbb{E}g_i^2g_k^2=1$ for $i\ne k$,
\begin{equation*}
  \mathbb{E}u_i^2=\frac1w,\qquad \mathbb{E}u_i^4=\frac{3}{w(w+2)},\qquad \mathbb{E}u_i^2u_k^2=\frac{1}{w(w+2)}.
\end{equation*}
Hence $\mathbb{E}d_j=\sum_i\lambda_i/w=\bar\lambda$, $\operatorname{Var}u_i^2=\frac{2(w-1)}{w^2(w+2)}$,
$\operatorname{Cov}(u_i^2,u_k^2)=-\frac{2}{w^2(w+2)}$, and
\begin{equation*}
  \begin{aligned}
  \operatorname{Var}d_j&=\frac{2(w-1)\sum_i\lambda_i^2-2\sum_{i\ne k}\lambda_i\lambda_k}{w^2(w+2)}\\
  &=\frac{2\bigl(w\sum_i\lambda_i^2-(\sum_i\lambda_i)^2\bigr)}{w^2(w+2)}=\frac{2\,s_\lambda^2}{w+2},
  \end{aligned}
\end{equation*}
since $w\sum_i\lambda_i^2-(\sum_i\lambda_i)^2=w^2s_\lambda^2$.\qed

\subsection{Proof of \cref{prop:share}}
\label{app:share}

With $D$ as in \cref{app:geometry}, $G(R)=ZRD^{-1}R^\top Z^\top$ and $Z=U_w\Sigma_w$, so
\[
  \begin{gathered}
  U_B^\top G(R)U_B=\Sigma_B\,[RD^{-1}R^\top]_{BB}\,\Sigma_B,\\
  \tr G(R)=\sum_{j=1}^m d_j^{-1}r_j^\top\Sigma_w^2r_j=\sum_{j=1}^m d_j^{-1}\,n\,d_j=nm,
  \end{gathered}
\]
where $\Sigma_B$ holds the singular values indexed by $B$.

If $R$ does not mix $A$ with $B$, then $m=w$ and $[RD^{-1}R^\top]_{BB}=R_BD_B^{-1}R_B^\top$. For $j\in B$, column $\rho_j$ of $R_B$ gives $d_j=\frac1n\rho_j^\top\Sigma_B^2\rho_j$, and hence
\[
  \tr\bigl(U_B^\top G(R)U_B\bigr)=\sum_{j\in B}d_j^{-1}\rho_j^\top\Sigma_B^2\rho_j=n|B|,
\]
and the share is $|B|/w$.

If all $d_j$ are equal, then $D=\bar d\,I_m$, and with $R_{B\cdot}\in\R^{|B|\times m}$ the rows of $R$ indexed by $B$, so that $R_{B\cdot}R_{B\cdot}^\top=I_{|B|}$,

\[
  \begin{gathered}
  \tr\bigl(U_B^\top G(R)U_B\bigr)=\tr\Sigma_B^2/\bar d=n\sum_{i\in B}\lambda_i/\bar d,\\
  \tr G(R)=n\sum_i\lambda_i/\bar d.
  \end{gathered}
\]
\qed

\section{Column Variances under Orthogonal Maps}
\label{app:givens}

\begin{figure}[ht]
  \centering
  \figslot{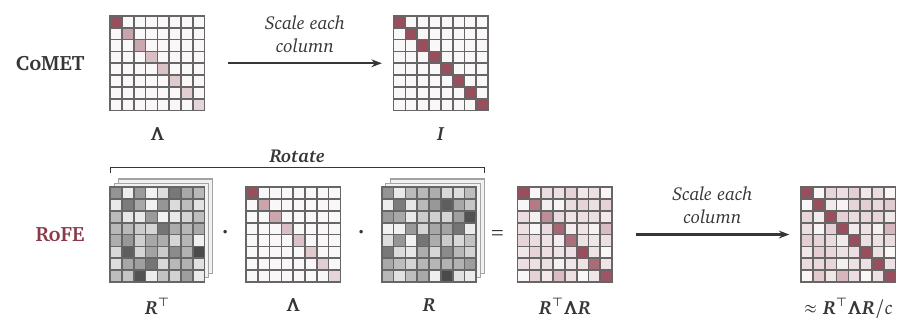}{\linewidth}
  \caption{\textbf{Where the spectrum sits.} Leading $8\times8$ block of the column covariance of the labeled rows of
  \cref{fig:spectrum}; shade: absolute value relative to the largest entry of the block; gray: absolute entries of $R$.
  The per-column scaling removes the diagonal $\boldsymbol\Lambda$; after $R$ the diagonal is nearly constant, the
  scaling divides it by about one common factor $c$, the mean column variance, and the spectrum stays in the
  correlations.}
  \label{fig:cov}
\end{figure}

\begin{figure}[ht]
  \centering
  \figslot{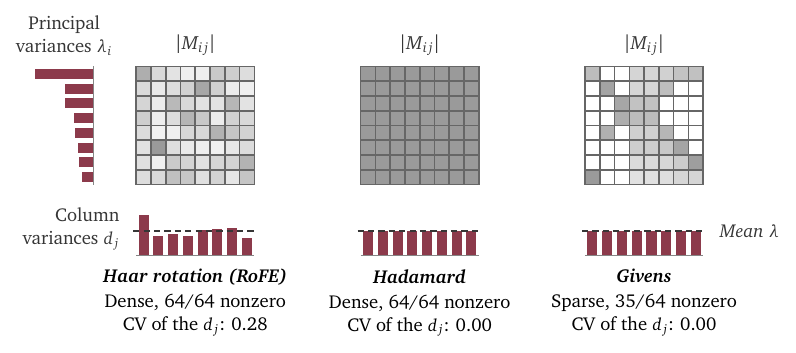}{0.85\linewidth}
  \caption{\textbf{Three orthogonal maps on one spectrum.} Left: the first eight principal variances $\lambda_i$ of the labeled
  rows of \cref{fig:spectrum}. Grids: absolute entries of each map; bars: the column variances $d_j$ after it, with
  their mean as a dashed line. Hadamard and Givens give equal variances and Haar nearly equal ones, but Givens leaves
  many entries zero.}
  \label{fig:mix}
\end{figure}

\Cref{fig:cov} shows the column covariance of the labeled rows of \cref{fig:spectrum} before and after RoFE's first
rotation, and \cref{fig:mix} compares three orthogonal maps on the same spectrum.

A product of Givens rotations, each mixing two columns, can also make all column variances equal. Such a product exists
by the Schur--Horn theorem~\citep{horn1954doubly}, and we build it column pair by column pair (\cref{app:basis}). It tests whether equal
variances alone carry the gain. With TabICL, Givens rotations stay within half a point of AUROC of random rotations at
ten labels per class and underperform them as the labels grow, and with TabPFN they underperform from ten labels per
class (\cref{tab:basis}). RoFE therefore uses dense rotations. One possible reason is that TabICL first embeds each
column on its own~\citep{qu2025tabicl}. A sparse mix leaves some columns with mostly signal and others with mostly
noise at the same variance, while a dense mix gives every column about the same share of each principal direction
(\cref{fig:mix}).

\section{Experimental Details}
\label{app:details}

\paragraph{Algorithm.}
\Cref{alg:rofe} lists the steps of \Method for one set of labeled rows. The rotations are drawn once for the set
and shared by every query row, and with several embeddings, lines 1 to 4 and the rotations of lines 6 and 7 are applied to each
embedding separately. In lines 8 and 9, $[a;b]$ stacks $a$ above $b$. The rotated scores of a row are thus followed by its table
columns.

\input{algorithms/rofe}

\begin{table}[!ht]
  \centering
  \caption{Tasks by source. Ranges over the tasks of a row; pool and query: rows in one fold, classes: classes per
  fold, columns: table columns.}
  \label{tab:task-summary}
  \footnotesize\setlength{\tabcolsep}{4pt}
  \input{tables/task_summary}
\end{table}

\paragraph{Image Tasks.}
\Cref{tab:tasks} lists the \NimgTasks{} image tasks. The MulTaBench tasks use the benchmark's loader and its first
five splits, and, as in MulTaBench, at most 10{,}000 training rows per fold. The 8 datasets released with MulTaBench outside its 40 datasets are marked $\dagger$. The MultiModalPFN tasks use the splits of its authors' code.

Seven tasks come from three datasets that both benchmarks use: PAD-UFES-20 (the same label and patients, different
preprocessing), PetFinder (age bin in MulTaBench, adoption speed in MultiModalPFN) and CBIS-DDSM (breast density in
MulTaBench, pathology of masses and of calcifications in MultiModalPFN). We keep each benchmark's tasks as they are.
Resampling the 20 datasets instead of the 24 tasks widens the 95\% intervals of RoFE $-$ CoMET and of RoFE $-$ the
linear probe by at most \NsrcWiden{} points and changes for none of them whether it includes zero. The 20 tasks of the
MulTaBench release, which hold each dataset once, give the same sign of the mean difference wherever it is more than
\NsrcSignTol{} points from zero.

The two CBIS-DDSM tasks of MultiModalPFN have three images per row, and their embeddings are concatenated before the
PCA. One PCA and one rotation therefore cover the three views.

The tasks come from
CelebA~\citep{liu2015celeba}, Hateful Memes~\citep{kiela2020hateful}, CMMD~\citep{cai2023cmmd},
CBIS-DDSM~\citep{lee2017cbis}, CheXpert~\citep{irvin2019chexpert}, HuBMAP~\citep{jain2023hubmap}, the ZooScan data
of \citet{grandremy2024zooscan}, OASIS~\citep{marcus2007oasis}, HAM10000~\citep{tschandl2018ham10000},
PAD-UFES-20~\citep{pacheco2020padufes}, PetFinder~\citep{petfinder2019kaggle} and the game data of MuG~\citep{lu2023mug}.
The others come from public Kaggle uploads listed by MulTaBench.

\begin{table}[!htb]%
  \centering
  \caption{The \NimgTasks{} image tasks. MulTaBench names drop the prefix MUL\_IMAGE\_ or BIN\_IMAGE\_. $\dagger$: released with MulTaBench outside its 40 datasets. Pool and query: rows in one fold. Columns: table columns.
  Images: images per row. PETFINDER and petfinder, CBIS\_DDSM, cbis\_mass and cbis\_calc, and PAD\_UFES\_LESION and
  pad\_ufes\_20 come from the same three datasets.}
  \label{tab:tasks}
  \scriptsize
  \input{tables/tasks}
\end{table}

\paragraph{Text Tasks.}
The \NtextTasks{} text tasks of \cref{tab:tasks-text} are the ten text classification tasks of MulTaBench and the ten released with it, and 12 of these 20
come from the benchmark of \citet{shi2021benchmarking}. The
text columns are those MulTaBench's loader marks as text. Each is written as ``name: value'', and they are joined
into one string with ``; '', shortest field first by median length over the task, and truncated to 512 tokens. E5 receives the prefix ``query: ''.

\begin{table}[!htbp]
  \centering
  \caption{The \NtextTasks{} text tasks. Names drop the prefix MUL\_TEXT\_ or BIN\_TEXT\_. $\dagger$: released with
  MulTaBench outside its 40 datasets. Pool and query: rows in one fold. Columns: table columns. Text fields: text
  columns joined into the embedded string.}
  \label{tab:tasks-text}
  \scriptsize
  \input{tables/tasks_text}
\end{table}

\paragraph{Classes.}
Hearthstone (11 classes), Spotify genres (114) and wine reviews (30) keep ten classes per fold, drawn uniformly at
random from the classes of the labeled pool with a fixed seed per task and fold. In every task, a class with fewer rows
in the labeled pool than the number drawn enters with all of its rows.

\paragraph{Models.}
TabICLv2 runs with its default eight ensemble members.
TabPFN 2.5 and TabPFN 3.5 run with the defaults of the
\texttt{tabpfn} package, eight members, under their non-commercial licenses, and TabPFN v2 runs with four members. The feature ``rotation'' of
the TabPFN implementation shifts the column order cyclically and does not mix the columns.

CoMET
and RoFE use the same PFN, the same members and the same random state. RoFE's rotations are drawn from fixed seeds. Every run uses a single NVIDIA A100 or L40S GPU, and the linear probe and the other
heads run on CPUs.

The linear probe standardizes the embedding on the labeled rows and fits a logistic regression
with an $L_2$ penalty (scikit-learn, at most 2000 iterations), its inverse strength chosen among 0.01, 0.1, 1 and 10 by
stratified three-fold cross-validation of the log-loss on the labeled rows. Missing table values are imputed, numbers by their
median and categories as a category of their own. Numbers are then standardized and categories one-hot encoded, with
unseen categories ignored.

\paragraph{CoMET.}
\label{app:comet}
CoMET releases no code for its pipeline, so we implement it from the description in its Sections~4 and~6.1. It takes
the class token of DINOv3 or the mean-pooled ELECTRA embedding and fits a PCA on the labeled rows with $\min(n-1,256)$
components, without whitening. TabICLv2 then runs on the scores and the table columns. The authors'
research repository, which holds their PAL (Pseudo Attention Labels) pooling, uses the same settings.

Three choices are ours and are shared with
RoFE, so that the two differ only in the interface. The text embedding averages over all tokens, the table columns reach the PFN as they are, without a table vectorizer, and TabICLv2 runs with its default eight members.
TabICLv2 then turns each categorical column into integer codes learned on the labeled rows, with one extra code for
missing and unseen values, and fills missing numbers with their mean on the labeled rows. TabPFN receives integer
codes learned on the fold's pool, with the categorical columns marked as such, and leaves out columns that have no value
on the labeled rows.

\section{Additional Results}
\label{app:more-results}

\FloatBarrier  %
\needspace{6\baselineskip}  %
\subsection{Gain for Each Encoder}
\label{app:encoders}

\begin{figure}[!ht]
  \centering
  \figslot{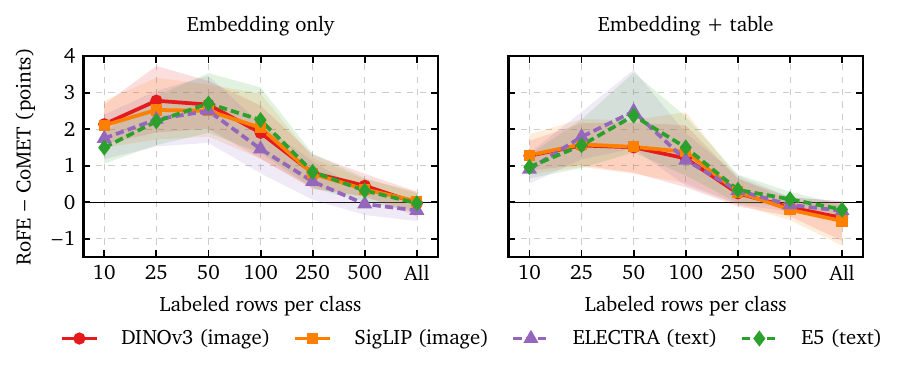}{\linewidth}
  \caption{\textbf{Gain over CoMET for four encoders.} RoFE minus CoMET in AUROC on the \NimgTasks{} image and \NtextTasks{} text
  tasks. Points: means over tasks; shaded bands: 95\% intervals over tasks.}
  \label{fig:labels}
\end{figure}

\Cref{fig:labels} puts the four encoders side by side. With and without the table, the gain over CoMET is positive
for every encoder at every number of labels from 10 to 100 per class. With the whole pool the two are comparable with
the embedding alone, and CoMET is slightly higher with the table (difference \NallJoint{} points).

\FloatBarrier  %
\needspace{6\baselineskip}  %
\subsection{Other PFNs}
\label{app:pfns-full}

\Cref{tab:pfns-full} is \cref{tab:pfns} with both encoders of each modality, with the table, and with the 95\%
intervals.

\begin{table}[!ht]
  \centering
  \caption{Gain from the rotation with the same PFN, \NimgTasks{} image and \NtextTasks{} text tasks. $\Delta$: RoFE $-$ CoMET with the same PFN, $\pm$ the half-width of its 95\% bootstrap interval over tasks.}
  \label{tab:pfns-full}
  \footnotesize\setlength{\tabcolsep}{4pt}
  \input{tables/pfns}
\end{table}

\FloatBarrier  %
\needspace{6\baselineskip}  %
\subsection{Column Variances on Every Task}
\label{app:spectrum-grid}

\begin{figure}[!ht]
  \centering
  \figslot{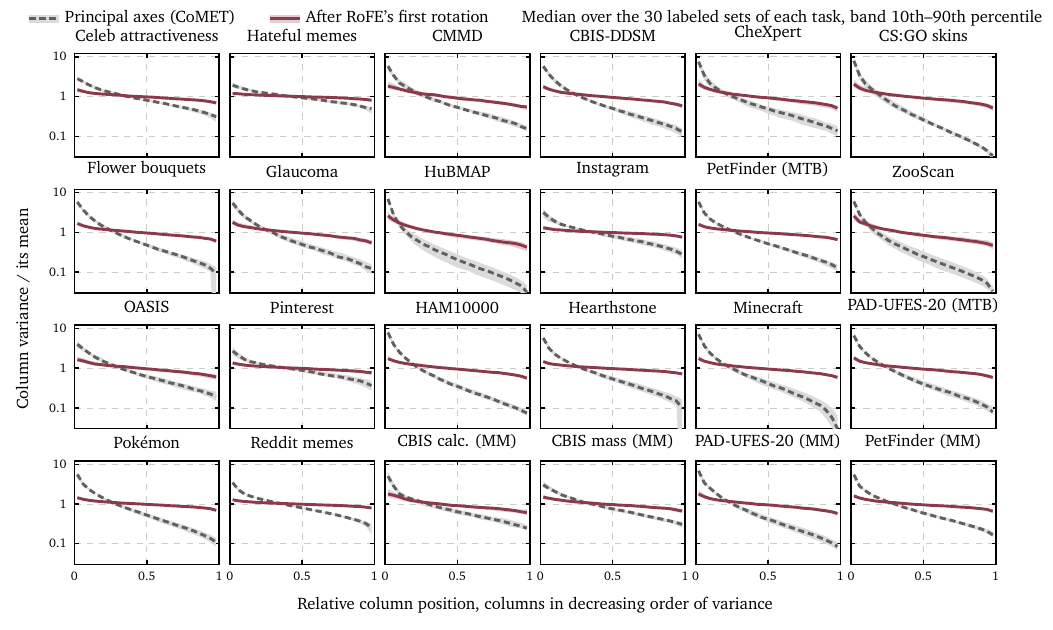}{\linewidth}
  \caption{\textbf{Column variances on every image task.} The \NgridSetsPerTask{} labeled sets of each task with ten labeled
  rows per class (DINOv3 and SigLIP, five folds, three draws), in the order of \cref{tab:tasks}; MTB: MulTaBench, MM:
  MultiModalPFN.}
  \label{fig:spectrum-grid}
\end{figure}

\Cref{fig:spectrum-grid} repeats \cref{fig:spectrum} on every image task. On the principal axes the largest column
variance is \NgridPc{} times the smallest at the median over tasks, and after one rotation it is \NgridRot{} times.

\FloatBarrier  %
\needspace{6\baselineskip}  %
\subsection{Other Heads}
\label{app:heads}

\Cref{tab:heads,tab:heads-text} add three heads with the settings of \citet{zhang2026tabpfn}, on the inputs of the linear probe
(\cref{app:details}), standardized on the labeled rows:
\begin{Itemize}
    \item five nearest neighbors with cosine distance, weighted by distance;
    \item a linear support vector machine with $C=1$ and Platt scaling, its solver stopped after $10^6$ iterations;~and
    \item a perceptron with hidden layers of 100 and 50 units, trained for at most 2000 iterations.
\end{Itemize}
On the image and the text tasks, the linear probe is the strongest of them.

\begin{table}[!ht]
  \centering
  \caption{Other heads on the image tasks. Mean over the \NimgTasks{} tasks. Bold: best in each block.}

  \label{tab:heads}
  \footnotesize\setlength{\tabcolsep}{4pt}
  \input{tables/heads}
\end{table}

\begin{table}[!ht]
  \centering
  \caption{Other heads on the text tasks. Mean over the \NtextTasks{} tasks. Bold: best in each block.}
  \label{tab:heads-text}
  \footnotesize\setlength{\tabcolsep}{4pt}
  \input{tables/heads_text}
\end{table}

\FloatBarrier
\needspace{6\baselineskip}
\subsection{Trained Baselines}
\label{app:trained}

MultiModalPFN~\citep{kim2026multimodalpfn} fine-tunes TabPFN v2 on the labeled rows together with learned projectors
for the image embedding. We run the authors' code and model with DINOv3 on the \NimgTasks{} image tasks (SigLIP's 1152
dimensions would need a change of the authors' model). Each task uses one of the authors' 39 configurations, the one
their grid selects on the whole pool by test accuracy, and with fewer labels the same configuration is fine-tuned on
the labeled rows. The runs with the image alone use the authors' code path without table
columns. CatBoost~\citep{prokhorenkova2018catboost} follows MulTaBench's
baseline~\citep{multabench2026}: 2000 iterations with early stopping on 10\% of the labeled rows, the table's
categorical columns handled natively, and each embedding reduced to 30 principal components, image embeddings
standardized first. With few labels the PCA keeps $\min(30,n_{\text{train}}-1)$ components and the validation part holds
at least one row per class. Both baselines use the labeled and query rows of the RoFE runs.
On the text tasks CatBoost is the only trained baseline. ConTextTab~\citep{spinaci2025contexttab} and
TabSTAR~\citep{arazi2025tabstar} encode the text cells inside a model of their own and lie outside the interface
studied here.

\Cref{tab:trained-image,tab:trained-text} give the results. RoFE outperforms MultiModalPFN at every number of labels,
by \NtbMmJointLo{} to \NtbMmJointHi{} points with the table and \NtbMmImgLo{} to \NtbMmImgHi{} with the image alone.
It also outperforms CatBoost with every encoder and setting, by \NtbCbLo{} to \NtbCbHi{}. Every 95\% interval lies above
zero. CatBoost underperforms CoMET throughout, and with the table MultiModalPFN is within two points of the linear probe
(\cref{tab:main-image}). Refitting CatBoost on all labeled rows, with the number of trees and the learning rate of the early-stopped fit, raises
it by \NcbRefit{} points, and RoFE still outperforms it, by \NcbRefitRofe{} points, every 95\% interval above zero.

The authors' fine-tuning keeps the checkpoint with the lowest loss on a fifth of the labeled rows. Choosing the
configuration by this loss as well (\cref{tab:mmpfn-config}, 10 to 100 labels per class) lowers MultiModalPFN by
\NmmValTab{} points with the table and raises it by \NmmValImg{} with the image alone. RoFE still outperforms it, by
\NmmValRofeTab{} and \NmmValRofeImg{} points. Even the best of the 39 configurations for each task and number of labels,
chosen by its test AUROC, underperforms RoFE, by \NmmBestRofeTab{} points with the table and \NmmBestRofeImg{} with the
image alone. In both cases every 95\% interval lies above zero.

\begin{table}[!ht]
  \centering
  \caption{Trained baselines on the image tasks. Mean over the \NimgTasks{} tasks; MultiModalPFN runs with DINOv3 only.
  Bold: best in each block. $\Delta$ X: RoFE $-$ X, with the half-width of the 95\% interval over tasks.}
  \label{tab:trained-image}
  \footnotesize\setlength{\tabcolsep}{3pt}
  \input{tables/trained_image}
\end{table}

\begin{table}[!ht]
  \centering
  \caption{Trained baselines on the text tasks. Mean over the \NtextTasks{} tasks, as in \cref{tab:trained-image}.}
  \label{tab:trained-text}
  \footnotesize\setlength{\tabcolsep}{3pt}
  \input{tables/trained_text}
\end{table}

\begin{table}[!ht]
  \centering
  \caption{MultiModalPFN with three choices of its configuration among the authors' 39, with DINOv3. Mean over the
  \NimgTasks{} image tasks. Authors: chosen on the whole pool by test accuracy, as in \cref{tab:trained-image}.
  Val.\ loss: lowest validation loss of the fine-tuning on each set of labeled rows. Best per task: highest test AUROC
  for each task and number of labels. Bold: best in each block. $\Delta$ X: RoFE $-$ X, with the half-width of the 95\%
  interval over tasks.}
  \label{tab:mmpfn-config}
  \footnotesize\setlength{\tabcolsep}{3pt}
  \input{tables/mmpfn_config}
\end{table}

\FloatBarrier  %
\needspace{6\baselineskip}  %
\subsection{PCA on Standardized Dimensions}
\label{app:zhang}

A variant of the interface standardizes each embedding dimension on the labeled rows, fits a PCA with $d$ components (here $d$ is a number of components, not a column variance $d_j$),
and passes the scores to TabPFN 2.5 or to a classical head~\citep{zhang2026tabpfn}. We follow their settings with $d$
among 8, 16, 32, 48, 64, 96 and all components, and with TabPFN 2.5, TabPFN 3.5 and six heads ($k$-nearest neighbors,
logistic regression, a linear support vector machine, a perceptron, a random forest and XGBoost).

At each number of
labels, \cref{tab:zhang} gives three choices: the $d$ with the highest mean AUROC over the tasks with TabPFN 2.5, the PFN or head and $d$
with the highest mean, and each task's highest value over all of them. The table sets them against CoMET with TabPFN 2.5 and RoFE with
three PFNs. The best $d$ with TabPFN 2.5 is 16 at ten labels per class and 48 or 64 at 100. The best PFN or head is
TabPFN 3.5 with 16 to 48 components, except logistic regression on all components with DINOv3 and the image alone at ten
labels per class.

With TabPFN 2.5 on both sides and the image alone, RoFE outperforms the best fixed $d$ at every number of labels from 10
to 100 per class with both encoders, by \NzhangSame{} points. RoFE with TabPFN 3.5 is at least as high as all three
choices, the best per task included. With the table and the same PFN, RoFE is higher at ten labels per class. From 25,
a PCA on 32 to 64 standardized dimensions is comparable to it and slightly higher in one case (TabPFN 3.5, SigLIP, 100
labels per class; 95\% interval of the difference \NzhangExcCI{}), since with the table the rotation gains less (\cref{app:fewer-columns}).

On the \NtextTasks{} text tasks (\cref{tab:zhang-text}), RoFE with TabPFN 2.5 outperforms the best fixed $d$ with the
text alone by \NzhangTextSame{} points, and the two are comparable only with E5 at 100 labels per class. RoFE with
TabPFN 3.5 is comparable to the best per task (difference \NzhangTextOracle{} points). With the table, a PCA on
standardized dimensions is comparable to RoFE already at ten labels per class and slightly higher in one case (TabPFN
2.5, E5, 100 labels per class; 95\% interval of the difference \NzhangTextExcCI{}).

\begin{table}[!ht]
  \centering
  \caption{PCA on standardized dimensions. Mean over the \NimgTasks{} image tasks. Bold: best in each block.
  $\Delta$ best fixed $d$: RoFE with TabPFN 2.5 $-$ the best fixed $d$; $\Delta$ best PFN or head: RoFE with TabPFN 3.5
  $-$ the best PFN or head and $d$; $\pm$ the half-width of the 95\% bootstrap interval over tasks.}

  \label{tab:zhang}
  \footnotesize\setlength{\tabcolsep}{4pt}
  \input{tables/zhang}
\end{table}

\begin{table}[!ht]
  \centering
  \caption{PCA on standardized dimensions on the text tasks. Mean over the \NtextTasks{} text tasks. Rows, bold and
  $\Delta$ as in \cref{tab:zhang}.}
  \label{tab:zhang-text}
  \footnotesize\setlength{\tabcolsep}{4pt}
  \input{tables/zhang_text}
\end{table}

\FloatBarrier  %
\needspace{6\baselineskip}  %
\subsection{Number of Rotations}
\label{app:kcurve}

\Cref{tab:kcurve} compares 1, 2, 4, 8 and 16 rotations with TabICLv2. One rotation underperforms four with the image
alone (DINOv3; difference \NkOne{} points) and with the table (\NkOneJoint{}). On the text tasks with the text alone,
the difference is \NkOneText{} (ELECTRA and E5, one and four rotations in the same runs). Eight or sixteen rotations add
\NkMore{} points. RoFE therefore uses four. With TabPFN v2 the four rotations carry a larger share of the gain: with the
image alone and DINOv3, one rotation gains \NpfnOneRot{} points over CoMET and four gain \NpfnFourRot{}
(\cref{tab:basis}).

\begin{table}[!ht]
  \centering
  \caption{Number of rotations, \NimgTasks{} image tasks. $\Delta$: the given number of rotations $-$ four, $\pm$ the
  half-width of its 95\% bootstrap interval over tasks.}

  \label{tab:kcurve}
  \footnotesize\setlength{\tabcolsep}{4pt}
  \input{tables/kcurve}
\end{table}

\FloatBarrier  %
\needspace{6\baselineskip}  %
\subsection{Dense Rotation and Standardization}
\label{app:basis}

\Cref{tab:basis} gives the numbers behind \cref{fig:basis}.
\begin{Itemize}
    \item The strong directions are the first $k$ principal
directions, with $k$ the Gavish--Donoho rank~\citep{gavish2014optimal} of the labeled rows, and the weak directions
are the other $w-k$.
    \item The Hadamard transform pads the scores with zero columns to $m=2^{\lceil\log_2 w\rceil}$ and
multiplies them by the normalized Sylvester--Hadamard matrix of order $m$. The PFN thus receives $m$ columns. These
span the same $w$ directions, and $m\le256$ since $w\le256$. With both PFNs the Hadamard transform is within
\NhadOne{} points of one random rotation, which keeps $w$ columns.
    \item The Givens
rotations equalize the column variances in at most $w-1$ steps: each step rotates the two columns that are not yet
fixed and have the largest and the smallest variance, so that the first reaches the mean variance.
    \item The first row runs
the PFN with four random states on the principal axes, as many runs as RoFE.
\end{Itemize}

\begin{table}[!ht]
  \centering
  \caption{Change of basis, the numbers behind \cref{fig:basis}, \NimgTasks{} image tasks. Strong and weak directions are those of \cref{prop:share}, and column variances are those of the labeled rows after the change of basis. $\Delta$: each
  change of basis $-$ CoMET with the same PFN, $\pm$ the half-width of its 95\% bootstrap interval over tasks. Bold:
  largest gain in each block. The RoFE rows of TabPFN v2 are those of \cref{tab:pfns-full}.}
  \label{tab:basis}
  \footnotesize\setlength{\tabcolsep}{4pt}
  \input{tables/basis}
\end{table}

\Cref{tab:scaling-tabpfn} repeats \cref{tab:scaling} with TabPFN~v2, whose network standardizes each column on the
labeled rows. Every ensemble member runs without TabPFN's column transforms and SVD features, which remove the column
scales themselves. With the principal axes the global scale changes the score by \NtpfnScaleCol{} points, and at the
same global scale rotated axes outperform principal axes by \NtpfnScaleRot{} points. TabPFN therefore also uses the spectrum
mainly through the correlations between columns.

\begin{table}[!ht]
  \centering
  \caption{Per-column scaling and global scaling with TabPFN~v2. Mean over the \NimgTasks{} image tasks, embedding
  alone, every member without a column transform, SVD features or fingerprint column. Rotated: one random rotation.
  Standardized: TabPFN's per-column scaling; otherwise one global scale sets the mean column variance of the labeled rows
  to one.}
  \label{tab:scaling-tabpfn}
  \footnotesize\setlength{\tabcolsep}{4pt}
  \input{tables/scaling_tabpfn}
\end{table}

\FloatBarrier  %
\needspace{6\baselineskip}  %
\subsection{Fewer PCA Columns}
\label{app:fewer-columns}

\Cref{tab:fewer-columns} gives CoMET, at each number of labels, encoder and setting, the number of columns with the
highest mean AUROC over the tasks, among 8, 16, 32, 64, $\sqrt n$ and the Gavish--Donoho
threshold~\citep{gavish2014optimal}. It also gives the two rules that need no choice, $\sqrt n$ and the Gavish--Donoho threshold,
on their own. Fewer columns help CoMET by
\NtruncHelps{} points from 25 labels per class (DINOv3, image alone). With the image alone RoFE still outperforms CoMET with its best number of columns, by \NtruncImg{}
points, and with either rule on its own, by \NruleImg{} points. With the table and from 25 labels per class, RoFE and
CoMET are comparable, whether CoMET keeps its best number of columns or follows either rule. CoMET with the
Gavish--Donoho threshold, its best choice there, is slightly higher in one case (DINOv3, 50 labels per class; 95\% interval of RoFE $-$ CoMET \NtruncExcCI{}). Fewer columns help CoMET about as much with the table (\NtruncHelpsJoint{} points), but the
rotation helps it less: RoFE $-$ CoMET is \NrofeJointMany{} points with the table, against \NrofeImgMany{} with the
image alone (\cref{tab:main-image}). With the table the two gains are therefore about equal. The text tasks follow the
same pattern (\cref{tab:fewer-columns-text}): with the text alone RoFE outperforms CoMET with its best number of
columns, by \NtruncText{} points, and with the table the two are comparable at every number of labels from 10 to 100
per class.

\Cref{tab:rotation-width} rotates fewer columns. The gain shrinks with the number of columns, since the dropped
directions are mostly weak ones and fewer remain whose share the rotation lowers (\cref{prop:share}), and with the image
alone it stays positive at every width. With TabICLv2 and the PCA cut to 32 columns, one rotation adds \NwidthICLImg{}
points, against \NwidthICLFull{} with $\min(n-1,256)$ columns (DINOv3). With TabPFN~2.5 on standardized dimensions
(\cref{app:zhang}), four rotations add \NwidthPFNsmall{} points with $d=8$ and \NwidthPFNlarge{} with $d=96$. With the
table and 32 columns, the gain is within \NwidthJoint{} of zero from 25 labels per class. With four rotations, a PCA cut
to 32 columns or to the Gavish--Donoho rank gains \NtruncFourImg{} points with the image alone and \NtruncFourJoint{} with
the table. With the table, a 32-column PCA with four rotations outperforms RoFE by \NtruncFourBeat{} points from 10 to
100 labels per class, as does a Gavish--Donoho PCA with four rotations from 25 labels per class. On the text tasks, one
rotation adds \NwidthTextSmall{} points on 32 columns and \NwidthTextFull{} on $\min(n-1,256)$ columns (ELECTRA, text
alone), and with the table and 32 columns it is again within \NwidthJoint{} of zero from 25 labels per class.

\begin{table}[!ht]
  \centering
  \caption{CoMET with fewer PCA columns, \NimgTasks{} image tasks, with its best number of columns at each number of
  labels and with the $\sqrt n$ and Gavish--Donoho rules. $\Delta$: the first minus the second, $\pm$ the half-width of
  its 95\% bootstrap interval over tasks.}

  \label{tab:fewer-columns}
  \footnotesize\setlength{\tabcolsep}{4pt}
  \input{tables/fewer_columns}
\end{table}

\begin{table}[!ht]
  \centering
  \caption{CoMET with fewer PCA columns on the \NtextTasks{} text tasks, as in \cref{tab:fewer-columns}.}
  \label{tab:fewer-columns-text}
  \footnotesize\setlength{\tabcolsep}{4pt}
  \input{tables/fewer_columns_text}
\end{table}

\begin{table}[!ht]
  \centering
  \caption{The rotation on fewer columns, \NimgTasks{} image and \NtextTasks{} text tasks. $\Delta$: rotated minus principal axes, $\pm$ the
  half-width of its 95\% bootstrap interval over tasks. TabICLv2: PCA fitted on the labeled rows, one rotation unless
  stated; at ten labels per class \NwidthUncut{} of the image tasks have at most 32 principal columns, which the
  32-column rows rotate in full.
  TabPFN~2.5: PCA on standardized dimensions with $d$ components, at most $n-1$, four rotations, image tasks only.}
  \label{tab:rotation-width}
  \footnotesize\setlength{\tabcolsep}{4pt}
  \input{tables/rotation_width}
\end{table}

\FloatBarrier  %
\needspace{6\baselineskip}  %
\subsection{Where the PCA Is Fitted}
\label{app:pca-source}

\Cref{tab:pca-source} fits the PCA on the labeled rows, as CoMET and RoFE do, on every row of the fold's pool with
the labels unused, or on 100{,}000 ImageNet training images embedded by the same encoder. Each PCA keeps CoMET's
number of columns $w=\min(n-1,256)$ and runs without and with RoFE's four rotations. On the two MultiModalPFN CBIS tasks, with
three images per row, the ImageNet PCA projects each view on its own and splits the columns between the views.

\Cref{tab:pca-source-256} keeps 256 columns instead.
With the rotation, the pool PCA is then at most \NpoolWideImg{} points above RoFE with the image alone and up to
\NpoolWideJoint{} points below it with the table.

\Cref{tab:pca-source-text,tab:pca-source-256-text} repeat the comparison on the text tasks, with the external PCA
fitted on the 48{,}000 texts of CoMET's four text classification sets (\cref{app:comet-benchmarks}), embedded by the
same encoder. The result is the same as with images, except that at ten labels per class the pool PCA with the rotation
is slightly below RoFE (difference \NpoolTextTen{} points).

\begin{table}[!ht]
  \centering
  \caption{Where the PCA is fitted, with $w=\min(n-1,256)$ columns as in CoMET. Mean over the \NimgTasks{} image tasks.
  Whole pool: every labeled row of the fold, labels unused. Bold: best in each block.}
  \label{tab:pca-source}
  \footnotesize\setlength{\tabcolsep}{4pt}
  \input{tables/pca_source}
\end{table}

\begin{table}[!ht]
  \centering
  \caption{Where the PCA is fitted, with 256 columns, against RoFE. Mean over the \NimgTasks{} image tasks.
  $w=\min(n-1,256)$. Bold: best in each block.}
  \label{tab:pca-source-256}
  \footnotesize\setlength{\tabcolsep}{4pt}
  \input{tables/pca_source_256}
\end{table}

\begin{table}[!ht]
  \centering
  \caption{Where the PCA is fitted on the text tasks, with $w=\min(n-1,256)$ columns. Mean over the \NtextTasks{} text
  tasks. CoMET's text sets: its four text classification sets, embedded by the same encoder. Bold: best in each block.}
  \label{tab:pca-source-text}
  \footnotesize\setlength{\tabcolsep}{4pt}
  \input{tables/pca_source_text}
\end{table}

\begin{table}[!ht]
  \centering
  \caption{Where the PCA is fitted on the text tasks, with 256 columns, against RoFE. Mean over the \NtextTasks{} text
  tasks. $w=\min(n-1,256)$. Bold: best in each block.}
  \label{tab:pca-source-256-text}
  \footnotesize\setlength{\tabcolsep}{4pt}
  \input{tables/pca_source_256_text}
\end{table}

\FloatBarrier  %
\needspace{6\baselineskip}  %
\subsection{Rotating Other Inputs}
\label{app:need}

\Cref{tab:need} rotates other inputs.
\begin{Itemize}\setlength{\emergencystretch}{2em}%
    \item On eight numeric binary tables of \citet{grinsztajn2022tree} (pol, house\_16H, MiniBooNE, eye\_movements, jannis,
default-of-credit-card-clients, heloc and Bioresponse), whose columns are measured variables, four rotations lower AUROC at every number of labels, in line with their finding that the orientation of
the data matters for tabular models.
    \item Without a PCA, the raw embedding outperforms CoMET by \NrawSupport{} points with
the image alone, RoFE outperforms the raw embedding by \NoursRaw{} points, and four rotations of the raw embedding add \NrawRot{}
points, the opposite of the numeric tables. With the table, the raw embedding underperforms CoMET (difference
\NrawSupportJoint{} points) and RoFE outperforms it by \NoursRawJoint{} points.
    \item A random orthonormal basis of a $w$-dimensional subspace of $\R^p$, drawn uniformly and independently of the data, in place of the span of
the labeled rows underperforms RoFE with the image alone (difference \NrandFrame{} points), most with ten labels per class. The rotation
therefore needs the span of the labeled rows.
\end{Itemize}

\begin{table}[!ht]
  \centering
  \caption{Rotating other inputs: eight numeric tables, and the \NimgTasks{} image tasks with DINOv3, image alone unless
  stated.
  $\Delta$: the first minus the second, $\pm$ the half-width of its 95\% bootstrap interval over tasks.}

  \label{tab:need}
  \footnotesize\setlength{\tabcolsep}{4pt}
  \input{tables/need}
\end{table}

\FloatBarrier  %
\needspace{6\baselineskip}  %
\subsection{CoMET's Benchmarks}
\label{app:comet-benchmarks}

\Cref{tab:comet-benchmarks} runs CoMET and RoFE on CoMET's own benchmarks: four ImageNet subsets, four text
classification sets, and eight multimodal tasks that we rebuild following CoMET's description. RoFE has the higher
accuracy on all three benchmarks at every number of labels from 10 to 100 per class, by \NcometBench{} points.

\begin{table}[!ht]
  \centering
  \caption{CoMET's benchmarks, in accuracy as CoMET reports. Labeled pools of the text and multimodal sets hold at most
  10{,}000 rows. $\Delta$: RoFE $-$ CoMET, $\pm$ the half-width of its 95\% bootstrap interval over tasks.}

  \label{tab:comet-benchmarks}
  \footnotesize\setlength{\tabcolsep}{4pt}
  \input{tables/comet_benchmarks}
\end{table}

\FloatBarrier  %
\needspace{6\baselineskip}  %
\subsection{CoMET with Patch Pooling}
\label{app:pal}

CoMET can pool the encoder's patch tokens with PAL before the PCA. \Cref{tab:pal} uses its authors' code and, at each
number of labels, the PAL variant that scores best.

With the image alone RoFE outperforms CoMET with PAL by \NpalImg{} points at 10 to 100 labels per class, and with the
table by \NpalJoint{} points.

\begin{table}[!ht]
  \centering
  \caption{CoMET with PAL pooling. Mean over the \NimgTasks{} image tasks. Bold: best in each block. $\Delta$: RoFE $-$
  (CoMET + PAL), $\pm$ the half-width of its 95\% bootstrap interval over tasks.}

  \label{tab:pal}
  \footnotesize\setlength{\tabcolsep}{4pt}
  \input{tables/pal}
\end{table}

\FloatBarrier  %
\needspace{6\baselineskip}  %
\subsection{Many Classes}
\label{app:many-classes}

Hearthstone, wine reviews and Spotify genres have 11, 30 and 114 classes, and the main results draw ten of them per fold.
\Cref{tab:many-classes} keeps every class, with the benchmark's loader, its 10{,}000-row sample and its splits as
before. Every query set holds every class of its labeled pool. Only TabICLv2 is run, in its own many-class
mode. Spotify genres has fewer than 250 labeled rows
per class in a fold, so it has no entries at 250 and 500 labels per class. With the table the gains are smaller at most levels, and with
wine reviews and ELECTRA CoMET is up to half a point higher from 25 labels per class.

\begin{table}[!ht]
  \centering
  \caption{Tasks with more than ten classes, every class kept, embedding alone unless stated. Mean over folds and draws.
  Bold: best in each block. $\Delta$: RoFE $-$ CoMET; with the table: the same with the table columns added. $\Delta$, ten classes: the same task with ten classes drawn per fold, as in the main results. --: fewer than 250 labeled rows per class.}
  \label{tab:many-classes}
  \footnotesize\setlength{\tabcolsep}{4pt}
  \input{tables/many_classes}
\end{table}

\subsection{Regression}
\label{app:regression}

The \NregTasks{} regression datasets released with MulTaBench, \NregCurated{} of them among its 40 datasets, are
\NregImg{} with an image and \NregTxt{} with text. They use the benchmark's loader, its 10{,}000-row sample and its first
five splits, as the classification tasks, which gives \NregPoolMin{} to \NregPoolMax{} labeled rows per fold. The labeled rows are drawn by their number $n$,
from 10 to 1000, in five strata of equal size by the quantiles of the target, three times per fold, and the whole pool is
used once. The PFN is TabICLv2's regressor with eight members and its mean prediction, on the same scores as for
classification. The reference is a ridge regression on the standardized embedding, with the table one-hot encoded, its
penalty chosen among $10^{-2},10^{-1},\dots,10^{4}$ by leave-one-out cross-validation on the labeled rows. $R^2$ is in percent, as in MulTaBench, averaged
over the draws and folds of a task and not clipped.

\Cref{tab:regression} gives medians over tasks, since a few tasks have $R^2$ below $-100\%$. These are
up to \NregFarComet{} tasks with CoMET and up to \NregFarRofe{} with RoFE, all at 250 labeled rows or fewer, and AIRBNB\_NYC with the ridge
regression and the table (\cref{tab:regression-tasks}). CoMET's median $R^2$ with the embedding alone
and DINOv3 or ELECTRA stays below zero up to 250 labeled rows, the levels at which the PCA keeps all $n-1$ directions,
and RoFE's is above zero from 50. With the embedding alone the ridge regression outperforms RoFE up to 50 labeled rows and
underperforms it from 250, and with the table RoFE outperforms it from 50 labeled rows. Classification shows the same
pattern, where the linear probe is comparable to RoFE at ten labels per class with the image alone
(\cref{sec:main-results}): with the fewest rows, a linear model whose penalty is tuned on the labeled rows is as strong
as the PFN or stronger. \Cref{tab:regression-source} splits the gain by
source.

\begin{table}[!ht]
  \centering
  \caption{Regression on the \NregTasks{} tasks released with MulTaBench, TabICLv2's regressor. Median $R^2$ over tasks;
  DINOv3 / ELECTRA: DINOv3 on the image tasks and ELECTRA on the text tasks. Ridge: ridge regression on the embedding.
  Bold: best in each block. $\Delta$: RoFE $-$ CoMET, mean over tasks with 95\% intervals, and its median. Tasks higher:
  tasks with $\Delta>0$, of 40 (38 at 500 and 37 at 1000 labeled rows, where the pool of the other tasks is smaller).
  All: the whole labeled pool.}
  \label{tab:regression}
  \footnotesize\setlength{\tabcolsep}{3pt}
  \input{tables/regression}
\end{table}

\begin{table}[!ht]
  \centering
  \caption{RoFE $-$ CoMET in $R^2$ by source, mean / median over tasks. Curated: among the 40 datasets of MulTaBench; other:
  released with it outside them. Number of tasks in parentheses.}
  \label{tab:regression-source}
  \scriptsize\setlength{\tabcolsep}{2.5pt}
  \input{tables/regression_source}
\end{table}

\begin{table}[!ht]
  \centering
  \caption{The \NregTasks{} regression tasks. Names drop the prefix REG\_IMAGE\_ or REG\_TEXT\_. $\dagger$: released
  with MulTaBench outside its 40 datasets. Pool: labeled rows in one fold. $R^2<-100\%$: methods whose mean $R^2$ over the
  folds and draws of the task falls below $-100\%$ at some number of labeled rows, encoder or setting.}
  \label{tab:regression-tasks}
  \scriptsize\setlength{\tabcolsep}{4pt}
  \input{tables/regression_tasks}
\end{table}

%% file: algorithms/rofe.tex
\begin{algorithm}[ht]
\caption{\Method: predictions for the query rows}
\label{alg:rofe}
\begin{algorithmic}[1]
\Require labeled embeddings $x_1,\dots,x_n\in\R^p$ with labels $y_1,\dots,y_n$; query embeddings
  $x'_1,\dots,x'_{n_q}$; table columns $t_i$, $t'_q$ (optional); a PFN $f$; the number of rotations $K$ (we use $4$)
\Ensure class probabilities $\hat p_q$ for every query row $q$
\State $\mu\gets\frac1n\sum_{i}x_i$ \Comment{\textit{labeled rows only}}
\State $X\gets[x_1-\mu,\dots,x_n-\mu]^\top$
\State $U\Sigma V^\top\gets\mathrm{SVD}(X)$;\quad $w\gets\min(n-1,256)$
\State $z_i\gets V_w^\top(x_i-\mu)$, \ $z'_q\gets V_w^\top(x'_q-\mu)$ \Comment{\textit{\cref{eq:scores}}}
\For{$k=1,\dots,K$}
  \State $G\in\R^{w\times w}$ with i.i.d.\ $\mathcal N(0,1)$ entries;\quad $(Q,B)\gets\mathrm{QR}(G)$, $B$ upper triangular
  \State $R_k\gets Q\,\diag\bigl(\mathrm{sign}(B_{11}),\dots,\mathrm{sign}(B_{ww})\bigr)$, $\mathrm{sign}(0)=1$ \Comment{\textit{Haar}}
  \State $S_k\gets\{([R_k^\top z_i;\,t_i],\,y_i): i=1,\dots,n\}$ \Comment{\textit{context}}
  \State $\hat p_{q,k}\gets f\bigl(S_k,\,[R_k^\top z'_q;\,t'_q]\bigr)$ for all $q$ \Comment{\textit{one pass}}
\EndFor
\State \Return $\hat p_q=\frac1K\sum_{k=1}^{K}\hat p_{q,k}$ for all $q$
\end{algorithmic}
\end{algorithm}

%% file: tables/task_summary.tex
\tablestyle
\begin{NiceTabular}{lrrrrr}
\CodeBefore
  \rowcolor{tablesection}{2,6}
\Body
\toprule
\textbf{Source} & \textbf{Tasks} & \textbf{Classes} & \textbf{Pool} & \textbf{Query} & \textbf{Columns} \\
\midrule
\Block[l]{1-6}{\textit{Image tasks}} & & & & & \\
MulTaBench & 12 & 2--10 & 540--10{,}000 & 60--2{,}000 & 3--39 \\
Released with MulTaBench & 8 & 2--10 & 391--6{,}723 & 45--747 & 3--27 \\
MultiModalPFN & 4 & 2--6 & 1{,}318--11{,}721 & 326--2{,}931 & 8--21 \\
\midrule
\Block[l]{1-6}{\textit{Text tasks}} & & & & & \\
MulTaBench & 10 & 2--10 & 878--10{,}000 & 175--2{,}000 & 1--35 \\
Released with MulTaBench & 10 & 2--10 & 718--10{,}000 & 82--2{,}000 & 3--65 \\
\bottomrule
\end{NiceTabular}

%% file: tables/tasks.tex
\tablestyle
\begin{NiceTabular}{llrrrrr}
\toprule
\textbf{Task} & \textbf{Source} & \textbf{Classes} & \textbf{Pool} & \textbf{Query} & \textbf{Columns} & \textbf{Images} \\
\midrule
CELEB\_ATTRACTIVENESS & MulTaBench & 2 & 9{,}998 & 2{,}000 & 39 & 1 \\
HATEFUL\_MEME & MulTaBench & 2 & 8{,}098 & 902 & 20 & 1 \\
MAMMOGRAPHY\_CMMD & MulTaBench & 2 & 4{,}680 & 522 & 4 & 1 \\
CBIS\_DDSM & MulTaBench & 4 & 1{,}527 & 169 & 8 & 1 \\
CHEXPERT & MulTaBench & 3 & 10{,}000 & 2{,}000 & 17 & 1 \\
CSGO\_SKIN\_PRICE & MulTaBench & 10 & 861 & 95 & 3 & 1 \\
FLOWER\_BOUQUETS & MulTaBench & 5 & 540 & 60 & 3 & 1 \\
GLAUCOMA\_SMDG & MulTaBench & 3 & 10{,}000 & 1{,}244 & 8 & 1 \\
HUBMAP\_HPA & MulTaBench & 10 & 10{,}000 & 1{,}258 & 5 & 1 \\
JUSTIN\_INSTAGRAM & MulTaBench & 5 & 10{,}000 & 319 & 6 & 1 \\
PETFINDER & MulTaBench & 8 & 10{,}000 & 1{,}465 & 17 & 1 \\
ZOOSCAN\_ZOOPLANKTON & MulTaBench & 10 & 10{,}000 & 2{,}000 & 28 & 1 \\
OASIS\_ALZHEIMERS$^\dagger$ & MulTaBench & 2 & 391 & 45 & 7 & 1 \\
PINTEREST\_POPULAR$^\dagger$ & MulTaBench & 2 & 1{,}352 & 152 & 15 & 1 \\
HAM10000\_LESION$^\dagger$ & MulTaBench & 7 & 6{,}723 & 747 & 3 & 1 \\
HEARTHSTONE\_CLASS$^\dagger$ & MulTaBench & 10 of 11 & 3{,}581 & 398 & 11 & 1 \\
MINECRAFT$^\dagger$ & MulTaBench & 7 & 5{,}481 & 608 & 7 & 1 \\
PAD\_UFES\_LESION$^\dagger$ & MulTaBench & 6 & 1{,}477 & 164 & 21 & 1 \\
POKEMON\_HEIGHT$^\dagger$ & MulTaBench & 10 & 970 & 107 & 27 & 1 \\
REDDIT\_MEMES$^\dagger$ & MulTaBench & 7 & 4{,}339 & 482 & 5 & 1 \\
cbis\_calc & MultiModalPFN & 2 & 1{,}545 & 326 & 8 & 3 \\
cbis\_mass & MultiModalPFN & 2 & 1{,}318 & 378 & 8 & 3 \\
pad\_ufes\_20 & MultiModalPFN & 6 & 1{,}838 & 460 & 21 & 1 \\
petfinder & MultiModalPFN & 5 & 11{,}721 & 2{,}931 & 19 & 1 \\
\bottomrule
\end{NiceTabular}

%% file: tables/tasks_text.tex
\tablestyle
\begin{NiceTabular}{llrrrrr}
\toprule
\textbf{Task} & \textbf{Source} & \textbf{Classes} & \textbf{Pool} & \textbf{Query} & \textbf{Columns} & \textbf{Text fields} \\
\midrule
FAKE\_JOB\_POSTING & MulTaBench & 2 & 9{,}998 & 1{,}272 & 2 & 3 \\
JIGSAW\_TOXICITY & MulTaBench & 2 & 9{,}998 & 2{,}000 & 29 & 2 \\
KICKSTARTER\_FUNDING & MulTaBench & 2 & 9{,}998 & 2{,}000 & 4 & 5 \\
DATA\_SCIENTIST\_SALARY & MulTaBench & 6 & 10{,}000 & 1{,}584 & 1 & 5 \\
MICHELIN\_RESTAURANTS & MulTaBench & 5 & 10{,}000 & 1{,}884 & 5 & 6 \\
PRODUCT\_SENTIMENT & MulTaBench & 4 & 4{,}582 & 509 & 1 & 1 \\
SPOTIFY\_GENRES & MulTaBench & 10 of 114 & 878 & 175 & 15 & 3 \\
US\_ACCIDENTS & MulTaBench & 4 & 10{,}000 & 2{,}000 & 35 & 9 \\
WINE\_REVIEW & MulTaBench & 10 of 30 & 3{,}216 & 643 & 3 & 2 \\
WOMEN\_CLOTHING\_REVIEW & MulTaBench & 5 & 10{,}000 & 1{,}878 & 8 & 2 \\
CALIFORNIA\_PRICES$^\dagger$ & MulTaBench & 2 & 9{,}998 & 2{,}000 & 19 & 20 \\
IMDB\_GENRE$^\dagger$ & MulTaBench & 2 & 718 & 82 & 7 & 4 \\
OSHA\_INJURY$^\dagger$ & MulTaBench & 2 & 4{,}361 & 486 & 12 & 4 \\
AMERICAN\_EAGLE\_PRICES$^\dagger$ & MulTaBench & 5 & 10{,}000 & 2{,}000 & 8 & 5 \\
BOOKS\_GOODREADS$^\dagger$ & MulTaBench & 3 & 3{,}571 & 396 & 5 & 9 \\
BOX\_OFFICE$^\dagger$ & MulTaBench & 5 & 6{,}668 & 740 & 9 & 10 \\
CONSUMER\_COMPLAINT$^\dagger$ & MulTaBench & 8 & 10{,}000 & 2{,}000 & 9 & 7 \\
KOREAN\_DRAMA$^\dagger$ & MulTaBench & 10 & 1{,}483 & 164 & 3 & 6 \\
MELBOURNE\_AIRBNB$^\dagger$ & MulTaBench & 10 & 10{,}000 & 1{,}831 & 65 & 23 \\
NEWS\_CHANNEL$^\dagger$ & MulTaBench & 6 & 10{,}000 & 2{,}000 & 16 & 1 \\
\bottomrule
\end{NiceTabular}

%% file: tables/pfns.tex
\tablestyle
\begin{NiceTabular}{lcccc}
\CodeBefore
  \rowcolor{tablesection}{3,8,13,18,23,28,33,38}
\Body
\toprule
\Block{2-1}{\textbf{PFN}} & \Block{1-4}{\textbf{$\Delta$ AUROC} by labeled rows per class} & & & \\
 & \textbf{10} & \textbf{25} & \textbf{50} & \textbf{100} \\
\midrule
\Block[l]{1-5}{\textit{Image only, DINOv3}} & & & & \\
TabICLv2 & +2.1\,{\scriptsize$\pm$0.5} & +2.8\,{\scriptsize$\pm$0.9} & +2.7\,{\scriptsize$\pm$0.7} & +1.9\,{\scriptsize$\pm$0.7} \\
TabPFN v2 & +2.8\,{\scriptsize$\pm$0.6} & +2.4\,{\scriptsize$\pm$0.5} & +2.0\,{\scriptsize$\pm$0.6} & +1.5\,{\scriptsize$\pm$0.6} \\
TabPFN 2.5 & +2.6\,{\scriptsize$\pm$0.6} & +2.1\,{\scriptsize$\pm$0.5} & +1.7\,{\scriptsize$\pm$0.6} & +0.9\,{\scriptsize$\pm$0.3} \\
TabPFN 3.5 & +2.4\,{\scriptsize$\pm$0.5} & +2.3\,{\scriptsize$\pm$0.5} & +1.8\,{\scriptsize$\pm$0.4} & +1.4\,{\scriptsize$\pm$0.5} \\
\midrule
\Block[l]{1-5}{\textit{Image + table, DINOv3}} & & & & \\
TabICLv2 & +1.3\,{\scriptsize$\pm$0.4} & +1.6\,{\scriptsize$\pm$0.6} & +1.5\,{\scriptsize$\pm$0.7} & +1.2\,{\scriptsize$\pm$0.9} \\
TabPFN v2 & +1.2\,{\scriptsize$\pm$0.6} & +1.1\,{\scriptsize$\pm$0.4} & +1.1\,{\scriptsize$\pm$0.5} & +0.8\,{\scriptsize$\pm$0.5} \\
TabPFN 2.5 & +1.0\,{\scriptsize$\pm$0.6} & +0.8\,{\scriptsize$\pm$0.3} & +0.8\,{\scriptsize$\pm$0.5} & +0.5\,{\scriptsize$\pm$0.4} \\
TabPFN 3.5 & +0.8\,{\scriptsize$\pm$0.4} & +0.7\,{\scriptsize$\pm$0.4} & +0.7\,{\scriptsize$\pm$0.4} & +0.6\,{\scriptsize$\pm$0.5} \\
\midrule
\Block[l]{1-5}{\textit{Image only, SigLIP}} & & & & \\
TabICLv2 & +2.1\,{\scriptsize$\pm$0.6} & +2.5\,{\scriptsize$\pm$0.8} & +2.5\,{\scriptsize$\pm$0.7} & +2.0\,{\scriptsize$\pm$0.9} \\
TabPFN v2 & +2.5\,{\scriptsize$\pm$0.6} & +2.7\,{\scriptsize$\pm$0.7} & +2.0\,{\scriptsize$\pm$0.6} & +1.5\,{\scriptsize$\pm$0.6} \\
TabPFN 2.5 & +2.5\,{\scriptsize$\pm$0.5} & +2.3\,{\scriptsize$\pm$0.7} & +1.6\,{\scriptsize$\pm$0.6} & +1.0\,{\scriptsize$\pm$0.4} \\
TabPFN 3.5 & +2.3\,{\scriptsize$\pm$0.5} & +2.4\,{\scriptsize$\pm$0.6} & +1.8\,{\scriptsize$\pm$0.6} & +1.4\,{\scriptsize$\pm$0.7} \\
\midrule
\Block[l]{1-5}{\textit{Image + table, SigLIP}} & & & & \\
TabICLv2 & +1.3\,{\scriptsize$\pm$0.5} & +1.6\,{\scriptsize$\pm$0.7} & +1.5\,{\scriptsize$\pm$0.7} & +1.4\,{\scriptsize$\pm$1.0} \\
TabPFN v2 & +1.3\,{\scriptsize$\pm$0.5} & +1.4\,{\scriptsize$\pm$0.6} & +1.1\,{\scriptsize$\pm$0.5} & +0.9\,{\scriptsize$\pm$0.5} \\
TabPFN 2.5 & +0.9\,{\scriptsize$\pm$0.5} & +1.0\,{\scriptsize$\pm$0.5} & +0.7\,{\scriptsize$\pm$0.5} & +0.6\,{\scriptsize$\pm$0.4} \\
TabPFN 3.5 & +0.8\,{\scriptsize$\pm$0.4} & +0.9\,{\scriptsize$\pm$0.5} & +0.7\,{\scriptsize$\pm$0.4} & +0.6\,{\scriptsize$\pm$0.5} \\
\midrule
\Block[l]{1-5}{\textit{Text only, ELECTRA}} & & & & \\
TabICLv2 & +1.8\,{\scriptsize$\pm$0.6} & +2.3\,{\scriptsize$\pm$0.8} & +2.5\,{\scriptsize$\pm$0.9} & +1.5\,{\scriptsize$\pm$0.7} \\
TabPFN v2 & +2.4\,{\scriptsize$\pm$1.0} & +3.0\,{\scriptsize$\pm$0.8} & +2.6\,{\scriptsize$\pm$0.7} & +1.8\,{\scriptsize$\pm$0.7} \\
TabPFN 2.5 & +2.4\,{\scriptsize$\pm$1.0} & +2.8\,{\scriptsize$\pm$0.7} & +1.9\,{\scriptsize$\pm$0.5} & +1.1\,{\scriptsize$\pm$0.3} \\
TabPFN 3.5 & +1.9\,{\scriptsize$\pm$0.8} & +2.8\,{\scriptsize$\pm$0.7} & +2.0\,{\scriptsize$\pm$0.4} & +1.3\,{\scriptsize$\pm$0.4} \\
\midrule
\Block[l]{1-5}{\textit{Text + table, ELECTRA}} & & & & \\
TabICLv2 & +0.9\,{\scriptsize$\pm$0.4} & +1.8\,{\scriptsize$\pm$0.6} & +2.5\,{\scriptsize$\pm$1.1} & +1.2\,{\scriptsize$\pm$0.7} \\
TabPFN v2 & +0.9\,{\scriptsize$\pm$0.5} & +1.4\,{\scriptsize$\pm$0.8} & +1.6\,{\scriptsize$\pm$0.7} & +1.0\,{\scriptsize$\pm$0.4} \\
TabPFN 2.5 & +0.3\,{\scriptsize$\pm$0.5} & +1.1\,{\scriptsize$\pm$0.7} & +0.8\,{\scriptsize$\pm$0.5} & +0.7\,{\scriptsize$\pm$0.3} \\
TabPFN 3.5 & +0.1\,{\scriptsize$\pm$0.5} & +0.9\,{\scriptsize$\pm$0.6} & +0.9\,{\scriptsize$\pm$0.4} & +0.7\,{\scriptsize$\pm$0.5} \\
\midrule
\Block[l]{1-5}{\textit{Text only, E5}} & & & & \\
TabICLv2 & +1.5\,{\scriptsize$\pm$0.4} & +2.2\,{\scriptsize$\pm$0.7} & +2.7\,{\scriptsize$\pm$0.8} & +2.3\,{\scriptsize$\pm$0.9} \\
TabPFN v2 & +1.8\,{\scriptsize$\pm$0.7} & +2.3\,{\scriptsize$\pm$0.8} & +1.9\,{\scriptsize$\pm$0.6} & +1.4\,{\scriptsize$\pm$0.5} \\
TabPFN 2.5 & +2.0\,{\scriptsize$\pm$0.8} & +2.1\,{\scriptsize$\pm$0.7} & +1.5\,{\scriptsize$\pm$0.5} & +0.7\,{\scriptsize$\pm$0.4} \\
TabPFN 3.5 & +1.8\,{\scriptsize$\pm$0.5} & +1.9\,{\scriptsize$\pm$0.6} & +1.6\,{\scriptsize$\pm$0.5} & +1.4\,{\scriptsize$\pm$0.6} \\
\midrule
\Block[l]{1-5}{\textit{Text + table, E5}} & & & & \\
TabICLv2 & +1.0\,{\scriptsize$\pm$0.3} & +1.6\,{\scriptsize$\pm$0.7} & +2.4\,{\scriptsize$\pm$1.1} & +1.5\,{\scriptsize$\pm$0.8} \\
TabPFN v2 & +0.8\,{\scriptsize$\pm$0.6} & +0.9\,{\scriptsize$\pm$0.4} & +0.9\,{\scriptsize$\pm$0.4} & +0.8\,{\scriptsize$\pm$0.4} \\
TabPFN 2.5 & +0.8\,{\scriptsize$\pm$0.5} & +0.7\,{\scriptsize$\pm$0.4} & +0.7\,{\scriptsize$\pm$0.4} & +0.6\,{\scriptsize$\pm$0.4} \\
TabPFN 3.5 & +0.7\,{\scriptsize$\pm$0.4} & +0.6\,{\scriptsize$\pm$0.3} & +0.6\,{\scriptsize$\pm$0.3} & +0.8\,{\scriptsize$\pm$0.5} \\
\bottomrule
\end{NiceTabular}

%% file: tables/heads.tex
\tablestyle
\begin{NiceTabular}{lcccc}
\CodeBefore
  \rowcolor{tablesection}{3,10,17,24}
  \rowcolor{ourrow}{9,16,23,30}
\Body
\toprule
\Block{2-1}{\textbf{Head}} & \Block{1-4}{\textbf{AUROC} ($\uparrow$)\\ by labeled rows per class} & & & \\
 & \textbf{10} & \textbf{25} & \textbf{50} & \textbf{100} \\
\midrule
\Block[l]{1-5}{\textit{Image only, DINOv3}} & & & & \\
$k$-NN & 66.9 & 69.6 & 71.5 & 72.9 \\
Linear SVM & 66.2 & 72.9 & 75.5 & 77.2 \\
MLP & 68.3 & 72.7 & 75.1 & 77.2 \\
Linear probe & \textbf{70.1} & 74.0 & 76.3 & 78.2 \\
CoMET & 68.0 & 72.0 & 75.2 & 78.3 \\
\textbf{RoFE} & \textbf{70.1} & \textbf{74.8} & \textbf{77.9} & \textbf{80.2} \\
\midrule
\Block[l]{1-5}{\textit{Image + table, DINOv3}} & & & & \\
$k$-NN & 67.7 & 70.5 & 72.3 & 73.7 \\
Linear SVM & 67.1 & 74.4 & 77.2 & 79.4 \\
MLP & 68.9 & 73.4 & 76.2 & 78.5 \\
Linear probe & 71.2 & 75.3 & 77.8 & 80.0 \\
CoMET & 73.9 & 78.2 & 80.9 & 83.5 \\
\textbf{RoFE} & \textbf{75.2} & \textbf{79.7} & \textbf{82.4} & \textbf{84.7} \\
\midrule
\Block[l]{1-5}{\textit{Image only, SigLIP}} & & & & \\
$k$-NN & 69.7 & 72.5 & 74.5 & 76.0 \\
Linear SVM & 70.1 & 75.4 & 78.3 & 80.0 \\
MLP & 71.3 & 75.6 & 78.4 & 80.4 \\
Linear probe & 72.9 & 76.7 & 79.2 & 80.9 \\
CoMET & 71.0 & 75.1 & 78.3 & 80.9 \\
\textbf{RoFE} & \textbf{73.1} & \textbf{77.6} & \textbf{80.8} & \textbf{82.9} \\
\midrule
\Block[l]{1-5}{\textit{Image + table, SigLIP}} & & & & \\
$k$-NN & 70.0 & 73.0 & 75.0 & 76.5 \\
Linear SVM & 70.7 & 76.4 & 79.6 & 81.6 \\
MLP & 71.9 & 76.1 & 79.2 & 81.3 \\
Linear probe & 73.5 & 77.5 & 80.2 & 82.2 \\
CoMET & 75.7 & 80.0 & 82.7 & 84.8 \\
\textbf{RoFE} & \textbf{77.0} & \textbf{81.5} & \textbf{84.2} & \textbf{86.2} \\
\bottomrule
\end{NiceTabular}

%% file: tables/heads_text.tex
\tablestyle
\begin{NiceTabular}{lcccc}
\CodeBefore
  \rowcolor{tablesection}{3,10,17,24}
  \rowcolor{ourrow}{9,16,23,30}
\Body
\toprule
\Block{2-1}{\textbf{Head}} & \Block{1-4}{\textbf{AUROC} ($\uparrow$)\\ by labeled rows per class} & & & \\
 & \textbf{10} & \textbf{25} & \textbf{50} & \textbf{100} \\
\midrule
\Block[l]{1-5}{\textit{Text only, ELECTRA}} & & & & \\
$k$-NN & 61.3 & 63.3 & 65.0 & 66.3 \\
Linear SVM & 61.2 & 67.0 & 70.4 & 72.6 \\
MLP & 62.7 & 66.6 & 69.7 & 72.1 \\
Linear probe & 64.2 & 68.1 & 71.0 & 73.5 \\
CoMET & 62.8 & 66.7 & 69.7 & 73.3 \\
\textbf{RoFE} & \textbf{64.5} & \textbf{68.9} & \textbf{72.2} & \textbf{74.8} \\
\midrule
\Block[l]{1-5}{\textit{Text + table, ELECTRA}} & & & & \\
$k$-NN & 62.0 & 64.1 & 65.8 & 67.2 \\
Linear SVM & 62.2 & 69.3 & 73.0 & 75.7 \\
MLP & 63.5 & 67.9 & 71.2 & 74.1 \\
Linear probe & 65.5 & 70.0 & 73.4 & 76.3 \\
CoMET & 71.5 & 75.0 & 77.3 & 80.7 \\
\textbf{RoFE} & \textbf{72.4} & \textbf{76.8} & \textbf{79.8} & \textbf{81.8} \\
\midrule
\Block[l]{1-5}{\textit{Text only, E5}} & & & & \\
$k$-NN & 66.8 & 69.1 & 70.9 & 72.5 \\
Linear SVM & 65.8 & 71.4 & 73.5 & 75.0 \\
MLP & 67.2 & 70.7 & 73.3 & 75.3 \\
Linear probe & \textbf{69.3} & 72.1 & 74.2 & 75.9 \\
CoMET & 67.4 & 70.5 & 73.0 & 75.7 \\
\textbf{RoFE} & 68.9 & \textbf{72.7} & \textbf{75.8} & \textbf{78.0} \\
\midrule
\Block[l]{1-5}{\textit{Text + table, E5}} & & & & \\
$k$-NN & 67.4 & 69.7 & 71.4 & 73.1 \\
Linear SVM & 66.8 & 72.8 & 75.2 & 77.2 \\
MLP & 67.8 & 71.5 & 74.2 & 76.5 \\
Linear probe & 70.3 & 73.3 & 75.8 & 77.8 \\
CoMET & 74.2 & 77.4 & 79.5 & 82.1 \\
\textbf{RoFE} & \textbf{75.2} & \textbf{79.0} & \textbf{81.9} & \textbf{83.6} \\
\bottomrule
\end{NiceTabular}

%% file: tables/trained_image.tex
\tablestyle
\begin{NiceTabular}{lccccccc}
\CodeBefore
  \rowcolor{tablesection}{3,10,17,22}
  \rowcolor{ourrow}{7,14,20,25}
\Body
\toprule
\Block{2-1}{\textbf{Method}} & \Block{1-7}{\textbf{AUROC} ($\uparrow$) by labeled rows per class} & & & & & & \\
 & \textbf{10} & \textbf{25} & \textbf{50} & \textbf{100} & \textbf{250} & \textbf{500} & \textbf{All} \\
\midrule
\Block[l]{1-8}{\textit{Image only, DINOv3}} & & & & & & & \\
MultiModalPFN & 60.6 & 65.7 & 68.7 & 73.0 & 77.6 & 79.8 & 80.6 \\
CatBoost & 65.0 & 70.4 & 73.8 & 76.1 & 79.2 & 80.6 & 80.7 \\
CoMET & 68.0 & 72.0 & 75.2 & 78.3 & 82.0 & 83.3 & \textbf{84.3} \\
\textbf{\Method} & \textbf{70.1} & \textbf{74.8} & \textbf{77.9} & \textbf{80.2} & \textbf{82.8} & \textbf{83.8} & \textbf{84.3} \\
$\Delta$ MultiModalPFN & +9.6\,{\scriptsize$\pm$2.2} & +9.1\,{\scriptsize$\pm$2.3} & +9.2\,{\scriptsize$\pm$2.4} & +7.2\,{\scriptsize$\pm$1.7} & +5.2\,{\scriptsize$\pm$1.4} & +4.0\,{\scriptsize$\pm$1.2} & +3.7\,{\scriptsize$\pm$1.0} \\
$\Delta$ CatBoost & +5.1\,{\scriptsize$\pm$1.2} & +4.4\,{\scriptsize$\pm$0.8} & +4.1\,{\scriptsize$\pm$0.7} & +4.1\,{\scriptsize$\pm$0.7} & +3.6\,{\scriptsize$\pm$0.7} & +3.2\,{\scriptsize$\pm$0.6} & +3.6\,{\scriptsize$\pm$0.8} \\
\midrule
\Block[l]{1-8}{\textit{Image + table, DINOv3}} & & & & & & & \\
MultiModalPFN & 72.3 & 76.0 & 78.8 & 81.4 & 84.4 & 85.5 & 86.5 \\
CatBoost & 71.2 & 76.4 & 79.6 & 81.9 & 85.2 & 86.0 & 86.2 \\
CoMET & 73.9 & 78.2 & 80.9 & 83.5 & 86.8 & \textbf{88.0} & \textbf{89.0} \\
\textbf{\Method} & \textbf{75.2} & \textbf{79.7} & \textbf{82.4} & \textbf{84.7} & \textbf{87.1} & 87.8 & 88.6 \\
$\Delta$ MultiModalPFN & +2.9\,{\scriptsize$\pm$1.4} & +3.7\,{\scriptsize$\pm$1.2} & +3.6\,{\scriptsize$\pm$1.2} & +3.3\,{\scriptsize$\pm$1.2} & +2.7\,{\scriptsize$\pm$1.1} & +2.3\,{\scriptsize$\pm$1.2} & +2.1\,{\scriptsize$\pm$1.3} \\
$\Delta$ CatBoost & +4.0\,{\scriptsize$\pm$1.1} & +3.3\,{\scriptsize$\pm$1.0} & +2.8\,{\scriptsize$\pm$0.8} & +2.8\,{\scriptsize$\pm$0.8} & +1.9\,{\scriptsize$\pm$0.8} & +1.8\,{\scriptsize$\pm$0.6} & +2.4\,{\scriptsize$\pm$0.8} \\
\midrule
\Block[l]{1-8}{\textit{Image only, SigLIP}} & & & & & & & \\
CatBoost & 68.8 & 73.2 & 76.5 & 78.9 & 82.2 & 83.4 & 83.3 \\
CoMET & 71.0 & 75.1 & 78.3 & 80.9 & 84.7 & 86.1 & \textbf{86.6} \\
\textbf{\Method} & \textbf{73.1} & \textbf{77.6} & \textbf{80.8} & \textbf{82.9} & \textbf{85.5} & \textbf{86.4} & \textbf{86.6} \\
$\Delta$ CatBoost & +4.3\,{\scriptsize$\pm$1.0} & +4.4\,{\scriptsize$\pm$1.0} & +4.2\,{\scriptsize$\pm$0.8} & +4.0\,{\scriptsize$\pm$0.9} & +3.3\,{\scriptsize$\pm$0.9} & +3.0\,{\scriptsize$\pm$0.8} & +3.3\,{\scriptsize$\pm$0.9} \\
\midrule
\Block[l]{1-8}{\textit{Image + table, SigLIP}} & & & & & & & \\
CatBoost & 73.2 & 78.2 & 81.4 & 83.7 & 86.7 & 87.4 & 87.6 \\
CoMET & 75.7 & 80.0 & 82.7 & 84.8 & 88.2 & \textbf{89.3} & \textbf{90.0} \\
\textbf{\Method} & \textbf{77.0} & \textbf{81.5} & \textbf{84.2} & \textbf{86.2} & \textbf{88.5} & 89.1 & 89.5 \\
$\Delta$ CatBoost & +3.8\,{\scriptsize$\pm$1.0} & +3.3\,{\scriptsize$\pm$0.9} & +2.8\,{\scriptsize$\pm$0.8} & +2.5\,{\scriptsize$\pm$0.7} & +1.8\,{\scriptsize$\pm$0.5} & +1.6\,{\scriptsize$\pm$0.6} & +1.9\,{\scriptsize$\pm$0.5} \\
\bottomrule
\end{NiceTabular}

%% file: tables/trained_text.tex
\tablestyle
\begin{NiceTabular}{lccccccc}
\CodeBefore
  \rowcolor{tablesection}{3,8,13,18}
  \rowcolor{ourrow}{6,11,16,21}
\Body
\toprule
\Block{2-1}{\textbf{Method}} & \Block{1-7}{\textbf{AUROC} ($\uparrow$) by labeled rows per class} & & & & & & \\
 & \textbf{10} & \textbf{25} & \textbf{50} & \textbf{100} & \textbf{250} & \textbf{500} & \textbf{All} \\
\midrule
\Block[l]{1-8}{\textit{Text only, ELECTRA}} & & & & & & & \\
CatBoost & 60.4 & 63.8 & 67.2 & 69.7 & 72.1 & 73.9 & 76.0 \\
CoMET & 62.8 & 66.7 & 69.7 & 73.3 & 76.5 & \textbf{78.7} & \textbf{81.3} \\
\textbf{\Method} & \textbf{64.5} & \textbf{68.9} & \textbf{72.2} & \textbf{74.8} & \textbf{77.1} & 78.6 & 81.1 \\
\rlap{$\Delta$ CatBoost}\phantom{$\Delta$ MultiModalPFN} & +4.2\,{\scriptsize$\pm$1.0} & +5.2\,{\scriptsize$\pm$1.2} & +5.0\,{\scriptsize$\pm$1.1} & +5.1\,{\scriptsize$\pm$0.8} & +5.0\,{\scriptsize$\pm$0.9} & +4.7\,{\scriptsize$\pm$1.0} & +5.2\,{\scriptsize$\pm$0.9} \\
\midrule
\Block[l]{1-8}{\textit{Text + table, ELECTRA}} & & & & & & & \\
CatBoost & 69.8 & 73.5 & 76.6 & 78.9 & 81.0 & 82.4 & 83.9 \\
CoMET & 71.5 & 75.0 & 77.3 & 80.7 & 83.7 & \textbf{85.1} & \textbf{86.8} \\
\textbf{\Method} & \textbf{72.4} & \textbf{76.8} & \textbf{79.8} & \textbf{81.8} & \textbf{84.0} & \textbf{85.1} & 86.6 \\
\rlap{$\Delta$ CatBoost}\phantom{$\Delta$ MultiModalPFN} & +2.6\,{\scriptsize$\pm$0.5} & +3.3\,{\scriptsize$\pm$1.0} & +3.2\,{\scriptsize$\pm$0.8} & +2.9\,{\scriptsize$\pm$0.8} & +3.0\,{\scriptsize$\pm$0.8} & +2.7\,{\scriptsize$\pm$0.8} & +2.7\,{\scriptsize$\pm$0.8} \\
\midrule
\Block[l]{1-8}{\textit{Text only, E5}} & & & & & & & \\
CatBoost & 64.8 & 68.8 & 72.2 & 74.5 & 77.0 & 78.4 & 79.8 \\
CoMET & 67.4 & 70.5 & 73.0 & 75.7 & 79.0 & 81.0 & \textbf{83.5} \\
\textbf{\Method} & \textbf{68.9} & \textbf{72.7} & \textbf{75.8} & \textbf{78.0} & \textbf{79.8} & \textbf{81.3} & 83.4 \\
\rlap{$\Delta$ CatBoost}\phantom{$\Delta$ MultiModalPFN} & +4.0\,{\scriptsize$\pm$1.1} & +3.9\,{\scriptsize$\pm$1.0} & +3.6\,{\scriptsize$\pm$0.8} & +3.5\,{\scriptsize$\pm$0.8} & +2.8\,{\scriptsize$\pm$0.7} & +2.9\,{\scriptsize$\pm$0.8} & +3.7\,{\scriptsize$\pm$0.9} \\
\midrule
\Block[l]{1-8}{\textit{Text + table, E5}} & & & & & & & \\
CatBoost & 72.4 & 76.5 & 79.5 & 81.3 & 83.5 & 84.6 & 85.6 \\
CoMET & 74.2 & 77.4 & 79.5 & 82.1 & 84.9 & 86.4 & \textbf{88.0} \\
\textbf{\Method} & \textbf{75.2} & \textbf{79.0} & \textbf{81.9} & \textbf{83.6} & \textbf{85.3} & \textbf{86.5} & 87.8 \\
\rlap{$\Delta$ CatBoost}\phantom{$\Delta$ MultiModalPFN} & +2.8\,{\scriptsize$\pm$1.1} & +2.5\,{\scriptsize$\pm$0.9} & +2.4\,{\scriptsize$\pm$0.6} & +2.3\,{\scriptsize$\pm$0.6} & +1.8\,{\scriptsize$\pm$0.5} & +1.8\,{\scriptsize$\pm$0.5} & +2.2\,{\scriptsize$\pm$0.7} \\
\bottomrule
\end{NiceTabular}

%% file: tables/mmpfn_config.tex
\tablestyle
\begin{NiceTabular}{lcccc}
\CodeBefore
  \rowcolor{tablesection}{3,10}
  \rowcolor{ourrow}{7,14}
\Body
\toprule
\Block{2-1}{\textbf{Method}} & \Block{1-4}{\textbf{AUROC} ($\uparrow$) by labeled rows per class} & & & \\
 & \textbf{10} & \textbf{25} & \textbf{50} & \textbf{100} \\
\midrule
\Block[l]{1-5}{\textit{Image only, DINOv3}} & & & & \\
MultiModalPFN (authors) & 60.6 & 65.7 & 68.7 & 73.0 \\
MultiModalPFN (val.\ loss) & 63.1 & 67.9 & 71.0 & 74.2 \\
MultiModalPFN (best per task) & 64.1 & 68.7 & 72.0 & 75.3 \\
\textbf{\Method} & \textbf{70.1} & \textbf{74.8} & \textbf{77.9} & \textbf{80.2} \\
$\Delta$ val.\ loss & +7.0\,{\scriptsize$\pm$1.9} & +6.8\,{\scriptsize$\pm$1.9} & +6.9\,{\scriptsize$\pm$1.6} & +5.9\,{\scriptsize$\pm$1.4} \\
$\Delta$ best per task & +6.1\,{\scriptsize$\pm$2.0} & +6.0\,{\scriptsize$\pm$1.9} & +5.9\,{\scriptsize$\pm$1.7} & +4.9\,{\scriptsize$\pm$1.5} \\
\midrule
\Block[l]{1-5}{\textit{Image + table, DINOv3}} & & & & \\
MultiModalPFN (authors) & 72.3 & 76.0 & 78.8 & 81.4 \\
MultiModalPFN (val.\ loss) & 70.7 & 75.6 & 78.8 & 81.1 \\
MultiModalPFN (best per task) & 73.5 & 77.5 & 80.2 & 82.2 \\
\textbf{\Method} & \textbf{75.2} & \textbf{79.7} & \textbf{82.4} & \textbf{84.7} \\
$\Delta$ val.\ loss & +4.5\,{\scriptsize$\pm$1.3} & +4.1\,{\scriptsize$\pm$1.2} & +3.7\,{\scriptsize$\pm$1.0} & +3.5\,{\scriptsize$\pm$1.1} \\
$\Delta$ best per task & +1.7\,{\scriptsize$\pm$1.5} & +2.2\,{\scriptsize$\pm$1.3} & +2.3\,{\scriptsize$\pm$1.2} & +2.5\,{\scriptsize$\pm$1.1} \\
\bottomrule
\end{NiceTabular}

%% file: tables/zhang.tex
\tablestyle
\begin{NiceTabular}{lcccc}
\CodeBefore
  \rowcolor{tablesection}{3,13,23,33}
  \rowcolor{ourrow}{8,9,10,18,19,20,28,29,30,38,39,40}
\Body
\toprule
\Block{2-1}{\textbf{Method}} & \Block{1-4}{\textbf{AUROC} ($\uparrow$) by labeled rows per class} & & & \\
 & \textbf{10} & \textbf{25} & \textbf{50} & \textbf{100} \\
\midrule
\Block[l]{1-5}{\textit{Image only, DINOv3}} & & & & \\
CoMET, TabPFN 2.5 & 68.1 & 73.0 & 75.9 & 78.5 \\
Best fixed $d$, TabPFN 2.5 & 69.1 & 73.8 & 76.7 & 78.8 \\
Best PFN or head and $d$ & 69.6 & 74.6 & 77.7 & 79.9 \\
Best per task & 70.8 & 75.2 & 78.1 & 80.2 \\
\textbf{RoFE, TabPFN 2.5} & 70.7 & 75.1 & 77.6 & 79.4 \\
\textbf{RoFE, TabICLv2} & 70.1 & 74.8 & 77.9 & 80.2 \\
\textbf{RoFE, TabPFN 3.5} & \textbf{71.1} & \textbf{75.6} & \textbf{78.2} & \textbf{80.3} \\
$\Delta$ best fixed $d$ & +1.6\,{\scriptsize$\pm$0.4} & +1.3\,{\scriptsize$\pm$0.4} & +0.8\,{\scriptsize$\pm$0.5} & +0.6\,{\scriptsize$\pm$0.5} \\
$\Delta$ best PFN or head & +1.6\,{\scriptsize$\pm$0.8} & +1.0\,{\scriptsize$\pm$0.4} & +0.5\,{\scriptsize$\pm$0.4} & +0.4\,{\scriptsize$\pm$0.3} \\
\midrule
\Block[l]{1-5}{\textit{Image + table, DINOv3}} & & & & \\
CoMET, TabPFN 2.5 & 74.9 & 79.4 & 81.6 & 83.6 \\
Best fixed $d$, TabPFN 2.5 & 75.6 & 80.0 & 82.5 & 84.4 \\
Best PFN or head and $d$ & 76.5 & 81.2 & 83.5 & 85.4 \\
Best per task & \textbf{77.0} & \textbf{81.5} & \textbf{83.8} & \textbf{85.6} \\
\textbf{RoFE, TabPFN 2.5} & 76.0 & 80.2 & 82.4 & 84.1 \\
\textbf{RoFE, TabICLv2} & 75.2 & 79.7 & 82.4 & 84.7 \\
\textbf{RoFE, TabPFN 3.5} & 76.9 & 81.1 & 83.2 & 85.2 \\
$\Delta$ best fixed $d$ & +0.4\,{\scriptsize$\pm$0.4} & +0.2\,{\scriptsize$\pm$0.3} & $-$0.1\,{\scriptsize$\pm$0.5} & $-$0.3\,{\scriptsize$\pm$0.5} \\
$\Delta$ best PFN or head & +0.4\,{\scriptsize$\pm$0.3} & $-$0.1\,{\scriptsize$\pm$0.3} & $-$0.3\,{\scriptsize$\pm$0.5} & $-$0.2\,{\scriptsize$\pm$0.4} \\
\midrule
\Block[l]{1-5}{\textit{Image only, SigLIP}} & & & & \\
CoMET, TabPFN 2.5 & 71.1 & 75.6 & 78.8 & 81.2 \\
Best fixed $d$, TabPFN 2.5 & 72.4 & 76.7 & 79.6 & 81.7 \\
Best PFN or head and $d$ & 72.4 & 77.4 & 80.5 & 82.7 \\
Best per task & 73.7 & 78.1 & 80.8 & \textbf{83.0} \\
\textbf{RoFE, TabPFN 2.5} & 73.6 & 77.9 & 80.4 & 82.2 \\
\textbf{RoFE, TabICLv2} & 73.1 & 77.6 & 80.8 & 82.9 \\
\textbf{RoFE, TabPFN 3.5} & \textbf{74.0} & \textbf{78.3} & \textbf{81.0} & \textbf{83.0} \\
$\Delta$ best fixed $d$ & +1.3\,{\scriptsize$\pm$0.3} & +1.3\,{\scriptsize$\pm$0.5} & +0.9\,{\scriptsize$\pm$0.4} & +0.5\,{\scriptsize$\pm$0.4} \\
$\Delta$ best PFN or head & +1.5\,{\scriptsize$\pm$0.5} & +1.0\,{\scriptsize$\pm$0.4} & +0.5\,{\scriptsize$\pm$0.3} & +0.3\,{\scriptsize$\pm$0.2} \\
\midrule
\Block[l]{1-5}{\textit{Image + table, SigLIP}} & & & & \\
CoMET, TabPFN 2.5 & 76.8 & 81.0 & 83.4 & 85.1 \\
Best fixed $d$, TabPFN 2.5 & 77.3 & 81.6 & 84.1 & 85.9 \\
Best PFN or head and $d$ & 78.1 & 82.7 & 84.9 & 86.8 \\
Best per task & \textbf{78.6} & \textbf{82.9} & \textbf{85.1} & \textbf{86.9} \\
\textbf{RoFE, TabPFN 2.5} & 77.8 & 82.0 & 84.1 & 85.7 \\
\textbf{RoFE, TabICLv2} & 77.0 & 81.5 & 84.2 & 86.2 \\
\textbf{RoFE, TabPFN 3.5} & 78.5 & 82.6 & 84.7 & 86.6 \\
$\Delta$ best fixed $d$ & +0.5\,{\scriptsize$\pm$0.4} & +0.3\,{\scriptsize$\pm$0.4} & $-$0.1\,{\scriptsize$\pm$0.3} & $-$0.2\,{\scriptsize$\pm$0.3} \\
$\Delta$ best PFN or head & +0.4\,{\scriptsize$\pm$0.3} & $-$0.1\,{\scriptsize$\pm$0.4} & $-$0.2\,{\scriptsize$\pm$0.3} & $-$0.2\,{\scriptsize$\pm$0.2} \\
\bottomrule
\end{NiceTabular}

%% file: tables/zhang_text.tex
\tablestyle
\begin{NiceTabular}{lcccc}
\CodeBefore
  \rowcolor{tablesection}{3,13,23,33}
  \rowcolor{ourrow}{8,9,10,18,19,20,28,29,30,38,39,40}
\Body
\toprule
\Block{2-1}{\textbf{Method}} & \Block{1-4}{\textbf{AUROC} ($\uparrow$) by labeled rows per class} & & & \\
 & \textbf{10} & \textbf{25} & \textbf{50} & \textbf{100} \\
\midrule
\Block[l]{1-5}{\textit{Text only, ELECTRA}} & & & & \\
CoMET, TabPFN 2.5 & 62.0 & 66.5 & 70.5 & 73.6 \\
Best fixed $d$, TabPFN 2.5 & 63.4 & 67.7 & 70.6 & 73.6 \\
Best PFN or head and $d$ & 64.1 & 68.5 & 71.7 & 74.5 \\
Best per task & \textbf{64.8} & 69.1 & 72.3 & \textbf{74.8} \\
\textbf{RoFE, TabPFN 2.5} & 64.4 & 69.2 & 72.3 & 74.7 \\
\textbf{RoFE, TabICLv2} & 64.5 & 68.9 & 72.2 & \textbf{74.8} \\
\textbf{RoFE, TabPFN 3.5} & 64.6 & \textbf{69.3} & \textbf{72.4} & \textbf{74.8} \\
$\Delta$ best fixed $d$ & +1.0\,{\scriptsize$\pm$0.5} & +1.5\,{\scriptsize$\pm$0.5} & +1.7\,{\scriptsize$\pm$0.6} & +1.1\,{\scriptsize$\pm$0.3} \\
$\Delta$ best PFN or head & +0.5\,{\scriptsize$\pm$0.7} & +0.8\,{\scriptsize$\pm$0.3} & +0.8\,{\scriptsize$\pm$0.3} & +0.4\,{\scriptsize$\pm$0.2} \\
\midrule
\Block[l]{1-5}{\textit{Text + table, ELECTRA}} & & & & \\
CoMET, TabPFN 2.5 & 72.3 & 75.9 & 78.9 & 80.9 \\
Best fixed $d$, TabPFN 2.5 & 73.0 & 76.7 & 79.4 & 81.6 \\
Best PFN or head and $d$ & 73.6 & 77.5 & 80.0 & 82.1 \\
Best per task & \textbf{74.2} & \textbf{78.0} & \textbf{80.4} & \textbf{82.3} \\
\textbf{RoFE, TabPFN 2.5} & 72.6 & 76.9 & 79.7 & 81.6 \\
\textbf{RoFE, TabICLv2} & 72.4 & 76.8 & 79.8 & 81.8 \\
\textbf{RoFE, TabPFN 3.5} & 73.3 & 77.5 & 80.2 & 82.0 \\
$\Delta$ best fixed $d$ & $-$0.4\,{\scriptsize$\pm$0.7} & +0.2\,{\scriptsize$\pm$0.7} & +0.3\,{\scriptsize$\pm$0.8} & 0.0\,{\scriptsize$\pm$0.4} \\
$\Delta$ best PFN or head & $-$0.3\,{\scriptsize$\pm$0.6} & 0.0\,{\scriptsize$\pm$0.5} & +0.1\,{\scriptsize$\pm$0.4} & $-$0.1\,{\scriptsize$\pm$0.3} \\
\midrule
\Block[l]{1-5}{\textit{Text only, E5}} & & & & \\
CoMET, TabPFN 2.5 & 67.4 & 70.9 & 73.8 & 76.3 \\
Best fixed $d$, TabPFN 2.5 & 68.3 & 72.0 & 74.7 & 76.8 \\
Best PFN or head and $d$ & 68.8 & 72.6 & 75.5 & 77.7 \\
Best per task & \textbf{69.6} & 73.2 & 75.8 & 77.9 \\
\textbf{RoFE, TabPFN 2.5} & 69.4 & 73.0 & 75.3 & 77.1 \\
\textbf{RoFE, TabICLv2} & 68.9 & 72.7 & 75.8 & \textbf{78.0} \\
\textbf{RoFE, TabPFN 3.5} & \textbf{69.6} & \textbf{73.4} & \textbf{75.9} & 77.9 \\
$\Delta$ best fixed $d$ & +1.1\,{\scriptsize$\pm$0.4} & +1.0\,{\scriptsize$\pm$0.4} & +0.7\,{\scriptsize$\pm$0.4} & +0.2\,{\scriptsize$\pm$0.4} \\
$\Delta$ best PFN or head & +0.8\,{\scriptsize$\pm$0.6} & +0.8\,{\scriptsize$\pm$0.3} & +0.5\,{\scriptsize$\pm$0.3} & +0.2\,{\scriptsize$\pm$0.2} \\
\midrule
\Block[l]{1-5}{\textit{Text + table, E5}} & & & & \\
CoMET, TabPFN 2.5 & 75.3 & 78.8 & 80.9 & 82.4 \\
Best fixed $d$, TabPFN 2.5 & 75.8 & 79.4 & 81.8 & 83.4 \\
Best PFN or head and $d$ & 76.3 & 80.1 & 82.2 & 83.9 \\
Best per task & \textbf{77.1} & \textbf{80.5} & \textbf{82.5} & \textbf{84.0} \\
\textbf{RoFE, TabPFN 2.5} & 76.0 & 79.5 & 81.7 & 83.0 \\
\textbf{RoFE, TabICLv2} & 75.2 & 79.0 & 81.9 & 83.6 \\
\textbf{RoFE, TabPFN 3.5} & 76.5 & 80.1 & 82.2 & 83.6 \\
$\Delta$ best fixed $d$ & +0.2\,{\scriptsize$\pm$0.4} & +0.1\,{\scriptsize$\pm$0.3} & $-$0.1\,{\scriptsize$\pm$0.3} & $-$0.4\,{\scriptsize$\pm$0.4} \\
$\Delta$ best PFN or head & +0.2\,{\scriptsize$\pm$0.3} & 0.0\,{\scriptsize$\pm$0.3} & $-$0.1\,{\scriptsize$\pm$0.3} & $-$0.3\,{\scriptsize$\pm$0.3} \\
\bottomrule
\end{NiceTabular}

%% file: tables/kcurve.tex
\tablestyle
\begin{NiceTabular}{lcccc}
\CodeBefore
  \rowcolor{tablesection}{3,7,11,15}
\Body
\toprule
\Block{2-1}{\textbf{Rotations}} & \Block{1-4}{\textbf{$\Delta$ AUROC} by labeled rows per class} & & & \\
 & \textbf{10} & \textbf{25} & \textbf{50} & \textbf{100} \\
\midrule
\Block[l]{1-5}{\textit{Image only, DINOv3}} & & & & \\
1 rotation & $-$0.4\,{\scriptsize$\pm$0.1} & $-$0.4\,{\scriptsize$\pm$0.1} & $-$0.3\,{\scriptsize$\pm$0.1} & $-$0.3\,{\scriptsize$\pm$0.1} \\
8 rotations & +0.1\,{\scriptsize$\pm$0.0} & +0.1\,{\scriptsize$\pm$0.0} & +0.1\,{\scriptsize$\pm$0.0} & +0.1\,{\scriptsize$\pm$0.0} \\
16 rotations & +0.1\,{\scriptsize$\pm$0.1} & +0.1\,{\scriptsize$\pm$0.0} & +0.1\,{\scriptsize$\pm$0.0} & +0.1\,{\scriptsize$\pm$0.0} \\
\midrule
\Block[l]{1-5}{\textit{Image + table, DINOv3}} & & & & \\
1 rotation & $-$0.4\,{\scriptsize$\pm$0.1} & $-$0.4\,{\scriptsize$\pm$0.1} & $-$0.2\,{\scriptsize$\pm$0.1} & $-$0.2\,{\scriptsize$\pm$0.1} \\
8 rotations & +0.1\,{\scriptsize$\pm$0.1} & +0.1\,{\scriptsize$\pm$0.0} & +0.1\,{\scriptsize$\pm$0.0} & 0.0\,{\scriptsize$\pm$0.0} \\
16 rotations & +0.1\,{\scriptsize$\pm$0.1} & +0.1\,{\scriptsize$\pm$0.0} & +0.1\,{\scriptsize$\pm$0.0} & +0.1\,{\scriptsize$\pm$0.0} \\
\midrule
\Block[l]{1-5}{\textit{Image only, SigLIP}} & & & & \\
1 rotation & $-$0.3\,{\scriptsize$\pm$0.1} & $-$0.3\,{\scriptsize$\pm$0.1} & $-$0.3\,{\scriptsize$\pm$0.1} & $-$0.3\,{\scriptsize$\pm$0.1} \\
8 rotations & 0.0\,{\scriptsize$\pm$0.0} & +0.1\,{\scriptsize$\pm$0.0} & 0.0\,{\scriptsize$\pm$0.0} & +0.1\,{\scriptsize$\pm$0.0} \\
16 rotations & 0.0\,{\scriptsize$\pm$0.1} & +0.1\,{\scriptsize$\pm$0.1} & 0.0\,{\scriptsize$\pm$0.1} & +0.1\,{\scriptsize$\pm$0.0} \\
\midrule
\Block[l]{1-5}{\textit{Image + table, SigLIP}} & & & & \\
1 rotation & $-$0.4\,{\scriptsize$\pm$0.1} & $-$0.4\,{\scriptsize$\pm$0.1} & $-$0.2\,{\scriptsize$\pm$0.1} & $-$0.2\,{\scriptsize$\pm$0.1} \\
8 rotations & 0.0\,{\scriptsize$\pm$0.0} & +0.1\,{\scriptsize$\pm$0.0} & +0.1\,{\scriptsize$\pm$0.0} & +0.1\,{\scriptsize$\pm$0.0} \\
16 rotations & +0.1\,{\scriptsize$\pm$0.0} & +0.1\,{\scriptsize$\pm$0.1} & +0.1\,{\scriptsize$\pm$0.0} & +0.1\,{\scriptsize$\pm$0.0} \\
\bottomrule
\end{NiceTabular}

%% file: tables/basis.tex
\tablestyle
\begin{NiceTabular}{lcccccc}
\CodeBefore
  \rowcolor{tablesection}{3,11,19,27}
  \rowcolor{ourrow}{10,18,26,34}
\Body
\toprule
\Block{2-1}{\textbf{Change of basis}} & \Block{2-1}{\textbf{Mixes strong}\\\textbf{and weak}} & \Block{2-1}{\textbf{Column}\\\textbf{variances}} & \Block{1-4}{\textbf{$\Delta$ AUROC} by labeled rows per class} & & & \\
 &  &  & \textbf{10} & \textbf{25} & \textbf{50} & \textbf{100} \\
\midrule
\Block[l]{1-7}{\textit{TabICLv2, image only, DINOv3}} & & & & & & \\
No rotation, four PFN seeds averaged & \xmark & unequal & +0.1\,{\scriptsize$\pm$0.1} & +0.2\,{\scriptsize$\pm$0.1} & +0.2\,{\scriptsize$\pm$0.1} & +0.1\,{\scriptsize$\pm$0.0} \\
Rotation within the strong directions & \xmark & unequal & +0.3\,{\scriptsize$\pm$0.2} & +1.1\,{\scriptsize$\pm$0.5} & +1.4\,{\scriptsize$\pm$0.5} & +1.0\,{\scriptsize$\pm$0.4} \\
Rotation within the weak directions & \xmark & unequal & +0.6\,{\scriptsize$\pm$0.3} & +1.1\,{\scriptsize$\pm$0.5} & +1.0\,{\scriptsize$\pm$0.4} & +0.7\,{\scriptsize$\pm$0.3} \\
Givens rotations (sparse) & \cmark & equal & +1.5\,{\scriptsize$\pm$0.5} & +2.0\,{\scriptsize$\pm$0.9} & +1.1\,{\scriptsize$\pm$0.6} & +0.6\,{\scriptsize$\pm$0.6} \\
Hadamard transform (dense) & \cmark & equal & +1.9\,{\scriptsize$\pm$0.6} & +2.5\,{\scriptsize$\pm$0.9} & +2.4\,{\scriptsize$\pm$0.7} & +1.5\,{\scriptsize$\pm$0.7} \\
One random rotation & \cmark & near equal & +1.7\,{\scriptsize$\pm$0.5} & +2.4\,{\scriptsize$\pm$0.9} & +2.4\,{\scriptsize$\pm$0.7} & +1.6\,{\scriptsize$\pm$0.7} \\
\textbf{RoFE: four random rotations averaged} & \cmark & near equal & \textbf{+2.1\,{\scriptsize$\pm$0.5}} & \textbf{+2.8\,{\scriptsize$\pm$0.9}} & \textbf{+2.7\,{\scriptsize$\pm$0.7}} & \textbf{+1.9\,{\scriptsize$\pm$0.7}} \\
\midrule
\Block[l]{1-7}{\textit{TabICLv2, image only, SigLIP}} & & & & & & \\
No rotation, four PFN seeds averaged & \xmark & unequal & +0.1\,{\scriptsize$\pm$0.1} & +0.2\,{\scriptsize$\pm$0.1} & +0.1\,{\scriptsize$\pm$0.1} & +0.1\,{\scriptsize$\pm$0.1} \\
Rotation within the strong directions & \xmark & unequal & +0.2\,{\scriptsize$\pm$0.3} & +1.1\,{\scriptsize$\pm$0.5} & +1.1\,{\scriptsize$\pm$0.5} & +1.0\,{\scriptsize$\pm$0.5} \\
Rotation within the weak directions & \xmark & unequal & +0.4\,{\scriptsize$\pm$0.3} & +0.8\,{\scriptsize$\pm$0.3} & +0.8\,{\scriptsize$\pm$0.4} & +0.6\,{\scriptsize$\pm$0.4} \\
Givens rotations (sparse) & \cmark & equal & +1.5\,{\scriptsize$\pm$0.7} & +1.7\,{\scriptsize$\pm$0.9} & +1.2\,{\scriptsize$\pm$0.6} & +1.0\,{\scriptsize$\pm$0.8} \\
Hadamard transform (dense) & \cmark & equal & +1.9\,{\scriptsize$\pm$0.7} & +2.3\,{\scriptsize$\pm$0.8} & +2.1\,{\scriptsize$\pm$0.7} & +1.7\,{\scriptsize$\pm$0.8} \\
One random rotation & \cmark & near equal & +1.8\,{\scriptsize$\pm$0.6} & +2.2\,{\scriptsize$\pm$0.8} & +2.2\,{\scriptsize$\pm$0.7} & +1.7\,{\scriptsize$\pm$0.8} \\
\textbf{RoFE: four random rotations averaged} & \cmark & near equal & \textbf{+2.1\,{\scriptsize$\pm$0.6}} & \textbf{+2.5\,{\scriptsize$\pm$0.8}} & \textbf{+2.5\,{\scriptsize$\pm$0.7}} & \textbf{+2.0\,{\scriptsize$\pm$0.9}} \\
\midrule
\Block[l]{1-7}{\textit{TabPFN v2, image only, DINOv3}} & & & & & & \\
No rotation, four PFN seeds averaged & \xmark & unequal & +0.3\,{\scriptsize$\pm$0.1} & +0.3\,{\scriptsize$\pm$0.2} & +0.4\,{\scriptsize$\pm$0.2} & +0.3\,{\scriptsize$\pm$0.1} \\
Rotation within the strong directions & \xmark & unequal & +0.5\,{\scriptsize$\pm$0.4} & +0.7\,{\scriptsize$\pm$0.4} & +0.5\,{\scriptsize$\pm$0.3} & +0.5\,{\scriptsize$\pm$0.3} \\
Rotation within the weak directions & \xmark & unequal & +0.5\,{\scriptsize$\pm$0.3} & +0.3\,{\scriptsize$\pm$0.2} & +0.4\,{\scriptsize$\pm$0.3} & +0.4\,{\scriptsize$\pm$0.2} \\
Givens rotations (sparse) & \cmark & equal & +1.2\,{\scriptsize$\pm$0.4} & +0.9\,{\scriptsize$\pm$0.5} & +0.5\,{\scriptsize$\pm$0.5} & +0.2\,{\scriptsize$\pm$0.5} \\
Hadamard transform (dense) & \cmark & equal & +1.5\,{\scriptsize$\pm$0.4} & +1.1\,{\scriptsize$\pm$0.5} & +0.8\,{\scriptsize$\pm$0.5} & +0.6\,{\scriptsize$\pm$0.5} \\
One random rotation & \cmark & near equal & +1.7\,{\scriptsize$\pm$0.5} & +1.2\,{\scriptsize$\pm$0.5} & +1.1\,{\scriptsize$\pm$0.6} & +0.7\,{\scriptsize$\pm$0.5} \\
\textbf{RoFE: four random rotations averaged} & \cmark & near equal & \textbf{+2.8\,{\scriptsize$\pm$0.6}} & \textbf{+2.4\,{\scriptsize$\pm$0.5}} & \textbf{+2.0\,{\scriptsize$\pm$0.6}} & \textbf{+1.5\,{\scriptsize$\pm$0.6}} \\
\midrule
\Block[l]{1-7}{\textit{TabPFN v2, image only, SigLIP}} & & & & & & \\
No rotation, four PFN seeds averaged & \xmark & unequal & +0.2\,{\scriptsize$\pm$0.1} & +0.3\,{\scriptsize$\pm$0.2} & +0.4\,{\scriptsize$\pm$0.2} & +0.3\,{\scriptsize$\pm$0.1} \\
Rotation within the strong directions & \xmark & unequal & +0.6\,{\scriptsize$\pm$0.5} & +0.9\,{\scriptsize$\pm$0.5} & +0.5\,{\scriptsize$\pm$0.3} & +0.5\,{\scriptsize$\pm$0.2} \\
Rotation within the weak directions & \xmark & unequal & +0.2\,{\scriptsize$\pm$0.2} & +0.3\,{\scriptsize$\pm$0.3} & +0.3\,{\scriptsize$\pm$0.2} & +0.2\,{\scriptsize$\pm$0.2} \\
Givens rotations (sparse) & \cmark & equal & +1.1\,{\scriptsize$\pm$0.5} & +1.3\,{\scriptsize$\pm$0.6} & +0.6\,{\scriptsize$\pm$0.5} & +0.3\,{\scriptsize$\pm$0.4} \\
Hadamard transform (dense) & \cmark & equal & +1.6\,{\scriptsize$\pm$0.5} & +1.4\,{\scriptsize$\pm$0.5} & +0.9\,{\scriptsize$\pm$0.5} & +0.6\,{\scriptsize$\pm$0.4} \\
One random rotation & \cmark & near equal & +1.6\,{\scriptsize$\pm$0.5} & +1.8\,{\scriptsize$\pm$0.7} & +1.2\,{\scriptsize$\pm$0.6} & +0.7\,{\scriptsize$\pm$0.4} \\
\textbf{RoFE: four random rotations averaged} & \cmark & near equal & \textbf{+2.5\,{\scriptsize$\pm$0.6}} & \textbf{+2.7\,{\scriptsize$\pm$0.7}} & \textbf{+2.0\,{\scriptsize$\pm$0.6}} & \textbf{+1.5\,{\scriptsize$\pm$0.6}} \\
\bottomrule
\end{NiceTabular}

%% file: tables/scaling_tabpfn.tex
\tablestyle
\begin{NiceTabular}{cccccc}
\CodeBefore
  \rowcolor{tablesection}{3,8}
\Body
\toprule
\Block{2-1}{\textbf{Rotated}} & \Block{2-1}{\textbf{Standardized}} & \Block{1-4}{\textbf{AUROC} ($\uparrow$)\\ by labeled rows per class} & & & \\
 &  & \textbf{10} & \textbf{25} & \textbf{50} & \textbf{100} \\
\midrule
\Block[l]{1-6}{\textit{Image only, DINOv3}} & & & & & \\
-- & -- & 68.4 & 72.3 & 75.0 & 77.4 \\
-- & \ding{51} & 67.9 & 72.5 & 75.2 & 77.7 \\
\ding{51} & -- & \textbf{69.4} & \textbf{73.6} & \textbf{76.4} & \textbf{78.3} \\
\ding{51} & \ding{51} & \textbf{69.4} & \textbf{73.6} & \textbf{76.4} & \textbf{78.3} \\
\midrule
\Block[l]{1-6}{\textit{Image only, SigLIP}} & & & & & \\
-- & -- & 71.3 & 75.1 & 78.0 & 80.0 \\
-- & \ding{51} & 70.9 & 75.1 & 78.1 & 80.5 \\
\ding{51} & -- & \textbf{72.4} & \textbf{76.7} & \textbf{79.2} & \textbf{81.2} \\
\ding{51} & \ding{51} & 72.3 & \textbf{76.7} & \textbf{79.2} & \textbf{81.2} \\
\bottomrule
\end{NiceTabular}

%% file: tables/fewer_columns.tex
\tablestyle
\begin{NiceTabular}{lcccc}
\CodeBefore
  \rowcolor{tablesection}{3,8,13,18}
\Body
\toprule
\Block{2-1}{\textbf{Comparison}} & \Block{1-4}{\textbf{$\Delta$ AUROC} by labeled rows per class} & & & \\
 & \textbf{10} & \textbf{25} & \textbf{50} & \textbf{100} \\
\midrule
\Block[l]{1-5}{\textit{Image only, DINOv3}} & & & & \\
CoMET, best columns vs.\ all & +1.4\,{\scriptsize$\pm$0.8} & +2.2\,{\scriptsize$\pm$1.2} & +2.0\,{\scriptsize$\pm$0.7} & +1.4\,{\scriptsize$\pm$0.6} \\
RoFE vs.\ CoMET, best columns & +0.7\,{\scriptsize$\pm$0.5} & +0.6\,{\scriptsize$\pm$0.5} & +0.6\,{\scriptsize$\pm$0.3} & +0.5\,{\scriptsize$\pm$0.2} \\
RoFE vs.\ CoMET, $\sqrt n$ columns & +3.6\,{\scriptsize$\pm$1.4} & +2.1\,{\scriptsize$\pm$0.9} & +1.7\,{\scriptsize$\pm$0.7} & +1.4\,{\scriptsize$\pm$0.4} \\
RoFE vs.\ CoMET, Gavish--Donoho & +3.1\,{\scriptsize$\pm$1.7} & +1.0\,{\scriptsize$\pm$0.8} & +0.7\,{\scriptsize$\pm$0.4} & +0.5\,{\scriptsize$\pm$0.2} \\
\midrule
\Block[l]{1-5}{\textit{Image + table, DINOv3}} & & & & \\
CoMET, best columns vs.\ all & +1.2\,{\scriptsize$\pm$0.7} & +2.1\,{\scriptsize$\pm$0.9} & +2.1\,{\scriptsize$\pm$0.8} & +1.5\,{\scriptsize$\pm$1.1} \\
RoFE vs.\ CoMET, best columns & +0.1\,{\scriptsize$\pm$0.5} & $-$0.5\,{\scriptsize$\pm$0.7} & $-$0.5\,{\scriptsize$\pm$0.6} & $-$0.3\,{\scriptsize$\pm$0.5} \\
RoFE vs.\ CoMET, $\sqrt n$ columns & +0.6\,{\scriptsize$\pm$0.7} & $-$0.1\,{\scriptsize$\pm$0.8} & $-$0.3\,{\scriptsize$\pm$0.7} & $-$0.2\,{\scriptsize$\pm$0.6} \\
RoFE vs.\ CoMET, Gavish--Donoho & +0.3\,{\scriptsize$\pm$0.8} & $-$0.5\,{\scriptsize$\pm$0.7} & $-$0.5\,{\scriptsize$\pm$0.6} & $-$0.3\,{\scriptsize$\pm$0.5} \\
\midrule
\Block[l]{1-5}{\textit{Image only, SigLIP}} & & & & \\
CoMET, best columns vs.\ all & +1.4\,{\scriptsize$\pm$0.9} & +1.9\,{\scriptsize$\pm$1.1} & +1.9\,{\scriptsize$\pm$0.7} & +1.6\,{\scriptsize$\pm$0.7} \\
RoFE vs.\ CoMET, best columns & +0.7\,{\scriptsize$\pm$0.5} & +0.6\,{\scriptsize$\pm$0.4} & +0.6\,{\scriptsize$\pm$0.4} & +0.5\,{\scriptsize$\pm$0.3} \\
RoFE vs.\ CoMET, $\sqrt n$ columns & +3.4\,{\scriptsize$\pm$1.3} & +2.2\,{\scriptsize$\pm$1.1} & +1.9\,{\scriptsize$\pm$0.9} & +1.5\,{\scriptsize$\pm$0.8} \\
RoFE vs.\ CoMET, Gavish--Donoho & +2.6\,{\scriptsize$\pm$1.4} & +1.1\,{\scriptsize$\pm$0.9} & +0.7\,{\scriptsize$\pm$0.4} & +0.5\,{\scriptsize$\pm$0.2} \\
\midrule
\Block[l]{1-5}{\textit{Image + table, SigLIP}} & & & & \\
CoMET, best columns vs.\ all & +1.2\,{\scriptsize$\pm$0.8} & +1.7\,{\scriptsize$\pm$0.8} & +1.7\,{\scriptsize$\pm$0.8} & +1.6\,{\scriptsize$\pm$1.0} \\
RoFE vs.\ CoMET, best columns & +0.1\,{\scriptsize$\pm$0.4} & $-$0.1\,{\scriptsize$\pm$0.4} & $-$0.2\,{\scriptsize$\pm$0.4} & $-$0.2\,{\scriptsize$\pm$0.2} \\
RoFE vs.\ CoMET, $\sqrt n$ columns & +1.1\,{\scriptsize$\pm$0.6} & +0.2\,{\scriptsize$\pm$0.6} & +0.2\,{\scriptsize$\pm$0.5} & +0.1\,{\scriptsize$\pm$0.3} \\
RoFE vs.\ CoMET, Gavish--Donoho & +0.5\,{\scriptsize$\pm$0.6} & $-$0.1\,{\scriptsize$\pm$0.5} & $-$0.2\,{\scriptsize$\pm$0.4} & $-$0.1\,{\scriptsize$\pm$0.2} \\
\bottomrule
\end{NiceTabular}

%% file: tables/fewer_columns_text.tex
\tablestyle
\begin{NiceTabular}{lcccc}
\CodeBefore
  \rowcolor{tablesection}{3,8,13,18}
\Body
\toprule
\Block{2-1}{\textbf{Comparison}} & \Block{1-4}{\textbf{$\Delta$ AUROC} by labeled rows per class} & & & \\
 & \textbf{10} & \textbf{25} & \textbf{50} & \textbf{100} \\
\midrule
\Block[l]{1-5}{\textit{Text only, ELECTRA}} & & & & \\
CoMET, best columns vs.\ all & +0.6\,{\scriptsize$\pm$0.5} & +1.7\,{\scriptsize$\pm$0.9} & +2.0\,{\scriptsize$\pm$1.1} & +1.1\,{\scriptsize$\pm$0.5} \\
RoFE vs.\ CoMET, best columns & +1.2\,{\scriptsize$\pm$0.3} & +0.6\,{\scriptsize$\pm$0.3} & +0.5\,{\scriptsize$\pm$0.3} & +0.3\,{\scriptsize$\pm$0.4} \\
RoFE vs.\ CoMET, $\sqrt n$ columns & +4.6\,{\scriptsize$\pm$1.4} & +4.1\,{\scriptsize$\pm$1.4} & +3.2\,{\scriptsize$\pm$1.0} & +2.7\,{\scriptsize$\pm$0.9} \\
RoFE vs.\ CoMET, Gavish--Donoho & +3.4\,{\scriptsize$\pm$1.3} & +1.9\,{\scriptsize$\pm$1.2} & +0.8\,{\scriptsize$\pm$0.5} & +0.3\,{\scriptsize$\pm$0.4} \\
\midrule
\Block[l]{1-5}{\textit{Text + table, ELECTRA}} & & & & \\
CoMET, best columns vs.\ all & +0.7\,{\scriptsize$\pm$0.4} & +1.7\,{\scriptsize$\pm$1.0} & +2.6\,{\scriptsize$\pm$1.3} & +1.3\,{\scriptsize$\pm$0.6} \\
RoFE vs.\ CoMET, best columns & +0.2\,{\scriptsize$\pm$0.3} & +0.1\,{\scriptsize$\pm$0.6} & $-$0.1\,{\scriptsize$\pm$0.4} & $-$0.1\,{\scriptsize$\pm$0.2} \\
RoFE vs.\ CoMET, $\sqrt n$ columns & +1.2\,{\scriptsize$\pm$1.0} & +1.0\,{\scriptsize$\pm$1.3} & +0.7\,{\scriptsize$\pm$0.9} & +0.8\,{\scriptsize$\pm$0.6} \\
RoFE vs.\ CoMET, Gavish--Donoho & +0.9\,{\scriptsize$\pm$0.9} & +0.4\,{\scriptsize$\pm$1.1} & $-$0.1\,{\scriptsize$\pm$0.4} & $-$0.1\,{\scriptsize$\pm$0.2} \\
\midrule
\Block[l]{1-5}{\textit{Text only, E5}} & & & & \\
CoMET, best columns vs.\ all & +0.8\,{\scriptsize$\pm$0.6} & +1.7\,{\scriptsize$\pm$1.0} & +2.4\,{\scriptsize$\pm$0.9} & +1.8\,{\scriptsize$\pm$0.9} \\
RoFE vs.\ CoMET, best columns & +0.7\,{\scriptsize$\pm$0.5} & +0.5\,{\scriptsize$\pm$0.4} & +0.4\,{\scriptsize$\pm$0.3} & +0.4\,{\scriptsize$\pm$0.2} \\
RoFE vs.\ CoMET, $\sqrt n$ columns & +3.5\,{\scriptsize$\pm$1.7} & +2.4\,{\scriptsize$\pm$1.4} & +1.3\,{\scriptsize$\pm$0.6} & +1.1\,{\scriptsize$\pm$0.3} \\
RoFE vs.\ CoMET, Gavish--Donoho & +4.5\,{\scriptsize$\pm$2.3} & +2.1\,{\scriptsize$\pm$1.6} & +0.8\,{\scriptsize$\pm$0.6} & +0.5\,{\scriptsize$\pm$0.3} \\
\midrule
\Block[l]{1-5}{\textit{Text + table, E5}} & & & & \\
CoMET, best columns vs.\ all & +1.0\,{\scriptsize$\pm$0.6} & +1.9\,{\scriptsize$\pm$1.2} & +2.5\,{\scriptsize$\pm$1.3} & +1.5\,{\scriptsize$\pm$0.8} \\
RoFE vs.\ CoMET, best columns & 0.0\,{\scriptsize$\pm$0.5} & $-$0.3\,{\scriptsize$\pm$0.8} & $-$0.1\,{\scriptsize$\pm$0.3} & 0.0\,{\scriptsize$\pm$0.3} \\
RoFE vs.\ CoMET, $\sqrt n$ columns & +1.2\,{\scriptsize$\pm$1.5} & +0.3\,{\scriptsize$\pm$1.3} & +0.1\,{\scriptsize$\pm$0.5} & +0.1\,{\scriptsize$\pm$0.4} \\
RoFE vs.\ CoMET, Gavish--Donoho & +1.6\,{\scriptsize$\pm$1.9} & +0.2\,{\scriptsize$\pm$1.3} & 0.0\,{\scriptsize$\pm$0.4} & 0.0\,{\scriptsize$\pm$0.4} \\
\bottomrule
\end{NiceTabular}

%% file: tables/rotation_width.tex
\tablestyle
\begin{NiceTabular}{lcccc}
\CodeBefore
  \rowcolor{tablesection}{3,14,19,30,35,38,41,44}
\Body
\toprule
\Block{2-1}{\textbf{PFN and columns}} & \Block{1-4}{\textbf{$\Delta$ AUROC} by labeled rows per class} & & & \\
 & \textbf{10} & \textbf{25} & \textbf{50} & \textbf{100} \\
\midrule
\Block[l]{1-5}{\textit{Image only, DINOv3}} & & & & \\
TabICLv2, 32 columns & +0.7\,{\scriptsize$\pm$0.3} & +0.6\,{\scriptsize$\pm$0.3} & +0.4\,{\scriptsize$\pm$0.2} & +0.2\,{\scriptsize$\pm$0.1} \\
TabICLv2, 32 columns, four rotations & +1.1\,{\scriptsize$\pm$0.4} & +0.9\,{\scriptsize$\pm$0.4} & +0.5\,{\scriptsize$\pm$0.2} & +0.3\,{\scriptsize$\pm$0.1} \\
TabICLv2, Gavish--Donoho, four rotations & +0.9\,{\scriptsize$\pm$0.5} & +0.7\,{\scriptsize$\pm$0.3} & +0.6\,{\scriptsize$\pm$0.2} & +0.6\,{\scriptsize$\pm$0.1} \\
TabICLv2, $\min(n-1,256)$ columns & +1.7\,{\scriptsize$\pm$0.5} & +2.4\,{\scriptsize$\pm$0.9} & +2.4\,{\scriptsize$\pm$0.7} & +1.6\,{\scriptsize$\pm$0.7} \\
TabPFN 2.5, $d=8$ & +0.8\,{\scriptsize$\pm$0.2} & +0.4\,{\scriptsize$\pm$0.2} & +0.2\,{\scriptsize$\pm$0.1} & +0.1\,{\scriptsize$\pm$0.1} \\
TabPFN 2.5, $d=16$ & +1.2\,{\scriptsize$\pm$0.3} & +0.6\,{\scriptsize$\pm$0.2} & +0.3\,{\scriptsize$\pm$0.1} & +0.2\,{\scriptsize$\pm$0.1} \\
TabPFN 2.5, $d=32$ & +1.9\,{\scriptsize$\pm$0.4} & +1.1\,{\scriptsize$\pm$0.3} & +0.7\,{\scriptsize$\pm$0.2} & +0.6\,{\scriptsize$\pm$0.2} \\
TabPFN 2.5, $d=48$ & +2.3\,{\scriptsize$\pm$0.5} & +1.5\,{\scriptsize$\pm$0.4} & +0.9\,{\scriptsize$\pm$0.3} & +0.8\,{\scriptsize$\pm$0.3} \\
TabPFN 2.5, $d=64$ & +2.5\,{\scriptsize$\pm$0.6} & +1.7\,{\scriptsize$\pm$0.5} & +1.1\,{\scriptsize$\pm$0.4} & +0.9\,{\scriptsize$\pm$0.3} \\
TabPFN 2.5, $d=96$ & +2.7\,{\scriptsize$\pm$0.6} & +1.9\,{\scriptsize$\pm$0.5} & +1.3\,{\scriptsize$\pm$0.5} & +1.0\,{\scriptsize$\pm$0.4} \\
\midrule
\Block[l]{1-5}{\textit{Image + table, DINOv3}} & & & & \\
TabICLv2, 32 columns & +0.4\,{\scriptsize$\pm$0.3} & 0.0\,{\scriptsize$\pm$0.3} & 0.0\,{\scriptsize$\pm$0.1} & $-$0.1\,{\scriptsize$\pm$0.1} \\
TabICLv2, 32 columns, four rotations & +0.6\,{\scriptsize$\pm$0.3} & +0.3\,{\scriptsize$\pm$0.3} & +0.1\,{\scriptsize$\pm$0.2} & 0.0\,{\scriptsize$\pm$0.2} \\
TabICLv2, Gavish--Donoho, four rotations & +0.4\,{\scriptsize$\pm$0.3} & +0.2\,{\scriptsize$\pm$0.2} & +0.1\,{\scriptsize$\pm$0.2} & +0.2\,{\scriptsize$\pm$0.2} \\
TabICLv2, $\min(n-1,256)$ columns & +0.9\,{\scriptsize$\pm$0.4} & +1.1\,{\scriptsize$\pm$0.6} & +1.3\,{\scriptsize$\pm$0.7} & +1.0\,{\scriptsize$\pm$0.8} \\
\midrule
\Block[l]{1-5}{\textit{Image only, SigLIP}} & & & & \\
TabICLv2, 32 columns & +0.6\,{\scriptsize$\pm$0.3} & +0.4\,{\scriptsize$\pm$0.2} & +0.3\,{\scriptsize$\pm$0.2} & +0.2\,{\scriptsize$\pm$0.1} \\
TabICLv2, 32 columns, four rotations & +0.9\,{\scriptsize$\pm$0.4} & +0.7\,{\scriptsize$\pm$0.4} & +0.5\,{\scriptsize$\pm$0.2} & +0.3\,{\scriptsize$\pm$0.2} \\
TabICLv2, Gavish--Donoho, four rotations & +0.8\,{\scriptsize$\pm$0.5} & +0.8\,{\scriptsize$\pm$0.3} & +0.7\,{\scriptsize$\pm$0.2} & +0.6\,{\scriptsize$\pm$0.2} \\
TabICLv2, $\min(n-1,256)$ columns & +1.8\,{\scriptsize$\pm$0.6} & +2.2\,{\scriptsize$\pm$0.8} & +2.2\,{\scriptsize$\pm$0.7} & +1.7\,{\scriptsize$\pm$0.8} \\
TabPFN 2.5, $d=8$ & +0.5\,{\scriptsize$\pm$0.2} & +0.5\,{\scriptsize$\pm$0.3} & +0.3\,{\scriptsize$\pm$0.1} & +0.3\,{\scriptsize$\pm$0.1} \\
TabPFN 2.5, $d=16$ & +1.0\,{\scriptsize$\pm$0.2} & +0.7\,{\scriptsize$\pm$0.3} & +0.4\,{\scriptsize$\pm$0.2} & +0.3\,{\scriptsize$\pm$0.2} \\
TabPFN 2.5, $d=32$ & +1.7\,{\scriptsize$\pm$0.4} & +1.2\,{\scriptsize$\pm$0.4} & +0.8\,{\scriptsize$\pm$0.3} & +0.5\,{\scriptsize$\pm$0.3} \\
TabPFN 2.5, $d=48$ & +2.0\,{\scriptsize$\pm$0.4} & +1.5\,{\scriptsize$\pm$0.6} & +1.0\,{\scriptsize$\pm$0.3} & +0.6\,{\scriptsize$\pm$0.3} \\
TabPFN 2.5, $d=64$ & +2.2\,{\scriptsize$\pm$0.4} & +1.6\,{\scriptsize$\pm$0.5} & +1.1\,{\scriptsize$\pm$0.4} & +0.8\,{\scriptsize$\pm$0.4} \\
TabPFN 2.5, $d=96$ & +2.4\,{\scriptsize$\pm$0.5} & +1.8\,{\scriptsize$\pm$0.6} & +1.3\,{\scriptsize$\pm$0.4} & +0.8\,{\scriptsize$\pm$0.4} \\
\midrule
\Block[l]{1-5}{\textit{Image + table, SigLIP}} & & & & \\
TabICLv2, 32 columns & +0.1\,{\scriptsize$\pm$0.2} & 0.0\,{\scriptsize$\pm$0.1} & 0.0\,{\scriptsize$\pm$0.2} & 0.0\,{\scriptsize$\pm$0.1} \\
TabICLv2, 32 columns, four rotations & +0.5\,{\scriptsize$\pm$0.3} & +0.2\,{\scriptsize$\pm$0.2} & +0.2\,{\scriptsize$\pm$0.1} & +0.1\,{\scriptsize$\pm$0.2} \\
TabICLv2, Gavish--Donoho, four rotations & +0.4\,{\scriptsize$\pm$0.2} & +0.3\,{\scriptsize$\pm$0.1} & +0.1\,{\scriptsize$\pm$0.1} & +0.1\,{\scriptsize$\pm$0.2} \\
TabICLv2, $\min(n-1,256)$ columns & +0.9\,{\scriptsize$\pm$0.6} & +1.2\,{\scriptsize$\pm$0.7} & +1.3\,{\scriptsize$\pm$0.7} & +1.1\,{\scriptsize$\pm$0.9} \\
\midrule
\Block[l]{1-5}{\textit{Text only, ELECTRA}} & & & & \\
TabICLv2, 32 columns & +0.4\,{\scriptsize$\pm$0.4} & +0.1\,{\scriptsize$\pm$0.2} & +0.2\,{\scriptsize$\pm$0.2} & +0.1\,{\scriptsize$\pm$0.1} \\
TabICLv2, $\min(n-1,256)$ columns & +1.3\,{\scriptsize$\pm$0.6} & +1.8\,{\scriptsize$\pm$0.8} & +2.0\,{\scriptsize$\pm$0.9} & +1.0\,{\scriptsize$\pm$0.7} \\
\midrule
\Block[l]{1-5}{\textit{Text + table, ELECTRA}} & & & & \\
TabICLv2, 32 columns & $-$0.2\,{\scriptsize$\pm$0.2} & 0.0\,{\scriptsize$\pm$0.2} & 0.0\,{\scriptsize$\pm$0.2} & 0.0\,{\scriptsize$\pm$0.1} \\
TabICLv2, $\min(n-1,256)$ columns & +0.5\,{\scriptsize$\pm$0.4} & +1.4\,{\scriptsize$\pm$0.7} & +2.1\,{\scriptsize$\pm$1.1} & +0.9\,{\scriptsize$\pm$0.7} \\
\midrule
\Block[l]{1-5}{\textit{Text only, E5}} & & & & \\
TabICLv2, 32 columns & +0.6\,{\scriptsize$\pm$0.3} & +0.3\,{\scriptsize$\pm$0.2} & +0.3\,{\scriptsize$\pm$0.2} & +0.3\,{\scriptsize$\pm$0.2} \\
TabICLv2, $\min(n-1,256)$ columns & +0.9\,{\scriptsize$\pm$0.5} & +1.9\,{\scriptsize$\pm$0.7} & +2.4\,{\scriptsize$\pm$0.8} & +2.0\,{\scriptsize$\pm$0.9} \\
\midrule
\Block[l]{1-5}{\textit{Text + table, E5}} & & & & \\
TabICLv2, 32 columns & +0.1\,{\scriptsize$\pm$0.3} & 0.0\,{\scriptsize$\pm$0.2} & 0.0\,{\scriptsize$\pm$0.1} & +0.1\,{\scriptsize$\pm$0.1} \\
TabICLv2, $\min(n-1,256)$ columns & +0.5\,{\scriptsize$\pm$0.3} & +1.1\,{\scriptsize$\pm$0.7} & +2.1\,{\scriptsize$\pm$1.1} & +1.3\,{\scriptsize$\pm$0.8} \\
\bottomrule
\end{NiceTabular}

%% file: tables/pca_source.tex
\tablestyle
\begin{NiceTabular}{lccccc}
\CodeBefore
  \rowcolor{tablesection}{3,10,17,24}
  \rowcolor{ourrow}{7,14,21,28}
\Body
\toprule
\Block{2-1}{\textbf{PCA fitted on}} & \Block{2-1}{\textbf{Rotated}} & \Block{1-4}{\textbf{AUROC} ($\uparrow$)\\ by labeled rows per class} & & & \\
 &  & \textbf{10} & \textbf{25} & \textbf{50} & \textbf{100} \\
\midrule
\Block[l]{1-6}{\textit{Image only, DINOv3}} & & & & & \\
Labeled rows (CoMET) & \xmark & 68.0 & 72.0 & 75.2 & 78.3 \\
Whole pool & \xmark & 67.2 & 72.1 & 75.7 & 78.6 \\
ImageNet & \xmark & 65.5 & 72.2 & 76.2 & 79.1 \\
\textbf{Labeled rows (RoFE)} & \cmark & \textbf{70.1} & \textbf{74.8} & \textbf{77.9} & \textbf{80.2} \\
Whole pool & \cmark & 69.9 & 74.7 & \textbf{77.9} & \textbf{80.2} \\
ImageNet & \cmark & 66.1 & 72.8 & 76.7 & 79.4 \\
\midrule
\Block[l]{1-6}{\textit{Image + table, DINOv3}} & & & & & \\
Labeled rows (CoMET) & \xmark & 73.9 & 78.2 & 80.9 & 83.5 \\
Whole pool & \xmark & 73.6 & 78.7 & 81.8 & 84.2 \\
ImageNet & \xmark & 72.7 & 78.5 & 81.6 & 84.1 \\
\textbf{Labeled rows (RoFE)} & \cmark & \textbf{75.2} & \textbf{79.7} & 82.4 & \textbf{84.7} \\
Whole pool & \cmark & 75.0 & 79.6 & \textbf{82.5} & 84.6 \\
ImageNet & \cmark & 73.3 & 78.9 & 81.9 & 84.3 \\
\midrule
\Block[l]{1-6}{\textit{Image only, SigLIP}} & & & & & \\
Labeled rows (CoMET) & \xmark & 71.0 & 75.1 & 78.3 & 80.9 \\
Whole pool & \xmark & 70.3 & 75.1 & 78.8 & 81.3 \\
ImageNet & \xmark & 69.7 & 75.6 & 79.4 & 81.9 \\
\textbf{Labeled rows (RoFE)} & \cmark & \textbf{73.1} & \textbf{77.6} & \textbf{80.8} & 82.9 \\
Whole pool & \cmark & 72.8 & \textbf{77.6} & \textbf{80.8} & \textbf{83.0} \\
ImageNet & \cmark & 70.4 & 76.2 & 79.7 & 82.1 \\
\midrule
\Block[l]{1-6}{\textit{Image + table, SigLIP}} & & & & & \\
Labeled rows (CoMET) & \xmark & 75.7 & 80.0 & 82.7 & 84.8 \\
Whole pool & \xmark & 75.6 & 80.6 & 83.5 & 85.5 \\
ImageNet & \xmark & 75.0 & 80.4 & 83.5 & 85.7 \\
\textbf{Labeled rows (RoFE)} & \cmark & \textbf{77.0} & 81.5 & 84.2 & \textbf{86.2} \\
Whole pool & \cmark & 76.9 & \textbf{81.6} & \textbf{84.3} & \textbf{86.2} \\
ImageNet & \cmark & 75.7 & 80.9 & 83.8 & 85.8 \\
\bottomrule
\end{NiceTabular}

%% file: tables/pca_source_256.tex
\tablestyle
\begin{NiceTabular}{lcccccc}
\CodeBefore
  \rowcolor{tablesection}{3,9,15,21}
  \rowcolor{ourrow}{4,10,16,22}
\Body
\toprule
\Block{2-1}{\textbf{PCA fitted on}} & \Block{2-1}{\textbf{Columns}} & \Block{2-1}{\textbf{Rotated}} & \Block{1-4}{\textbf{AUROC} ($\uparrow$)\\ by labeled rows per class} & & & \\
 &  &  & \textbf{10} & \textbf{25} & \textbf{50} & \textbf{100} \\
\midrule
\Block[l]{1-7}{\textit{Image only, DINOv3}} & & & & & & \\
\textbf{Labeled rows (RoFE)} & $w$ & \cmark & 70.1 & 74.8 & 77.9 & \textbf{80.2} \\
Whole pool & 256 & \xmark & 62.6 & 70.0 & 74.7 & 78.4 \\
Whole pool & 256 & \cmark & \textbf{70.3} & \textbf{75.0} & \textbf{78.0} & \textbf{80.2} \\
ImageNet & 256 & \xmark & 68.9 & 73.8 & 76.6 & 79.2 \\
ImageNet & 256 & \cmark & 69.7 & 74.3 & 77.1 & 79.4 \\
\midrule
\Block[l]{1-7}{\textit{Image + table, DINOv3}} & & & & & & \\
\textbf{Labeled rows (RoFE)} & $w$ & \cmark & \textbf{75.2} & \textbf{79.7} & \textbf{82.4} & \textbf{84.7} \\
Whole pool & 256 & \xmark & 70.2 & 77.2 & 81.0 & 84.1 \\
Whole pool & 256 & \cmark & 73.9 & 79.2 & 82.3 & 84.6 \\
ImageNet & 256 & \xmark & 72.9 & 78.4 & 81.6 & 84.1 \\
ImageNet & 256 & \cmark & 73.6 & 78.9 & 81.9 & 84.3 \\
\midrule
\Block[l]{1-7}{\textit{Image only, SigLIP}} & & & & & & \\
\textbf{Labeled rows (RoFE)} & $w$ & \cmark & 73.1 & 77.6 & \textbf{80.8} & \textbf{82.9} \\
Whole pool & 256 & \xmark & 66.0 & 72.7 & 77.7 & 80.9 \\
Whole pool & 256 & \cmark & \textbf{73.2} & \textbf{77.9} & \textbf{80.8} & \textbf{82.9} \\
ImageNet & 256 & \xmark & 72.4 & 76.9 & 79.9 & 82.0 \\
ImageNet & 256 & \cmark & 73.0 & 77.4 & 80.2 & 82.2 \\
\midrule
\Block[l]{1-7}{\textit{Image + table, SigLIP}} & & & & & & \\
\textbf{Labeled rows (RoFE)} & $w$ & \cmark & \textbf{77.0} & \textbf{81.5} & \textbf{84.2} & \textbf{86.2} \\
Whole pool & 256 & \xmark & 72.3 & 79.0 & 82.6 & 85.2 \\
Whole pool & 256 & \cmark & 76.1 & 81.4 & 84.1 & 86.1 \\
ImageNet & 256 & \xmark & 75.5 & 80.6 & 83.6 & 85.7 \\
ImageNet & 256 & \cmark & 75.9 & 81.1 & 83.8 & 85.9 \\
\bottomrule
\end{NiceTabular}

%% file: tables/pca_source_text.tex
\tablestyle
\begin{NiceTabular}{lccccc}
\CodeBefore
  \rowcolor{tablesection}{3,10,17,24}
  \rowcolor{ourrow}{7,14,21,28}
\Body
\toprule
\Block{2-1}{\textbf{PCA fitted on}} & \Block{2-1}{\textbf{Rotated}} & \Block{1-4}{\textbf{AUROC} ($\uparrow$)\\ by labeled rows per class} & & & \\
 &  & \textbf{10} & \textbf{25} & \textbf{50} & \textbf{100} \\
\midrule
\Block[l]{1-6}{\textit{Text only, ELECTRA}} & & & & & \\
Labeled rows (CoMET) & \xmark & 62.8 & 66.7 & 69.7 & 73.3 \\
Whole pool & \xmark & 62.4 & 66.7 & 70.1 & 73.6 \\
CoMET's text sets & \xmark & 62.3 & 67.2 & 71.2 & 74.2 \\
\textbf{Labeled rows (RoFE)} & \cmark & \textbf{64.5} & \textbf{68.9} & \textbf{72.2} & \textbf{74.8} \\
Whole pool & \cmark & 63.7 & 68.7 & 72.1 & \textbf{74.8} \\
CoMET's text sets & \cmark & 62.4 & 67.6 & 71.5 & 74.3 \\
\midrule
\Block[l]{1-6}{\textit{Text + table, ELECTRA}} & & & & & \\
Labeled rows (CoMET) & \xmark & 71.5 & 75.0 & 77.3 & 80.7 \\
Whole pool & \xmark & 71.5 & 75.6 & 78.7 & 81.3 \\
CoMET's text sets & \xmark & 71.2 & 75.8 & 79.1 & 81.4 \\
\textbf{Labeled rows (RoFE)} & \cmark & \textbf{72.4} & \textbf{76.8} & \textbf{79.8} & \textbf{81.8} \\
Whole pool & \cmark & 72.1 & 76.6 & 79.7 & \textbf{81.8} \\
CoMET's text sets & \cmark & 71.5 & 76.1 & 79.5 & 81.6 \\
\midrule
\Block[l]{1-6}{\textit{Text only, E5}} & & & & & \\
Labeled rows (CoMET) & \xmark & 67.4 & 70.5 & 73.0 & 75.7 \\
Whole pool & \xmark & 66.8 & 70.5 & 73.7 & 76.6 \\
CoMET's text sets & \xmark & 66.6 & 71.5 & 74.8 & 77.1 \\
\textbf{Labeled rows (RoFE)} & \cmark & \textbf{68.9} & \textbf{72.7} & \textbf{75.8} & \textbf{78.0} \\
Whole pool & \cmark & 68.3 & 72.5 & 75.7 & \textbf{78.0} \\
CoMET's text sets & \cmark & 67.4 & 72.2 & 75.2 & 77.5 \\
\midrule
\Block[l]{1-6}{\textit{Text + table, E5}} & & & & & \\
Labeled rows (CoMET) & \xmark & 74.2 & 77.4 & 79.5 & 82.1 \\
Whole pool & \xmark & 73.8 & 77.8 & 80.8 & 83.0 \\
CoMET's text sets & \xmark & 73.4 & 78.2 & 81.2 & 83.0 \\
\textbf{Labeled rows (RoFE)} & \cmark & \textbf{75.2} & \textbf{79.0} & \textbf{81.9} & \textbf{83.6} \\
Whole pool & \cmark & 74.7 & 78.7 & 81.8 & 83.5 \\
CoMET's text sets & \cmark & 74.1 & 78.7 & 81.6 & 83.3 \\
\bottomrule
\end{NiceTabular}

%% file: tables/pca_source_256_text.tex
\tablestyle
\begin{NiceTabular}{lcccccc}
\CodeBefore
  \rowcolor{tablesection}{3,7,11,15}
  \rowcolor{ourrow}{4,8,12,16}
\Body
\toprule
\Block{2-1}{\textbf{PCA fitted on}} & \Block{2-1}{\textbf{Columns}} & \Block{2-1}{\textbf{Rotated}} & \Block{1-4}{\textbf{AUROC} ($\uparrow$)\\ by labeled rows per class} & & & \\
 &  &  & \textbf{10} & \textbf{25} & \textbf{50} & \textbf{100} \\
\midrule
\Block[l]{1-7}{\textit{Text only, ELECTRA}} & & & & & & \\
\textbf{Labeled rows (RoFE)} & $w$ & \cmark & \textbf{64.5} & \textbf{68.9} & \textbf{72.2} & \textbf{74.8} \\
Whole pool & 256 & \xmark & 60.1 & 65.0 & 69.3 & 73.3 \\
Whole pool & 256 & \cmark & 64.2 & 68.8 & \textbf{72.2} & \textbf{74.8} \\
\midrule
\Block[l]{1-7}{\textit{Text + table, ELECTRA}} & & & & & & \\
\textbf{Labeled rows (RoFE)} & $w$ & \cmark & \textbf{72.4} & \textbf{76.8} & \textbf{79.8} & \textbf{81.8} \\
Whole pool & 256 & \xmark & 68.5 & 73.8 & 78.0 & 81.1 \\
Whole pool & 256 & \cmark & 70.4 & 76.3 & 79.7 & \textbf{81.8} \\
\midrule
\Block[l]{1-7}{\textit{Text only, E5}} & & & & & & \\
\textbf{Labeled rows (RoFE)} & $w$ & \cmark & 68.9 & 72.7 & \textbf{75.8} & \textbf{78.0} \\
Whole pool & 256 & \xmark & 63.6 & 68.6 & 72.7 & 76.3 \\
Whole pool & 256 & \cmark & \textbf{69.0} & \textbf{72.9} & 75.7 & 77.9 \\
\midrule
\Block[l]{1-7}{\textit{Text + table, E5}} & & & & & & \\
\textbf{Labeled rows (RoFE)} & $w$ & \cmark & \textbf{75.2} & \textbf{79.0} & \textbf{81.9} & \textbf{83.6} \\
Whole pool & 256 & \xmark & 70.2 & 75.9 & 80.0 & 82.7 \\
Whole pool & 256 & \cmark & 73.4 & 78.3 & 81.7 & 83.4 \\
\bottomrule
\end{NiceTabular}

%% file: tables/need.tex
\tablestyle
\begin{NiceTabular}{lcccc}
\toprule
\Block{2-1}{\textbf{Comparison}} & \Block{1-4}{\textbf{$\Delta$ AUROC} by labeled rows per class} & & & \\
 & \textbf{10} & \textbf{25} & \textbf{50} & \textbf{100} \\
\midrule
Real tables: four rotations vs.\ none & $-$1.2\,{\scriptsize$\pm$1.1} & $-$0.9\,{\scriptsize$\pm$0.8} & $-$1.0\,{\scriptsize$\pm$0.8} & $-$1.2\,{\scriptsize$\pm$0.5} \\
Raw embedding: four rotations vs.\ none & +0.5\,{\scriptsize$\pm$0.2} & +0.7\,{\scriptsize$\pm$0.3} & +0.4\,{\scriptsize$\pm$0.2} & +0.4\,{\scriptsize$\pm$0.2} \\
Raw embedding vs.\ CoMET & +1.6\,{\scriptsize$\pm$0.5} & +1.1\,{\scriptsize$\pm$1.1} & +1.3\,{\scriptsize$\pm$0.8} & +0.8\,{\scriptsize$\pm$0.9} \\
RoFE vs.\ raw embedding & +0.5\,{\scriptsize$\pm$0.4} & +1.7\,{\scriptsize$\pm$0.8} & +1.4\,{\scriptsize$\pm$0.6} & +1.1\,{\scriptsize$\pm$0.5} \\
Raw embedding vs.\ CoMET, with the table & $-$2.7\,{\scriptsize$\pm$1.7} & $-$2.7\,{\scriptsize$\pm$1.8} & $-$1.7\,{\scriptsize$\pm$1.3} & $-$1.3\,{\scriptsize$\pm$1.1} \\
RoFE vs.\ raw embedding, with the table & +4.0\,{\scriptsize$\pm$1.7} & +4.2\,{\scriptsize$\pm$1.6} & +3.2\,{\scriptsize$\pm$1.0} & +2.5\,{\scriptsize$\pm$0.7} \\
Random orthonormal basis of the same width vs.\ RoFE & $-$2.2\,{\scriptsize$\pm$1.0} & $-$0.7\,{\scriptsize$\pm$0.3} & $-$0.5\,{\scriptsize$\pm$0.2} & $-$0.3\,{\scriptsize$\pm$0.2} \\
\bottomrule
\end{NiceTabular}

%% file: tables/comet_benchmarks.tex
\tablestyle
\begin{NiceTabular}{lcccc}
\toprule
\Block{2-1}{\textbf{Benchmark (tasks), encoders}} & \Block{1-4}{\textbf{$\Delta$ accuracy} by labeled rows per class} & & & \\
 & \textbf{10} & \textbf{25} & \textbf{50} & \textbf{100} \\
\midrule
ImageNet subsets (4), DINOv3 & +1.4\,{\scriptsize$\pm$1.0} & +1.6\,{\scriptsize$\pm$0.6} & +0.8\,{\scriptsize$\pm$0.2} & +0.4\,{\scriptsize$\pm$0.1} \\
Text only (4), ELECTRA & +2.9\,{\scriptsize$\pm$0.9} & +2.5\,{\scriptsize$\pm$1.4} & +2.0\,{\scriptsize$\pm$1.6} & +0.4\,{\scriptsize$\pm$0.4} \\
Multimodal (8), DINOv3 + ELECTRA & +1.8\,{\scriptsize$\pm$1.3} & +2.2\,{\scriptsize$\pm$1.0} & +2.0\,{\scriptsize$\pm$2.4} & +0.6\,{\scriptsize$\pm$1.2} \\
\bottomrule
\end{NiceTabular}

%% file: tables/pal.tex
\tablestyle
\begin{NiceTabular}{lcccc}
\CodeBefore
  \rowcolor{tablesection}{3,8}
  \rowcolor{ourrow}{6,11}
\Body
\toprule
\Block{2-1}{\textbf{Method}} & \Block{1-4}{\textbf{AUROC} ($\uparrow$) by labeled rows per class} & & & \\
 & \textbf{10} & \textbf{25} & \textbf{50} & \textbf{100} \\
\midrule
\Block[l]{1-5}{\textit{Image only, DINOv3}} & & & & \\
CoMET & 68.0 & 72.0 & 75.2 & 78.3 \\
CoMET + PAL & 66.6 & 72.5 & 76.4 & 79.0 \\
\textbf{RoFE} & \textbf{70.1} & \textbf{74.8} & \textbf{77.9} & \textbf{80.2} \\
$\Delta$ & +3.5\,{\scriptsize$\pm$1.0} & +2.3\,{\scriptsize$\pm$0.7} & +1.5\,{\scriptsize$\pm$0.5} & +1.1\,{\scriptsize$\pm$0.5} \\
\midrule
\Block[l]{1-5}{\textit{Image + table, DINOv3}} & & & & \\
CoMET & 73.9 & 78.2 & 80.9 & 83.5 \\
CoMET + PAL & 73.7 & 79.0 & 82.1 & 84.4 \\
\textbf{RoFE} & \textbf{75.2} & \textbf{79.7} & \textbf{82.4} & \textbf{84.7} \\
$\Delta$ & +1.5\,{\scriptsize$\pm$0.7} & +0.7\,{\scriptsize$\pm$0.6} & +0.3\,{\scriptsize$\pm$0.5} & +0.2\,{\scriptsize$\pm$0.3} \\
\bottomrule
\end{NiceTabular}

%% file: tables/many_classes.tex
\tablestyle
\begin{NiceTabular}{lccccccc}
\CodeBefore
  \rowcolor{tablesection}{3,10,17,24,31,38}
  \rowcolor{ourrow}{6,13,20,27,34,41}
\Body
\toprule
\Block{2-1}{\textbf{Method}} & \Block{1-7}{\textbf{AUROC} ($\uparrow$) by labeled rows per class} & & & & & & \\
 & \textbf{10} & \textbf{25} & \textbf{50} & \textbf{100} & \textbf{250} & \textbf{500} & \textbf{All} \\
\midrule
\Block[l]{1-8}{\textit{Hearthstone (11 classes), image only, DINOv3}} & & & & & & & \\
Linear probe & \textbf{68.3} & \textbf{72.3} & 75.4 & 78.2 & 81.6 & 82.1 & 82.0 \\
CoMET & 62.8 & 63.6 & 73.3 & 79.7 & 85.1 & 86.4 & 87.2 \\
\textbf{RoFE} & 66.5 & 72.0 & \textbf{76.7} & \textbf{80.2} & \textbf{85.5} & \textbf{86.6} & \textbf{87.9} \\
$\Delta$ & +3.7 & +8.3 & +3.4 & +0.4 & +0.4 & +0.2 & +0.7 \\
$\Delta$, with the table & +2.8 & +5.6 & +1.5 & +0.5 & +0.3 & +0.1 & +0.3 \\
$\Delta$, ten classes & +2.6 & +6.2 & +2.5 & +0.7 & +0.1 & $-$0.1 & +0.2 \\
\midrule
\Block[l]{1-8}{\textit{Hearthstone (11 classes), image only, SigLIP}} & & & & & & & \\
Linear probe & \textbf{75.1} & 79.3 & 81.9 & 84.4 & 86.9 & 87.5 & 87.8 \\
CoMET & 69.3 & 72.0 & 81.3 & 85.8 & 89.7 & 90.6 & 91.3 \\
\textbf{RoFE} & 74.2 & \textbf{79.6} & \textbf{83.1} & \textbf{86.2} & \textbf{90.0} & \textbf{90.8} & \textbf{91.5} \\
$\Delta$ & +5.0 & +7.6 & +1.7 & +0.4 & +0.3 & +0.2 & +0.3 \\
$\Delta$, with the table & +3.6 & +3.9 & +0.6 & +0.3 & +0.3 & +0.2 & +0.1 \\
$\Delta$, ten classes & +3.2 & +4.3 & +1.3 & +0.4 & +0.1 & +0.2 & 0.0 \\
\midrule
\Block[l]{1-8}{\textit{Wine reviews (30 classes), text only, ELECTRA}} & & & & & & & \\
Linear probe & 74.5 & 80.2 & \textbf{84.1} & \textbf{87.1} & \textbf{89.4} & \textbf{90.2} & \textbf{90.6} \\
CoMET & 67.4 & 78.6 & 83.5 & 86.3 & 88.4 & 89.2 & 89.7 \\
\textbf{RoFE} & \textbf{75.2} & \textbf{80.7} & \textbf{84.1} & 86.6 & 88.3 & 89.0 & 89.3 \\
$\Delta$ & +7.8 & +2.1 & +0.6 & +0.3 & $-$0.1 & $-$0.2 & $-$0.4 \\
$\Delta$, with the table & +1.3 & $-$0.1 & $-$0.4 & $-$0.4 & $-$0.5 & $-$0.4 & $-$0.5 \\
$\Delta$, ten classes & +1.6 & +1.7 & +0.2 & $-$0.2 & $-$0.7 & $-$0.7 & $-$0.7 \\
\midrule
\Block[l]{1-8}{\textit{Wine reviews (30 classes), text only, E5}} & & & & & & & \\
Linear probe & 92.5 & 94.6 & 95.7 & 96.7 & 97.2 & 97.5 & 97.6 \\
CoMET & 92.1 & 94.7 & 96.2 & \textbf{97.2} & \textbf{97.8} & \textbf{98.0} & \textbf{98.1} \\
\textbf{RoFE} & \textbf{93.0} & \textbf{95.4} & \textbf{96.4} & \textbf{97.2} & \textbf{97.8} & \textbf{98.0} & \textbf{98.1} \\
$\Delta$ & +0.9 & +0.6 & +0.2 & 0.0 & 0.0 & 0.0 & 0.0 \\
$\Delta$, with the table & +0.8 & +0.3 & 0.0 & $-$0.1 & $-$0.1 & $-$0.1 & $-$0.1 \\
$\Delta$, ten classes & +0.3 & +0.4 & 0.0 & 0.0 & 0.0 & 0.0 & 0.0 \\
\midrule
\Block[l]{1-8}{\textit{Spotify genres (114 classes), text only, ELECTRA}} & & & & & & & \\
Linear probe & \textbf{78.4} & 83.3 & 86.6 & 88.6 & -- & -- & 88.6 \\
CoMET & 76.0 & 82.7 & 86.4 & 89.0 & -- & -- & 89.0 \\
\textbf{RoFE} & \textbf{78.4} & \textbf{83.8} & \textbf{87.2} & \textbf{89.6} & -- & -- & \textbf{89.6} \\
$\Delta$ & +2.4 & +1.0 & +0.9 & +0.6 & -- & -- & +0.6 \\
$\Delta$, with the table & +0.9 & +0.6 & +0.5 & +0.3 & -- & -- & +0.3 \\
$\Delta$, ten classes & +2.4 & +2.3 & +0.5 & +0.2 & -- & -- & +0.2 \\
\midrule
\Block[l]{1-8}{\textit{Spotify genres (114 classes), text only, E5}} & & & & & & & \\
Linear probe & 84.3 & 88.5 & 90.7 & 92.3 & -- & -- & 92.3 \\
CoMET & \textbf{85.1} & 89.8 & 92.1 & 93.9 & -- & -- & 93.9 \\
\textbf{RoFE} & 84.9 & \textbf{90.3} & \textbf{93.1} & \textbf{94.9} & -- & -- & \textbf{94.9} \\
$\Delta$ & $-$0.2 & +0.5 & +1.0 & +1.0 & -- & -- & +1.0 \\
$\Delta$, with the table & +0.1 & +0.3 & +0.3 & +0.3 & -- & -- & +0.3 \\
$\Delta$, ten classes & +0.6 & +1.7 & +0.5 & +0.4 & -- & -- & +0.3 \\
\bottomrule
\end{NiceTabular}

%% file: tables/regression.tex
\tablestyle
\begin{NiceTabular}{lcccccccc}
\CodeBefore
  \rowcolor{tablesection}{3,10,17,24}
  \rowcolor{ourrow}{5,12,19,26}
\Body
\toprule
\Block{2-1}{\textbf{Method}} & \Block{1-8}{\textbf{$R^2$ (\%)} by labeled rows} & & & & & & & \\
 & \textbf{10} & \textbf{25} & \textbf{50} & \textbf{100} & \textbf{250} & \textbf{500} & \textbf{1000} & \textbf{All} \\
\midrule
\Block[l]{1-9}{\textit{Embedding only, DINOv3 / ELECTRA}} & & & & & & & & \\
CoMET & $-$16.7 & $-$25.3 & $-$11.3 & $-$8.8 & $-$9.6 & 21.0 & 33.7 & \textbf{50.5} \\
\textbf{\Method} & $-$9.6 & $-$11.5 & 0.9 & 11.6 & \textbf{24.6} & \textbf{28.6} & \textbf{34.9} & 49.8 \\
Ridge & \textbf{$-$1.1} & \textbf{0.8} & \textbf{7.9} & \textbf{13.7} & 21.8 & 25.5 & 30.0 & 37.1 \\
$\Delta$, mean & +8.8\,{\scriptsize$\pm$2.7} & +23.1\,{\scriptsize$\pm$10.7} & +18.4\,{\scriptsize$\pm$6.6} & +25.3\,{\scriptsize$\pm$10.4} & +59.2\,{\scriptsize$\pm$37.4} & +8.2\,{\scriptsize$\pm$3.5} & +2.4\,{\scriptsize$\pm$1.8} & +0.1\,{\scriptsize$\pm$0.7} \\
$\Delta$, median & +6.8 & +9.3 & +11.8 & +15.2 & +29.2 & +6.0 & +1.4 & $-$0.3 \\
Tasks higher & 36 & 40 & 40 & 39 & 39 & 34 & 31 & 18 \\
\midrule
\Block[l]{1-9}{\textit{Embedding only, SigLIP / E5}} & & & & & & & & \\
CoMET & $-$5.0 & $-$5.5 & 4.9 & 7.3 & 8.1 & 32.4 & \textbf{44.0} & 59.5 \\
\textbf{\Method} & $-$1.2 & 0.7 & 12.3 & 22.9 & \textbf{35.6} & \textbf{38.4} & 43.9 & \textbf{59.6} \\
Ridge & \textbf{2.4} & \textbf{7.5} & \textbf{15.3} & \textbf{23.9} & 31.8 & 32.5 & 37.9 & 45.7 \\
$\Delta$, mean & +5.5\,{\scriptsize$\pm$1.5} & +10.9\,{\scriptsize$\pm$6.4} & +11.2\,{\scriptsize$\pm$4.9} & +20.3\,{\scriptsize$\pm$8.8} & +52.2\,{\scriptsize$\pm$29.8} & +7.6\,{\scriptsize$\pm$2.9} & +2.3\,{\scriptsize$\pm$1.6} & +0.2\,{\scriptsize$\pm$0.8} \\
$\Delta$, median & +5.5 & +6.7 & +6.4 & +9.7 & +23.5 & +4.6 & +1.2 & 0.0 \\
Tasks higher & 38 & 39 & 40 & 39 & 40 & 38 & 30 & 19 \\
\midrule
\Block[l]{1-9}{\textit{Embedding + table, DINOv3 / ELECTRA}} & & & & & & & & \\
CoMET & $-$5.0 & $-$1.8 & 14.5 & 19.9 & 21.4 & 42.0 & 49.7 & \textbf{66.8} \\
\textbf{\Method} & $-$0.6 & \textbf{6.9} & \textbf{19.4} & \textbf{32.2} & \textbf{45.9} & \textbf{46.8} & \textbf{54.3} & 66.5 \\
Ridge & \textbf{$-$0.4} & 4.1 & 9.9 & 17.4 & 29.8 & 36.8 & 42.1 & 49.9 \\
$\Delta$, mean & +5.6\,{\scriptsize$\pm$2.1} & +15.1\,{\scriptsize$\pm$10.1} & +12.8\,{\scriptsize$\pm$3.4} & +18.6\,{\scriptsize$\pm$8.2} & +44.0\,{\scriptsize$\pm$31.2} & +7.8\,{\scriptsize$\pm$7.5} & +2.8\,{\scriptsize$\pm$3.6} & $-$0.6\,{\scriptsize$\pm$0.6} \\
$\Delta$, median & +3.9 & +7.6 & +9.2 & +10.9 & +19.5 & +2.0 & +0.2 & $-$0.5 \\
Tasks higher & 36 & 36 & 39 & 39 & 37 & 30 & 21 & 15 \\
\midrule
\Block[l]{1-9}{\textit{Embedding + table, SigLIP / E5}} & & & & & & & & \\
CoMET & 0.2 & 5.6 & 21.0 & 30.5 & 36.8 & 49.9 & 58.2 & \textbf{73.5} \\
\textbf{\Method} & 0.7 & \textbf{10.0} & \textbf{26.5} & \textbf{36.2} & \textbf{55.2} & \textbf{56.3} & \textbf{61.0} & 72.1 \\
Ridge & \textbf{2.6} & 9.0 & 16.9 & 24.7 & 35.4 & 39.5 & 46.1 & 56.3 \\
$\Delta$, mean & +2.5\,{\scriptsize$\pm$2.1} & +7.9\,{\scriptsize$\pm$3.7} & +8.5\,{\scriptsize$\pm$2.7} & +14.5\,{\scriptsize$\pm$7.2} & +38.0\,{\scriptsize$\pm$26.0} & +7.0\,{\scriptsize$\pm$6.3} & +2.6\,{\scriptsize$\pm$2.8} & 0.0\,{\scriptsize$\pm$0.8} \\
$\Delta$, median & +2.7 & +5.5 & +5.8 & +7.8 & +11.9 & +2.1 & +0.3 & $-$0.2 \\
Tasks higher & 35 & 39 & 39 & 39 & 38 & 31 & 22 & 18 \\
\bottomrule
\end{NiceTabular}

%% file: tables/regression_source.tex
\tablestyle
\begin{NiceTabular}{lcccccccc}
\CodeBefore
  \rowcolor{tablesection}{3,8,13,18}
\Body
\toprule
\Block{2-1}{\textbf{Tasks}} & \Block{1-8}{\textbf{$\Delta$ $R^2$ (\%), mean / median} by labeled rows} & & & & & & & \\
 & \textbf{10} & \textbf{25} & \textbf{50} & \textbf{100} & \textbf{250} & \textbf{500} & \textbf{1000} & \textbf{All} \\
\midrule
\Block[l]{1-9}{\textit{Embedding only, DINOv3 / ELECTRA}} & & & & & & & & \\
Curated (18) & +7.7/+8.1 & +21.2/+11.2 & +15.7/+11.7 & +16.1/+13.0 & +27.5/+27.1 & +5.5/+4.8 & +0.8/+1.2 & $-$0.6/$-$0.9 \\
Other (22) & +9.6/+6.8 & +24.7/+8.4 & +20.6/+11.8 & +32.8/+17.0 & +85.2/+33.7 & +10.2/+6.7 & +3.6/+1.7 & +0.6/+0.2 \\
Image tasks (20) & +7.0/+6.9 & +9.0/+7.1 & +9.8/+7.6 & +13.5/+12.1 & +25.1/+25.5 & +6.3/+6.0 & +1.9/+1.8 & +0.6/+0.3 \\
Text tasks (20) & +10.5/+6.6 & +37.3/+19.6 & +26.9/+19.6 & +37.1/+24.9 & +93.3/+36.8 & +9.9/+5.6 & +2.8/+1.1 & $-$0.5/$-$0.8 \\
\midrule
\Block[l]{1-9}{\textit{Embedding only, SigLIP / E5}} & & & & & & & & \\
Curated (18) & +4.5/+4.3 & +8.1/+6.0 & +10.9/+6.9 & +11.8/+8.4 & +34.7/+23.5 & +6.5/+4.7 & +1.1/+1.0 & $-$1.1/$-$1.0 \\
Other (22) & +6.3/+5.7 & +13.2/+6.9 & +11.5/+5.7 & +27.3/+10.3 & +66.6/+25.7 & +8.5/+4.5 & +3.2/+1.2 & +1.3/+0.2 \\
Image tasks (20) & +5.6/+5.7 & +8.0/+6.9 & +9.1/+5.9 & +10.6/+9.1 & +19.2/+12.3 & +5.4/+4.1 & +1.4/+1.3 & +0.1/+0.1 \\
Text tasks (20) & +5.4/+4.5 & +13.8/+6.0 & +13.3/+6.9 & +30.0/+17.7 & +85.3/+38.3 & +9.7/+5.0 & +3.1/+1.0 & +0.4/$-$0.6 \\
\midrule
\Block[l]{1-9}{\textit{Embedding + table, DINOv3 / ELECTRA}} & & & & & & & & \\
Curated (18) & +6.6/+5.0 & +13.3/+10.1 & +12.7/+11.1 & +13.3/+12.7 & +20.6/+19.5 & +3.3/+2.6 & +0.2/0.0 & $-$1.1/$-$0.9 \\
Other (22) & +4.8/+3.4 & +16.5/+5.9 & +13.0/+8.5 & +22.8/+10.2 & +63.1/+19.3 & +11.0/+1.0 & +4.7/+0.2 & $-$0.2/$-$0.1 \\
Image tasks (20) & +6.4/+4.0 & +5.3/+6.3 & +8.4/+6.5 & +10.2/+9.9 & +14.0/+13.5 & +2.0/+1.0 & +0.1/+0.1 & $-$0.4/$-$0.1 \\
Text tasks (20) & +4.9/+3.8 & +24.9/+12.3 & +17.3/+16.2 & +26.9/+14.6 & +73.9/+30.1 & +12.9/+3.4 & +5.0/+0.3 & $-$0.8/$-$0.8 \\
\midrule
\Block[l]{1-9}{\textit{Embedding + table, SigLIP / E5}} & & & & & & & & \\
Curated (18) & +1.5/+2.8 & +6.0/+5.0 & +9.6/+6.4 & +9.8/+8.3 & +26.0/+13.0 & +3.6/+2.4 & +0.5/+0.4 & $-$0.7/$-$0.6 \\
Other (22) & +3.4/+2.6 & +9.5/+6.5 & +7.6/+4.4 & +18.4/+7.4 & +47.8/+11.9 & +9.4/+1.6 & +4.2/+0.3 & +0.6/0.0 \\
Image tasks (20) & +1.7/+2.7 & +7.2/+6.5 & +7.8/+5.2 & +7.6/+6.9 & +10.3/+7.3 & +1.9/+1.8 & +0.2/+0.1 & $-$0.1/0.0 \\
Text tasks (20) & +3.3/+2.8 & +8.7/+4.6 & +9.2/+7.4 & +21.4/+13.1 & +65.7/+21.4 & +11.5/+2.5 & +4.6/+0.8 & +0.1/$-$0.3 \\
\bottomrule
\end{NiceTabular}

%% file: tables/regression_tasks.tex
\tablestyle
\begin{NiceTabular}{lrl@{\hspace{14pt}}lrl}
\toprule
\textbf{Image task} & \textbf{Pool} & \textbf{$R^2<-100\%$} & \textbf{Text task} & \textbf{Pool} & \textbf{$R^2<-100\%$} \\
\midrule
AIRBNB\_NYC$^\dagger$ & 10{,}000 & Ridge & AIRBNB\_SEATTLE$^\dagger$ & 3{,}437 & -- \\
AMAZON\_BEST\_SELLER & 3{,}140 & -- & ANIME\_PLANET$^\dagger$ & 10{,}000 & -- \\
AMAZON\_PACKAGES & 10{,}000 & -- & BABIES\_PRICES & 4{,}577 & -- \\
DVM\_CAR$^\dagger$ & 10{,}000 & -- & BOOK\_PRICE & 4{,}491 & -- \\
FLIPKART\_RATIO$^\dagger$ & 10{,}000 & -- & BOOK\_READABILITY & 4{,}252 & -- \\
GOIAS\_HOUSES$^\dagger$ & 8{,}847 & -- & CHOCOLATE\_BAR\_RATINGS$^\dagger$ & 1{,}616 & -- \\
HNM\_FASHION & 10{,}000 & -- & FIFA22\_WAGES$^\dagger$ & 10{,}000 & CoMET \\
KAMERNET\_SIZE$^\dagger$ & 1{,}475 & -- & MERCARI\_MARKETPLACE & 10{,}000 & -- \\
KHAADI\_CLOTHES & 360 & CoMET, \Method & MONTGOMERY\_SALARIES & 8{,}306 & -- \\
LAHAINA\_AUCTION$^\dagger$ & 872 & -- & RAMEN\_RATINGS$^\dagger$ & 3{,}695 & -- \\
LETTERBOXD\_MOVIES & 10{,}000 & -- & ROTTEN\_TOMATOES & 6{,}443 & -- \\
MANGO\_MASS & 492 & -- & SCIMAGOJR\_IMPACT & 10{,}000 & -- \\
MKPHOTO\_BOTS & 10{,}000 & -- & USED\_CAR\_PAKISTAN$^\dagger$ & 10{,}000 & CoMET \\
PAINTING\_PRICE & 10{,}000 & -- & USED\_CAR\_SAUDI$^\dagger$ & 7{,}232 & CoMET \\
ROMANIA\_PRICE$^\dagger$ & 10{,}000 & -- & VANCOUVER\_SALARIES & 10{,}000 & -- \\
SAO\_PAULO\_HOUSES$^\dagger$ & 10{,}000 & -- & VIDEO\_GAMES\_SALES & 10{,}000 & CoMET, \Method \\
SOCAL\_HOUSES$^\dagger$ & 10{,}000 & -- & WIKILIQ\_PRICES$^\dagger$ & 10{,}000 & CoMET, \Method \\
TOKOPEDIA\_WEIGHT$^\dagger$ & 1{,}078 & -- & WINE\_POLISH\_MARKET$^\dagger$ & 2{,}023 & CoMET \\
WATCH\_TIER$^\dagger$ & 1{,}134 & -- & WINE\_VIVINO\_SPAIN$^\dagger$ & 7{,}785 & -- \\
ZEPTO\_PRICE$^\dagger$ & 10{,}000 & -- & ZOMATO\_RESTAURANTS & 10{,}000 & -- \\
\bottomrule
\end{NiceTabular}